\def\gmode{} %%use this to include figures

 \def\gdriver{dvipdfmx}%for dvipdfmx to make pdf from dvi
\newcommand{\dtitle}[1]{\title{ \if \gmode \else
\color{red} Demo mode!\\
comment out \textbackslash def \textbackslash gmode\{demo\} at the header to include figures \color{black}\\
\fi
#1 }}
\ifx\pdfoutput\undefined
\else
 \def\gdriver{}
\fi
\documentclass[\gmode,\gdriver,10pt,twocolumn,letterpaper]{article}

\usepackage{iccv}
\usepackage{times}
\usepackage{epsfig}
\usepackage{graphicx}
\usepackage{amsmath}
\usepackage{amssymb}

\usepackage{multirow}

\usepackage{color}
\newcommand{\fnote}[1]{}

\newcommand{\knote}[1]{{\color{red} \bf #1 \color{black}}}

\newcommand{\bnote}[1]{{\color{red} \bf #1 \color{black}}}
\newcommand{\onote}[1]{{\color{cyan} \bf #1 \color{black}}}
\newcommand{\mnote}[1]{}

\newcommand{\mcut}[1]{}
\newcommand{\kcut}[1]{}
\newcommand{\jptext}[1]{}

\ifx\pdfoutput\undefined
\else
 \usepackage{CJKutf8}
 \renewcommand{\fnote}[1]{}
 \renewcommand{\knote}[1]{}
 \renewcommand{\mnote}[1]{}
 \renewcommand{\bnote}[1]{}
\fi

\newcommand{\argmax}{\mathop{\rm argmax}\limits}

\if 0

\newcommand{\ie}{{\it i.e.}}
\newcommand{\eg}{{\it e.g.}}

\fi

\iccvfinalcopy % *** Uncomment this line for the final submission

\def\iccvPaperID{2216} % *** Enter the ICCV Paper ID here

\ificcvfinal\fi
\begin{document}

%%%%%%%%% TITLE
\dtitle{
\vspace{-4mm}
Temporal shape super-resolution by intra-frame motion encoding \\using
high-fps structured light
%Temporal shape super-resolution for structured light system \\ using 
%high-speed alternating pattern projection
\vspace{-2mm}
}
%       Temporal shape super-resolution for structured light using coded motion blur}

\author{
Yuki Shiba, Satoshi Ono\\
Kagoshima University, Japan\\
%Kagoshima, Japan \\
%{\tt\small \{kawasaki,ono\}@ibe.kagoshima-u.ac.jp}
\and
Ryo Furukawa, Shinsaku Hiura\\
Hiroshima City University, Japan\\
%Hiroshima, Japan\\ 
%{\tt\small \{ryo-f,hiura\}@hiroshima-cu.ac.jp}
\and
Hiroshi Kawasaki\\
Kyushu University, Japan\\
%Fukuoka, Japan\\
%{\tt\small kawasaki@ait.kyushu-u.ac.jp}
}

%\onecolumn

\graphicspath{{../170721-cvpr-light-coded-motion-blur/}{./}}

\maketitle
%\thispagestyle{empty}

%%%%%%%%% ABSTRACT
\begin{abstract}
%Active 3D scanning techniques for dynamic scenes
%are becoming increasingly important due to strong demands in
%various fields; self-driving cars and surveillance, to name just two. However, 
%    there is just a few solution yet.

\vspace{-2mm}
One of the solutions of depth imaging of moving scene is to project
a static pattern on the object and use  just a single image for reconstruction.
However, if the motion of the object is too fast with respect to the exposure
time of the image sensor, patterns on the captured image are blurred and
reconstruction fails.
In this paper, we impose multiple projection patterns into each single captured 
image to realize temporal super resolution of the depth image sequences.
With our method, multiple patterns are projected onto the object with higher
fps than possible with a camera. In this case, the observed pattern varies
depending on the depth and motion of the object, so we can extract temporal
information of the scene from each single image. The decoding process is 
realized using a learning-based approach where no geometric calibration is needed.
Experiments confirm the effectiveness of our method where sequential shapes
are reconstructed from a single image. Both quantitative evaluations and comparisons
with recent techniques were also conducted.

\end{abstract}

%%%%%%%%% BODY TEXT
\section{Introduction}
\vspace{-2mm}

%\knoteA{動くシーンのアクティブ3D計測は大事だよ}

\kcut{ここを修正する。
Active scanning systems for capturing dynamic scenes
have been intensively investigated in response to strong demands from
various fields, \eg, self-driving cars, surveillance, medical applications, and 
so forth~\cite{Kinect,DBLP:conf/eccv/FurukawaMSTYK16}.
}
Active depth imaging systems for capturing dynamic scenes
have been intensively investigated in response to strong demands from
various fields.
Previous work on dynamic scene-capture mainly used
light projectors which project a static structured light pattern
%, where positional information of projector image plane is embedded in %a certain area of 
%the pattern, 
onto the object. 
Recently, time of flight (TOF) sensors have improved the capability of real-time 
and high resolution capturing~\cite{Kinect2}.
%available and intensively researched. 
%
%, however, high fps is basically difficult to realize.
%
%\knoteA{でも実現は難しい}
%
However, from the wide variety of real-time 3D scanning systems available, including commercial 
products, it is still a difficult task to realize high-speed 3D scans of fast-moving 
objects. One main reason for this difficulty comes from severe limitations on the light intensity of 
pattern projectors to sufficiently expose all the pixels of the image 
sensor in a short period of time.
This is a common problem for all active scanning systems including both structured 
light and TOF sensors.
%Since the insufficiency of light intensity is also a problem for a normal camera to 
%realize high speed imaging, hardware based solutions have been researched.
% and developed for camera system. 
%
%\knoteA{今回我々はこれを解決する→プロジェクタのfps＞＞カメラのfpsを利用して、一
%枚の画像内にたくさんの情報を埋め込むアプローチ}
%
Another reason is that it is still not 
common for imaging sensors to be capable of high speed capturing due to hardware 
limitations.

\begin{figure}[htb]
%\vspace{-5mm}
\begin{center}
	\includegraphics[width=60mm]{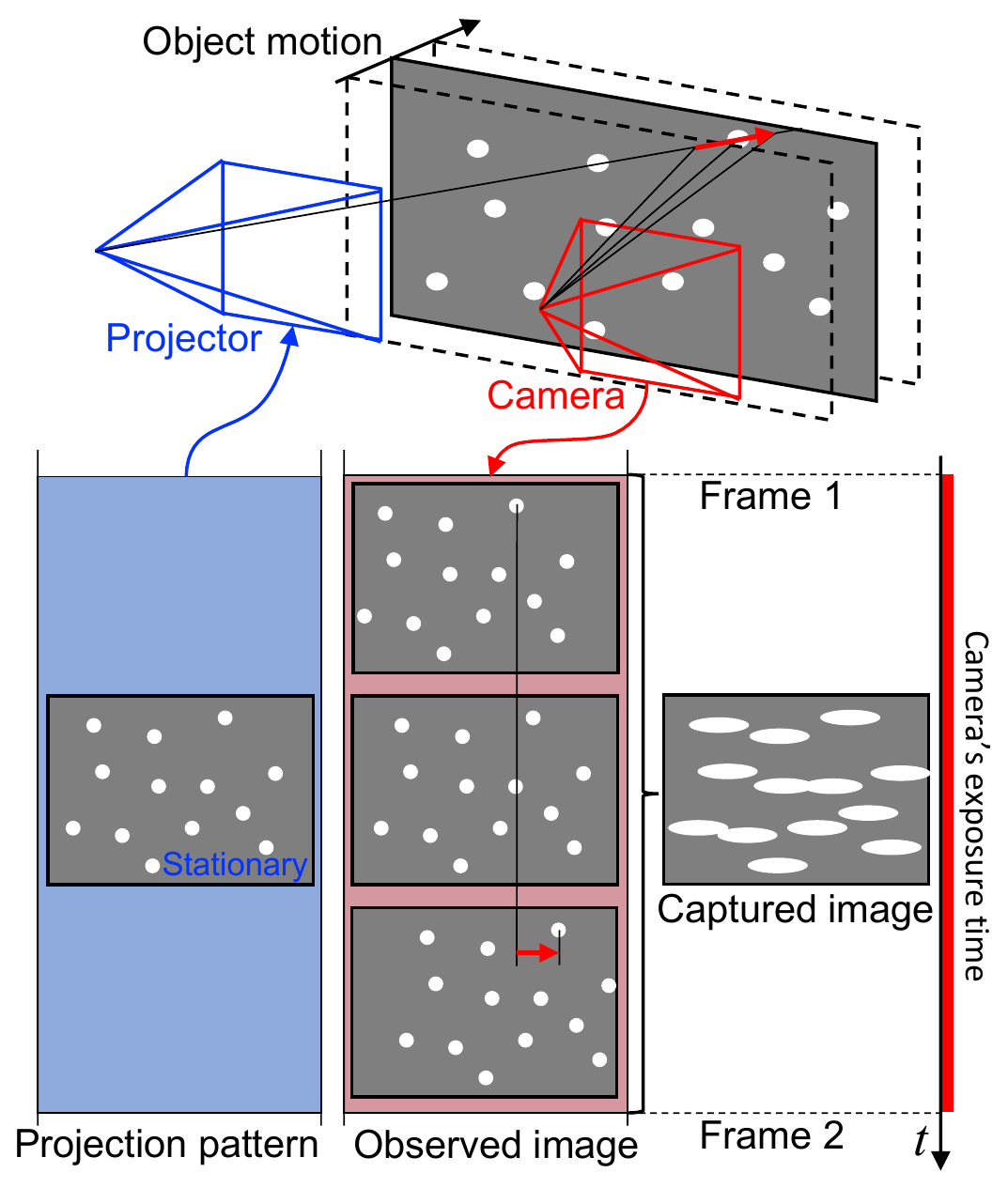}\\
 	\vspace{-2mm}
(a) Static pattern projection on fast moving object
 	\vspace{2mm}

	\includegraphics[width=60mm]{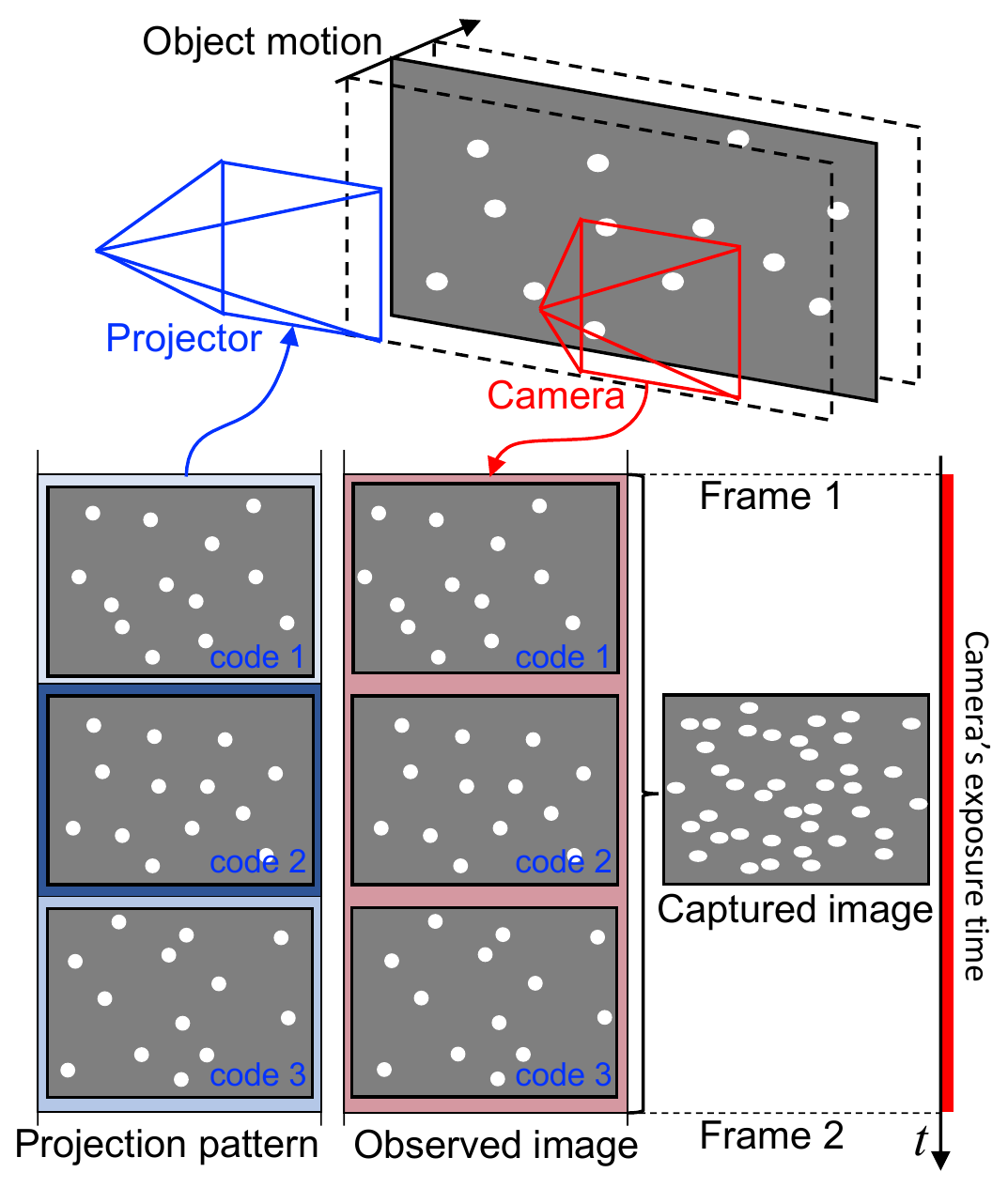}\\
 	\vspace{-2mm}
(b) Multiple pattern projection on fast moving object
 	\vspace{2mm}
	\caption{Differences between motion blur and  intra-frame motion encoding.}
	\label{fig:temporal_integration}
	\vspace*{-3mm}
\end{center}
\end{figure}

On the other hand, based on recent progress on precision devices fundamental to
 Digital Light
Processing (DLP) technology, such as  
micro-electromechanical systems (MEMS), extremely high fps (\eg 10,000fps) is 
readily available.
Therefore, in general, the switching speed of patterns on a projector is much faster 
than that of a camera. Based on this fact, 
we propose a new solution to reconstruct 3D shapes of fast moving 
objects %which moves faster than the shutter speed of camera 
by projecting multiple patterns onto 
    the object with a higher fps than that of a camera.
Since it is possible to alter patterns with extremely high fps,
the projected pattern on a moving object is not easily
    blurred out and a sharp patterns are effectively preserved (Fig.~\ref{fig:temporal_integration}(b)).
%consisting of accumlated multiple patterns  are kept.
%Since the projected 
%    pattern alters by multiple times faster than shutter speed, patterns are not 
%    blurred, but a sharp patterns, which is an accumlation of multiple patterns, 
%    can be captured.
By using these sharp patterns with our reconstruction algorithm,
% to the single captured image where multiple patterns are accumulated, 
not only a single shape, but also sequential 
shapes can be recovered; we call this temporal shape super-resolution.

%It should be noted that projector fps is usually high, at 60Hz or 
%more, in order for  degradation to not be noticeable by the
%human visual system. However, it is not 
%easy to realize high fps image capturing due to several hardware limitations: 
%image sensors require a certain exposure time; broad band width is required for data transfer;
%buffering requires large memory, writing to HDD is slow, and so forth. Rather 
%than viewing these as actual limitations,
%our technique takes full advantage of the
%mismatch between the projector and camera's frequency to embed information of object 
% motion into a captured image.
%
%Since it is still hard to capture high fps image sequence because of hardware 
%limitation, such as exposure time for image sensor, band width for data transfer, 
%size on memory, writing time for slower storage like HDD and capacity of HDD also.
The basic idea of a reconstruction algorithm is straightforward; 
we simultaneously search the depth as well as the velocity of the object's surface
in order to best describe the captured image for each pixel by synthesizing the 
image using an image 
database which is captured in advance. 
This is a multi-dimensional search of all pixels and the calculation time 
becomes enormous if brute-force search is applied.  
To reduce the computational time, we first obtain initial depth and velocity estimation under constant velocity assumption, 
then, we refine the estimation allowing varying velocities. 
%frames for each patch to extend stereo corresponding point
%search to compensate the velocity of the 
%point of object surface. 
%To increase accuracy, we adopt database driven approach in
%To achieve this, we coded not only the depth, but also 
%the coded motion blur in our system.
%Unlike the common structured light system which captures
%streaks of pattern if motion is faster than shutter speed, with our technique, 
%coded motion blur with embedded information is captured.
%
%This is a simple algorithm, however, with the best of our knowledge, such system has 
%never been proposed yet. %before.

%\knoteA{カメラでも実は似たようなものがある。それにInspireされたものである（以下
%をもっとコンパクトに）}

%\knoteA{コントリビューションは以下の通り（もっとコンパクトに）}

Several experiments were conducted using an off-the-shelf DLP projector module~\cite{LightCrafter4500} to confirm the effectiveness of our method with
quantitative and qualitative evaluation, %as well as comparison with recent successful commercial devices~\cite{Kinect}, 
proving that our system can
recover the shape of a fast moving object with better accuracy than previous techniques.
This paper proposed the following:
%The contribution of the paper is as follows:
\begin{enumerate}
%\item Motion information was embedded using high fps pattern projection as for coded motion blur.
%\item Unique motion information embedding technique with varying motion blur 
%      which depends on the velocity of the object when projected by high fps pattern
%\item Since motion blur varies depending on the velocity of the object when it
%is subjected to our high fps pattern projection, unique motion information is
%embedded in the resulting captured image.
\vspace{-2mm}
\item Multiple patterns projected into a single frame to encode both depth and 
      motion information, in order to recover 
      sequential shapes of fast moving objects from a single captured image.
%\item A high-speed projection pattern alternating technique, which 
%      can embed information of both object shapes and motions to recover 
%      sequential shapes of fast moving objects from single captured image, is proposed.
%generated by the variance in motion blur caused when the object moves at
%different velocities while subjected to high fps projection patterns.
%\item Motion information embedding technique with varying motion blur 
%whose appearance depends on the velocity of the object when projected by high fps pattern is proposed.
%Motion information was embedded as a unique appearance of motion blur 
%depending on velocity of the object using high fps pattern projection; we 
%call it coded motion blur in the paper.
%\item By analyzing the coded motion blur, motion information as well as depth 
%      information was retrieved from a single captured image
\vspace{-2mm}
\item Learning-based reconstruction algorithm to avoid geometric calibration as 
      well as efficient search algorithm to reduce calculation time. 
%      coarse to fine approach to reduce calculation time. % are proposed.
%\item Information about fast-moving objects, which cannot be reconstructed using ordinary real-time 
%      scanners was reconstructed.
\vspace{-2mm}
\item An actual system be built with an off-the-shelf DLP projector module with an 
      ordinary 
      camera to realize 1,800fps reconstruction. 
%\knoteA{
Note that no special setup, 
      synchronization nor extra devices are required.
%      }
%, to conduct comprehensive evaluations assessing
%      the effectiveness %and limitations 
%of the technique.
\end{enumerate}

\jptext{
・3次元計測は色んな分野で大切です。→特に動きのある3次元計測が重要

・Structured lightによるワンショットがある。TOFよりは、ステレオの方が精度がいい。

・しかしStructured lightによる計測の弱点として、輝度不足のために高速撮影ができないことが挙げられる

・データ転送やデータ処理などのボトルネックで普通のマシンでは、高fpsは難しい。

・一方で、プロジェクタは、DLPで使用されるDMD素子は非常に高速にパターンを切り替えられる
（しかし人の目はせいぜい30-60fpsしか見えない。もったいない）

　→つまり、カメラのfpsとプロジェクタのfpsにはミスマッチがある

・提案手法では、このミスマッチに着目し、低速撮影において、パターンを高速に切り替
えることで、動きのある物体形状の時間超解像を実現する

・具体的には、低速撮影で起きるモーションブラーを、投影パターンを高速に切り替える
ことでコード化することで、通常のステレオと同様のアプローチで、デプスと移動速度の
両方を同時に推定する。

・実験で有効性を示す。
}

In section \ref{sec:related}, we explain the related work.
Then, 
%in the section \ref{sec:theory}, we will explain the theory of temporal 
%super-resolution with high fps pattern projection.
in section \ref{sec:overview}, system configuration and overview of the 
algorithm are explained followed by a detailed explanation 
%of implementation 
in section \ref{sec:implement}.
Finally, we evaluate the accuracy of the method 
%by using the real data 
followed by limitations, 
and 
conclude the paper.

\section{Related work}%\knote{日浦先生}}
\label{sec:related}

In general, high frame rate of the input image is essential for analyzing 
rapidly moving objects. Since the sampling theory~\cite{shannon1949} determines the upper bound 
of recoverable temporal frequency for given sampling rates, specially designed 
image sensors~\cite{kleinfelder200110000,ultrahighspeed} have been intensively explored to achieve higher temporal 
resolution of input image sequences. However, we cannot ignore the fundamental 
trade-off between temporal and spatial resolution caused by the limited 
performance of analog-to-digital (AD) conversion, and by the storage capacity of the 
captured images. To overcome this trade-off, redundancy of the moving scenes has 
been exploited. Gupta et al~\cite{gupta:2009} synthesized high-resolution videos 
by combining low-resolution videos and a limited number of high-resolution 
key frames.
In order to 
efficiently preserve the motion information of the scene in a single captured image, 
a technique using a coded aperture, in which aperture shape changes faster than 
shutter speed, has been proposed~\cite{nagahara2010programmable,bub2010temporal}. Similarly, Hitomi~\cite{hitomi2011video} reconstructed a short video from a single 
coded exposure image, which sampled the scene randomly in terms of both spatial 
and temporal aspects, whereas Nagahara and Llull~\cite{nagahara2016high,llull2013coded} change exposure time for 
each pixel faster than shutter speed.
In those cases, the single input image can also be
regarded as a temporal sequence of sparsely sampled images. In other words, the 
corrected array of  pixel values is not an instantaneous image but a culled 
version of the temporal image sequence. Similar techniques that encode temporal 
information in a limited number of images have been proposed~\cite{reddy2011p2c2, park2009multiscale, sankaranarayanan2010compressive,makihara2010temporal,shechtman2005space}.
However, these methods still share the unavoidable critical issue of costly special sensors 
with controllable exposure timing specified independently for each pixel. 

Temporally coded light blinking faster than the sensor rate is also 
effective to increase the temporal information in a video with a limited frame 
rate~\cite{veeraraghavan2011coded}. In this case we can use ordinary imaging devices,
but from the viewpoint of the sampling scheme of the redundant spatio-temporal array,
it is far from optimal since all pixels are sampled simultaneously. In other words, the effect
of homogeneous blinking light has similarity to the technique of coded exposure~\cite{raskar2006coded}
so it is difficult to recover motion picture from a single input image as with Hitomi's method~\cite{hitomi2011video}
using pixel-wise individual exposure coding. Also blinking illumination is not versatile in daily-use cameras.
%, and it does not work in a stationary strong light source such as sunlight.

Contrary to the existing work listed above, our method of temporal 
super-resolution of 3D shapes is not only efficient in terms of sampling 
scheme and minimum cost for ordinary imaging devices, but is also natural in a shape-measuring context, 
because the method of projecting artificial light onto the object is not 
eccentric for active depth measurement~\cite{Sagawa:ICCV2011,Ulusoy:3DIM09}.
%\knote{←日浦先生、これであってますでしょうか？：あってます．川崎研のワンショットアクティブの論文を幾つかcite でつけて下さい} 
The proposed pattern of projected light is encoded spatially 
and temporally to maximize the exploitation of the motion information of the 
moving shape. 

% その他のコメント
% 「ワンショットスキャンは輝度不足のために光速撮影ができない」とありますが，それよりも，カメラのフレームレートを上げるとAD変換のレートが上がり取り込みが難しいという方がいいかもしれません．普通のフレームレートのカメラで録画するだけで，それよりも高いフレームレートの動きが撮れる，ということを全面に出したほうがいいのでは？

There are several papers, which 
use multiple patterns to reconstruct
a moving object~\cite{Taguchi2012,weise07:_fast_scann}, however, they capture each pattern in individual
frames and do not capture multiple exposure of patterns
into a single frame, it is difficult for them to achieve reconstruction of 
faster motion than camera fps nor temporal super-resolution.

\section{Overview}
\label{sec:overview}

\subsection{Configuration of the system}%\knote{川崎}}

\begin{figure}[tb]
%\vspace{-5mm}
%\begin{center}
\centering
	\begin{minipage}{0.64\hsize}
	\includegraphics[width=53mm]{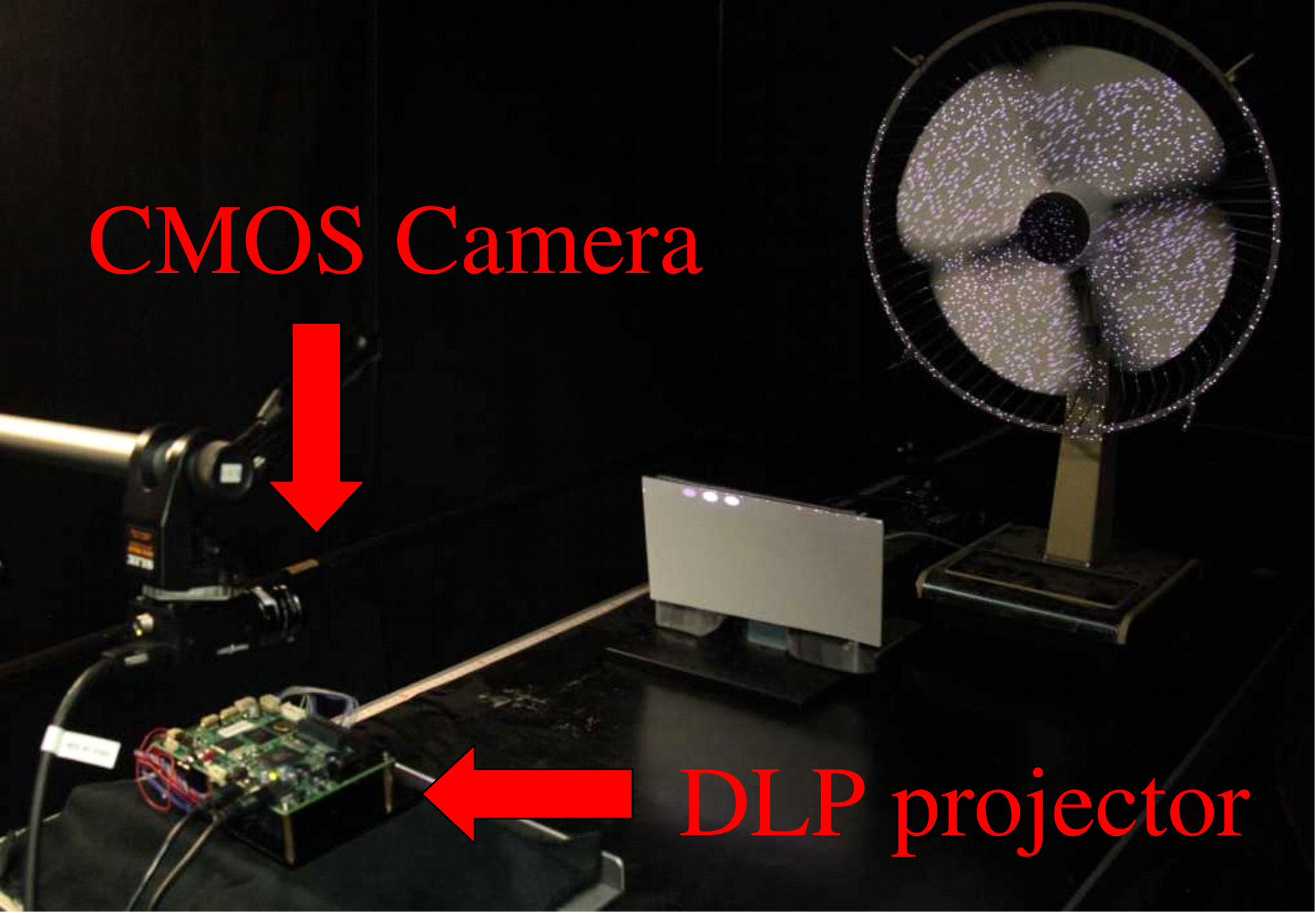}
	\end{minipage}
	\begin{minipage}{0.3\hsize}
\centering
	\includegraphics[width=30mm]{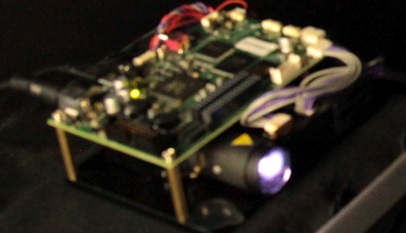}\\
Close up view of a DLP projector.
	\end{minipage}
 	\vspace{2mm}
	\caption{System configuration. Only a off-the-shelf camera and a projector are used.}% \knote{計測風景}}
	\label{fig:actualequipment}
%\end{center}
\vspace{-3mm}
\end{figure}

%\knoteA{
The system setup for our experiment is the same as for any common active stereo setup
and does not require special setup.
Nor do we require synchronization 
between the projector and the camera. 
The only difference from 
%the typical active stereo setup 
conventional systems
is that temporally coded patterns are projected onto the object.
Thus, if a fast-changing pattern can be projected, most previous structured 
light systems can be used with our method; such simplicity and generalities of the methodology are 
important considerations of our work.
Fig.~\ref{fig:actualequipment} shows an example setup using a high-fps DLP 
projector, which is used for real experiments.
%}

%The projector 
%and a camera are placed at
%%\fnote{
%a certain baseline,
%%}
%and 
%a light pattern is projected onto the object. 
%We ensure the projector and camera are calibrated so that both the
%intrinsic and extrinsic parameters of the camera and projector are known.
%Synchronization is not required with our system, because the parameter for 
%search frame is estimated, as explained in the following section. (note that this feature greatly 
%simplifies the system configuration and widens the applicable area of the system.)

%realize temporal super-resolution of 3D shapes.

\jptext{
●セットアップについて

プロジェクタ1台＋カメラ1台のシステム

ただし、プロジェクタは高速にパターン切り替え（一般にプロジェクタの方が早い）
}

\subsection{Basic idea}

With our technique, motion information is embedded into the accumulation of multiple patterns. Then a
series of shapes are temporally super-resolved by extracting the individual 
pattern from the pattern set.
The basic theory and overview of the technique
is explained in Fig.~\ref{fig:temporal_integration}.
%As shown in 
%Fig.~\ref{fig:temporal_integration}(a), 
If there is no motion in the scene, 
a sharp pattern is observed by the camera.
If there is motion in the scene, the position of the projected pattern on the 
object surface will move depending on the depth and velocity of the object,
and thus, the observed image subsequently degrades as shown in Fig.~\ref{fig:temporal_integration}(a).
%show different artifacts, 
In this process, if the scene motion is fast, motion blur increases
because of a large position change of the pattern on the object surface, and less motion blur 
will be observed if the motion is slow.
To overcome the problem, we project multiple patterns with a higher fps than 
that of the camera, as shown in 
Fig.~\ref{fig:temporal_integration}(b). With the method, the projected pattern is rapidly 
switched to the next pattern, thus several patterns are integrated on the object surface
and captured by image sensors in one frame.
Unlike the static pattern projection where pattern is blurred, 
high frequency patterns, which consist of multiple patterns with different codes, 
are captured without blur. Since the integrated pattern varies depending on which patterns are 
used, how fast the object moves and
 how deep the object exists, 
%the range of depth in the object,
it is necessary 
to estimate those parameters simultaneously. 
%\knoteA{
Also, projected patterns are significantly affected by defocus and aberrations.
%}
%For solution, we take a simple approach as follows. 
We solve those issues by taking a simple approach as follows:
First, we create a pattern database which stores independent patterns 
projected on a flat white board set at varying depth. 
%Then, images are synthesized the image 
%with a certain set of parameters to find the values which maximize the similarity 
%between the captured image and the synthesized one.
Once the database has been created, it can be used as the basis for comparison 
with the image capturing the moving target object. Thus, the image of the moving 
target object can be measured for depth and velocity by seeking the 
corresponding pattern combinations from a vast database.

%
%Then, instead of conducting a two step approach such as removing blur from 
%the captured image and then estimating the velocity of the object from motion 
%blur, we simultaneously estimate initial shape and motion information
%by conducting a multidimensional search to maximize the similarity 
%between the captured image and the synthesized image using the parameters of
%depth and velocity for each pattern; 
%note that 
%thanks to temporally coded patterns with high frequency, the process becomes more 
%robust and precise than with a static pattern, and this has been confirmed by our experiment.

\jptext{
原理をここで書く

１．プロカムが前提

２．早いと投影パターンはボケます。

３．ボケには動き情報あり

４．動き情報が取り出せて、初期形状があれば、それでスパレゾできる。

５．しかしボケからそれを取り出すのは難しい

６．しかも初期形状もボケてて作れない。

５．両方を同時に解決する手法を提案。

６．具体的には投影パターンをコード化することで実現される

先日のスライドを使う
}

\subsection{Algorithm overview}%\knote{川崎}}
\label{subsec:algorithm}

\begin{figure}[tb]
\vspace{-2mm}
\begin{center}
	\includegraphics[width=80mm]{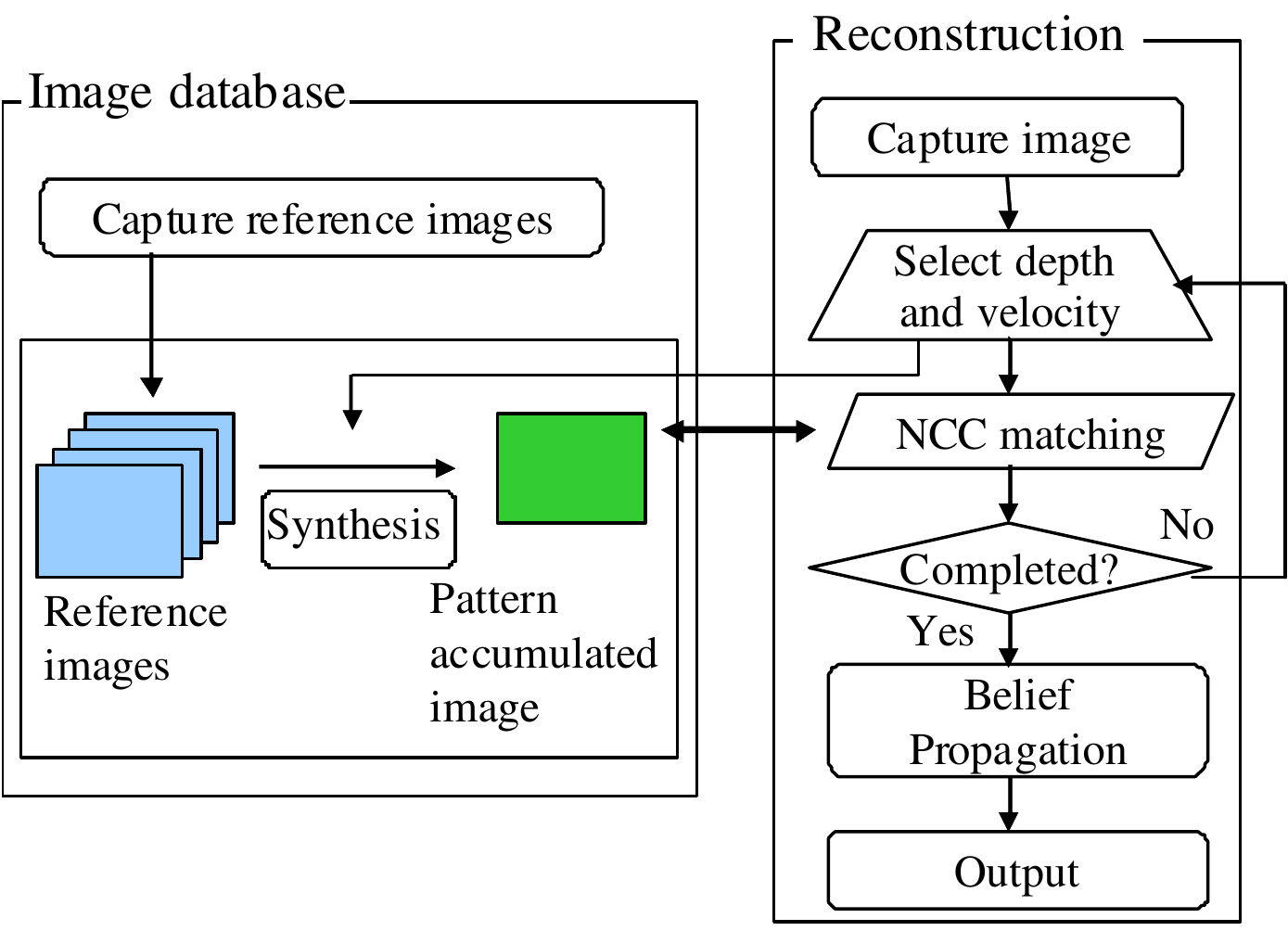}\\
	\caption{Algorithm overview}
	\label{fig:algorithm_overview}
\vspace{-7mm}
\end{center}
\end{figure}

Fig.\ref{fig:algorithm_overview} shows the diagram of the algorithm.
The technique mainly consists of two parts: image database creation and 
shape reconstruction. %four steps:
%In our implementation, 
Note that image database creation is an offline process performed only once.
To this end, reference images are captured by changing the depth 
of a planar board with a known position on which the set of randomly distributed 
dense dot patterns 
are projected. % using coded aperture
% for stereo matching. 
This process is repeated for all the independent patterns respectively.
%\knoteA{
Since such a database approach can solve open 
problems such as strong 
defocus blur, etc. for a projector-camera system, its efficacy is currently being intensively researched~\cite{kawasaki2015active,fanello16:_hyper,karsch2012depth}.
%The main strength of such a database approach is that it can solve the open 
%problems for a projector-camera system, such as strong 
%defocus blur, aberrations, etc.
%Because of the advantage, a database approach for 
%structured light systems is currently being intensively researched~\cite{kawasaki2015active,fanello16:_hyper,karsch2012depth}.
%}

%The reason why such a database approach is adopted is that it is a simple way to 
%avoid complicated issues which exist in a projector system, such as strong 
%defocus blur, aberrations, etc. 
%%of projector, difficulty on calibration, etc. 
%Further, a database approach for 
%structured systems is currently being intensively researched and several advantages for
%robustness and accuracy have been reported~\cite{kawasaki2015active,fanello16:_hyper,karsch2012depth}.

%~\cite{Fanello_2016_CVPR,kawasaki2015active,fanello16:_hyper,karsch2012depth}.
%explained in the Sec.~\ref{sec:discussion_distrt}. 
%Since pattern integration process varies depending on depth and motion of a object, stereo matching 
%will fail %if matching window size is large and the angle difference is large.
%unless those information is known.
%the velocity of object as well pattern changing information is known.
%We solve the problem by capturing all the pattern independently and synthesizing 
%the integrated pattern using the velocity of the object as explained in Sec. \ref{sec:Motion_compensation}.
%We therefore synthesize new reference images of slanted planes from captured image set with various orientations.
%(Sec. \ref{sec:slanted_plane}).

%Next, 
%to recover the shape, 
During the shape reconstruction phase,
we captured the target object by alternating projection patterns faster than 
camera fps.
% and
%%specially designed pattern with the light field projector using coded aperture. % is installed.
%by conducting stereo matching between the captured image and the 
%image database. % to recover the shape. %find correspondences.
%At the matching process, 
%\knoteA{
Then, the velocity and the shape of the object is recovered through two steps; the first is an initial depth estimation 
assuming a constant velocity for each pixel, and the second  
is a re-optimization step estimating the independent velocity 
%using the initial depth 
according to a non-linear optimization method;
note that errors 
accumulated by an assumption of constant velocity are efficiently decreased by 
the next re-optimization step.
Both steps conduct stereo matching between the captured image and the 
image database, for estimation. Reference image patches are synthesized by using values of velocities and depths of the 
object, and the patches of the captured image are compared with the reference patches
using normalized cross correlation (NCC).
In our experiments, 16x16 and 24x24 pixels windows are used for calculation.
%The detailed matching process is
%described in Sec.~\ref{sec:depth_velo_est}.
As for the pattern design, we adopted a standard dense random 
dot pattern used by real-time scanning 
systems~\cite{Kinect}. %,realsense:sr300,artec:pat07}.

\section{Implementation}
\label{sec:implement}
%詳細を説明する

\subsection{Motion compensation database construction}%\knote{芝（小野先生）}}

In our method, learning-based approach is adopted instead of geometric 
calibration. As mentioned in Sec.\ref{subsec:algorithm}, 
%The main reason is to avoid lens distortion and aberration 
%derived from complicated optics of projector. Although the database creation
%process is arduous, recent studies on learning based
%active 3D scan has proven to show better performance~\cite{Fanello_2016_CVPR} in 
%some cased. 
%And thus, 
the huge data size for learning-based approach is not just simply a weakness, 
but should be considered with a trade-off between accuracy, and we take accuracy 
in our method.

%●パターンについて
%
%ランダムドットを10枚

\jptext{
提案手法において投影されるパターンは，互いに異なるランダムドットパター
ン10枚である．
%これを，カメラの露光時間が1周期となるように投影を行う．
本論文では，カメラの露光時間を300ミリ秒から1秒程度とし，露光時間内に投
影パターンが1・7枚程度写るように投影パターンの切り替えを行う．
%間隔で切り替え，1秒周期で投影する．
ランダムドットパターンを投影する理由は，微小な窓領域内において，奥行き
および速度に対してなるべくユニークなパターンとするため，および，ブラー
がより明確に見えるためである．

-------------------
（芝）
ランダムドットパターン10種類。
パターンの切り替えの秒数はシャッタースピード300msに何フレーム分収まるかで測定したため
パターンの切り替え速度は実験によってことなります。
超解像での実験では、より加速度がわかりやすくするために露光時間を1秒(予定)で撮影。
プロジェクタのPCからプロジェクタに信号を送る際、信号をプロジェクタから投影する際のいくつかの遅延が重なるため
変化しない領域に常にパターンごとにIDのような模様をつけて投影している。
}

%●データベース方式
%
%その方が、安定なので
%撮り方などについて説明する（10枚別々に1mmずつとか）

\jptext{復元を安定的に行うため，synthesized imageからなるdatabaseを構築する．
構築した装置を用いてパターンを投影しつつ，動く平面板（reference plane）
を撮影する．このとき，1mm単位で奥行きを変えつつ，10パターンを個別に撮
影する（ブラーは含まれない）． 次節で後述するように，速度に応じたパター
ンを事前に合成するのではなく，オンザフライで合成する．}

The image database is created by capturing the actual scene where a planar reference board on a
motorized stage is moved from $d_{min}$ to $d_{max}$ between the
projector and the camera.
% In-focus distance is ●● mm \pm ●● mm for the projector.
Each pattern is projected on the board and captured
each time  the board is moved one predefined unit length (0.5 mm in our 
experiment for sufficient accuracy in relation to data size), thus the captured
static reference images do not involve motion blur.
Because of the baseline between the projector and the camera, 
the observed pattern is shifted with disparities depending on the 
distance to the board. 
This capturing process is repeated $N_{p_{max}}$ times. 
%The reference image patches dependent on
%depths and velocities are
%created on-the-fly from the image database as described in
%Sec.~\ref{ssec:implement_estimation}.

As for the projection pattern,
% employed using the proposed technique is time-variant and
%consists of multiple dense random dot patterns.
we use multiple dense random dot patterns %for our system;
%Camera's exposure time is set to 300 [ms] to 1 [s], and
%the reason why the random dot pattern is used is 
to make the
projected pattern as unique as possible and to make blur clearly
visible in all directions.
%The patterns change so that $N_{p_{max}}$ patterns are captured within one
%shot. 

%\subsection{Simultaneous depth and velocity estimation}%\knote{芝（小野先生）}}
\subsection{Initial depth and velocity estimation assuming constant velocity}%\knote{芝（小野先生）}}
\label{ssec:implement_estimation}

%
%\begin{figure}[tb]
%%\vspace{-5mm}
%\centering
%%	\includegraphics[width=0.48\textwidth]{arxiv_figs/makingRefImg_coarse.png}   
%        \includegraphics[width=0.48\textwidth]{arxiv_figs/makingRefImg_coarse-eps-converted-to.pdf}
%        \caption{Synthesize of velocity-sensitive reference image
%	used for coarse matching.
%%  \knote{パターンの図。それらを重ね合せた図も}
%    } \label{fig:makingCoarseRefPattern}
%%\end{figure}
%~\\
%%\begin{figure}[tb]
%%\vspace{-5mm}
%%	\includegraphics[width=0.48\textwidth]{arxiv_figs/makingRefImg_fine.png}
%       	\includegraphics[width=0.48\textwidth]{arxiv_figs/makingRefImg_fine-eps-converted-to.pdf}         
%%        \includegraphics[width=0.48\textwidth]{arxiv_figs/makingRefImg-eps-converted-to.pdf}
%        \caption{Synthesize of velocity-sensitive reference image
%	used for fine matching.
%%  \knote{速度によって、パターンが変わる様子の図を載せる}
%    }
%\label{fig:makingFineRefPattern}
%\end{figure}
%
%\if 0
\begin{figure}[tb]
\vspace{-2mm}
\centering
        \includegraphics[width=0.48\textwidth]{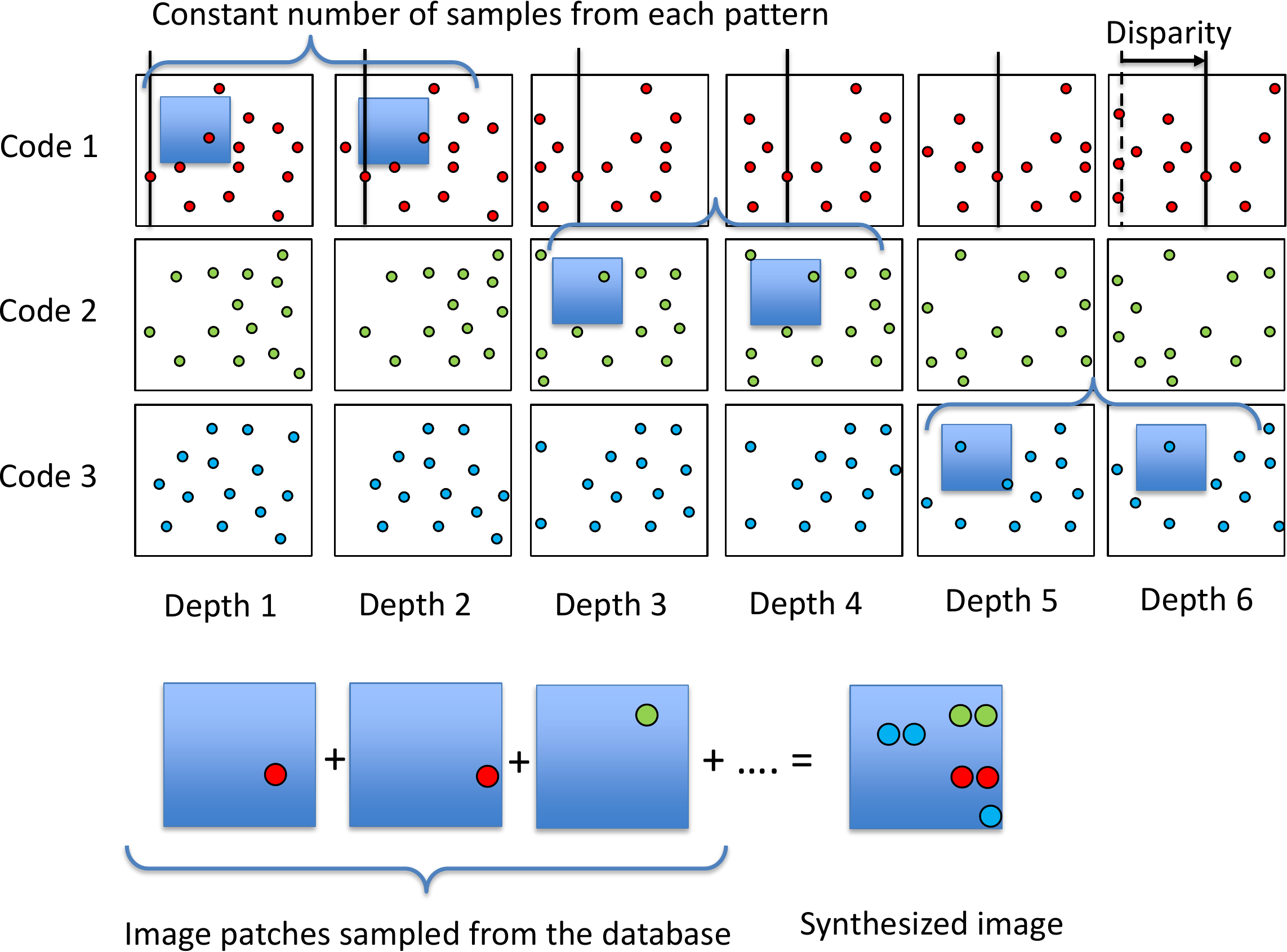}
        \caption{Synthesis of reference image patches
	for constant velocities.
%  \knote{パターンの図。それらを重ね合せた図も}
    } \label{fig:makingCoarseRefPattern}
\vspace{-2mm}
\end{figure}
%\fi

%●2次元Matchingベース

\jptext{
前節で撮影したreference image databaseをもとにdepthとvelocityの
multidimensional searchを行う． 速度に応じたreference imageを事前に作
成すると，得られる画像の総数は，depthの解像度分（100mm・400mmの301段
階）×速度の解像度分（from 1 pattern per exposure to 10 or 30
patterns = 10～30枚）（1秒間に動く速度を現在仮定して推定している（全探
索（Coarse）で大まかにするのであれば現在最大30枚：超解像実験ではこのく
らい行う予定))で，トータル（実験によって変わるが）約9000（今の環境では
3000枚）枚もの膨大な枚数が必要となる．さらに，撮影されたフレームが異な
る場合も想定すると，その枚数は膨大になる．このため，オンザフライで
matching patternを合成する．}

The proposed technique simultaneously estimates depth and velocity
using a multidimensional search with the reference image database.
The reference images that can be observed depends on depths 
and velocity. 
However, generating reference image patches in advance requires
huge amounts of memory and computational time for matching because the
combination of depth ($|d_{max}-d_{min}|$ levels) and velocity ($v_{max}$ levels) creates
enormous amounts of data, even if constant velocity is assumed. 
\kcut{
In reality, 
the observable images can be approximately synthesized by sampling image patches from the 
depth-dependent image database. 
%we estimate both velocities for each pattern 
%and the pattern set used is actually projected onto the object and integrated in 
%one frame from possible combinations of $_{N_{p{num}}}C_{N_p}$ sets.
Therefore, reference image patches are
generated on-the-fly, depending on a captured pattern set that is generated
using marker patterns as described in
Sec.~\ref{ssec:synchronization}.
}
%●速度を考慮したマッチングパターンの生成方法
%→いろんな速度を想定して、積分によりパターンを作る
%予め作ると膨大になるので、オンザフライ方式

\jptext{
次節で後述する方法により，撮影されたフレームを知ることができる．この情
報を用いて，matching patternを合成する．様々な奥行きおよび速度を想定し
て撮影した画像の積分を行い，動きがある対象物体を撮影した際に得られる画
像を合成する．奥行きは301段階，速度は0mm/sから100mm/sまで101段階で切り
替えて画像の合成を行う．(グラフ作成時)}

%Considering various depth and velocity for pixels, 
To reduce the computational cost, 
we first obtain coarse initial solution by assuming constant velocity of the surface.
The initial solution is refined later. 
%a coarse-to-fine approach is adopted in this method. 
%Here, coarse search means that we assume constant velocity of the surface, 
%and thus, we can avoid complexity of the combinations with varying velocities.  
\kcut{
In the initial search, 
%Initially, coarse 
depth and constant velocity for 
each pixel are estimated using %synthesized 
matching reference image patches created from the reference image database
and the patches from the captured image.  
}
The velocity is changed from $-v_{max}$ 
to $v_{max}$ %at the intervals of 1 mm/s, 
and the depth is changed from
$d_{min}$ to $d_{max}$. % at the intervals of 1 mm. 
In total, $2 \times v_{max} \times |d_{max}-d_{min}|$
the reference image patches are generated for the captured
pattern set. 
%\onote{【前の段落と数字や単位が合わない？ ＞ 芝君】}

%●NCCを使用
%
%captured imageとmatching patternとの照合はNCCにより行う． 明るさの差を
%キャンセルして照合を行える（4.2で記述済み？）．
%
%●Coarse to fineで、非線形な推定する→これがTemporal super-resoの根拠

\jptext{
depthとvelocityの推定は，coarse to fine approachで行う．coarseの段階で
は線形なmotionを想定してdepthおよびvelocityを荒く推定し，fineの段階で
非線形的な動きを推定する． すなわち，coarseの段階では，計測対象範囲の
全てのdepthおよび速度を候補としてmatching patternを生成し，推定を行う．
また，fineの段階では，pattern間毎に独立してdepthおよび速度の推定を行う．
このとき，隣接するpattern間ではdepthおよび速度が大きく変化しないことを
想定し，変化の上限を与える（★具体的な値は？？）．
}

In the initial estimation process, 
reference image patches
 $I^{const}(v,d_0)$
of depth at the start point $d_0$ and constant velocity $v$ are
generated by 
%convolving each static reference image with
%the same linear blur kernel and integrating the convolved
%images 
collecting image patches from the image database
and integrating them 
as shown in Fig.~\ref{fig:makingCoarseRefPattern}.
Note that, since constant velocity is assumed, 
the same number of patches are sampled from each pattern code.
We calculate
%on-the-fly 
%reference image 
%with constant 
%velocity assumption 
$I^{const}(v,d_0)$
by:
\begin{equation}
I^{const}(v,d_0) = \frac{1}{N_p}\frac{T_E}{T_E^{ref}}
            \sum_{n=1}^{N_p}
               \sum_{d = d_0 + (n-1)\Delta d}^{d_0 + n \Delta d}  
                  \frac{I^{static}_{p(n+N_{start}), d}}{\Delta d + 1}
\end{equation}
%\onote{【確認をお願いします ＞ 芝君】}
%
where $T_E$ and $T_E^{ref}$ are exposure time at measurement and
database construction, respectively.  $p(n)$ is $n$-th observed
pattern, $I^{static}_{p(n), d} $ is static reference image of pattern
$p(n)$ at depth $d$ in the image database, and $\Delta d$ is the moving distance while the
pattern is projected, \ie, $\Delta d = v \frac{T_E}{N_p}$.
%, considering all depth and velocity as candidates,
Then, all the generated reference images are compared to the captured
image, and initial estimated values of depth and velocities are computed by 
\begin{equation}
d, v = \argmax_{d, v} NCC\left(I^{const}(v, d), I^{cap}\right)
\end{equation}
where $I^{cap}$ is a captured image.

\if 0
\begin{figure*}[t]
%\vspace{-5mm}
	\begin{minipage}{0.48\hsize}
\centering
        \includegraphics[width=1.0\textwidth]{arxiv_figs/synthimage1-crop.pdf}
        \caption{Synthesis of reference image patches
	for constant velocities.
%  \knote{パターンの図。それらを重ね合せた図も}
        }
        \label{fig:makingCoarseRefPattern}
	\end{minipage}
%\end{figure}
%
%\begin{figure}[tb]
%\vspace{-5mm}
	\begin{minipage}{0.48\hsize}
       	\includegraphics[width=1.0\textwidth]{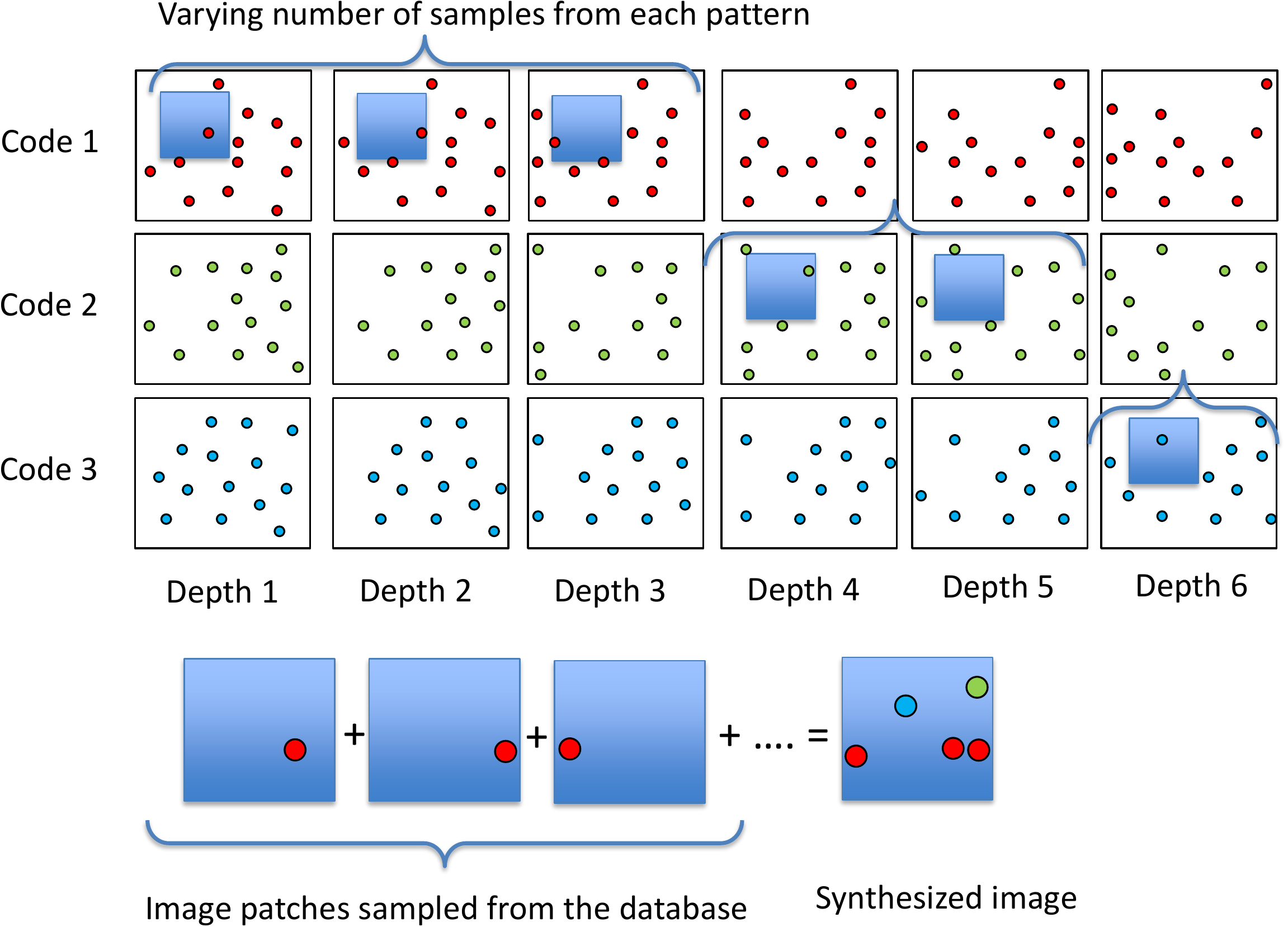}         
        \caption{Synthesize of reference image patches
	for varying velocities.
%  \knote{速度によって、パターンが変わる様子の図を載せる}
        }
        \label{fig:makingFineRefPattern}
	\end{minipage}
\end{figure*}
\fi

\subsection{Refinement of depth and velocity estimation}%\knote{芝（小野先生）}}
%\if 0
\begin{figure}[tb]
\vspace{-2mm}
       	\includegraphics[width=0.48\textwidth]{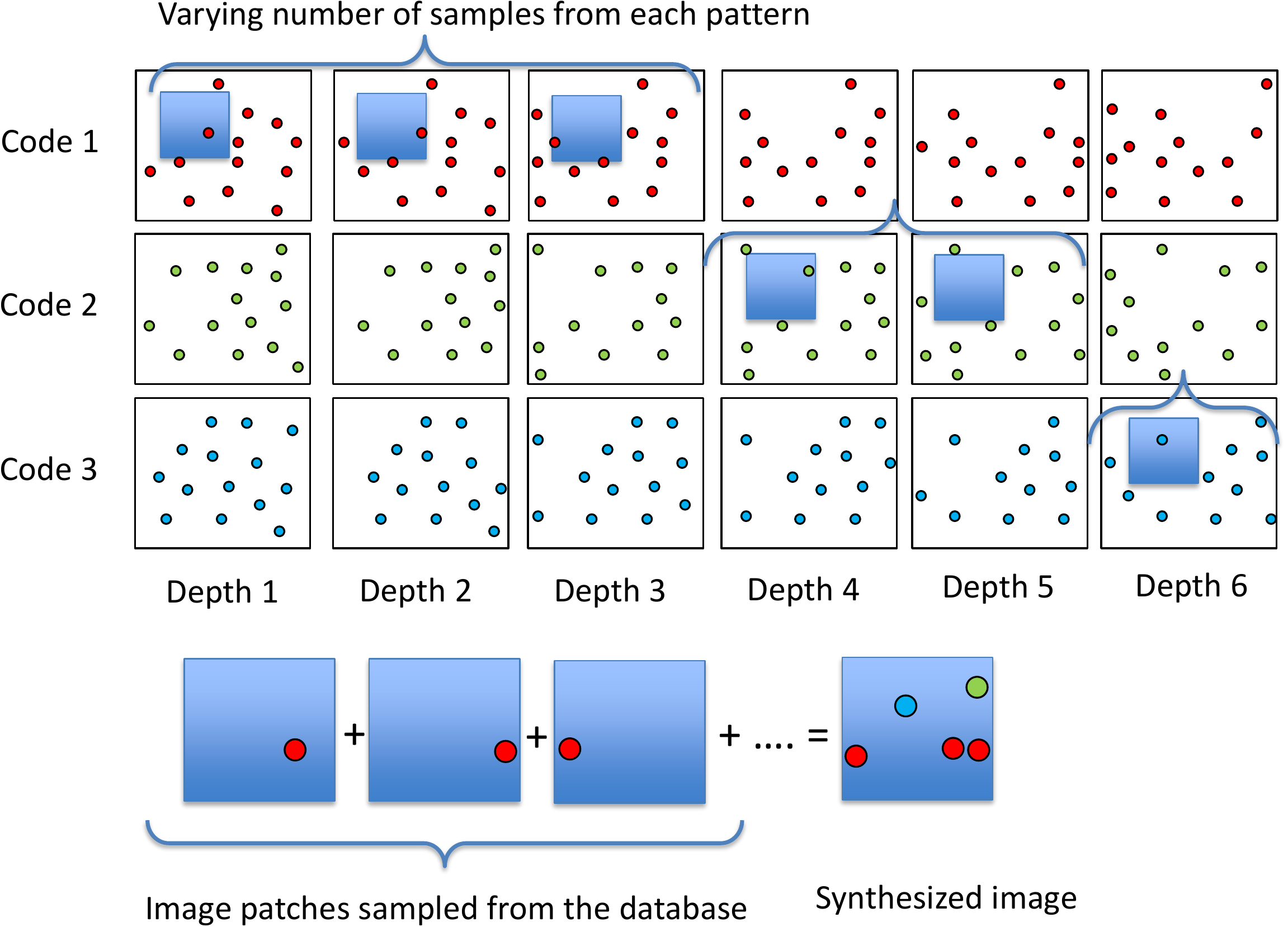}         
        \caption{Synthesize of reference image patches
	for varying velocities.
%  \knote{速度によって、パターンが変わる様子の図を載せる}
    }
\label{fig:makingFineRefPattern}
\vspace{-2mm}
\end{figure}
%\fi

%Then, fine parameter estimation of
% depth, and for each velocity for each pattern, is employed. 
% 
%At the coarse level, a
%uniform linear motion during exposure is assumed, and the depth at the
%start point of exposure and the consistent velocity during the exposure
%are estimated, whereas in the fine process, nonuniform motion of the object is estimated.

After obtaining the initial solution, 
the depth and velocity estimation is further refined. 
In the refinement process, nonuniform motion of the object is estimated.
Velocities for each interval between the projected
patterns
%$\boldsymbol{v} = (v_{p_1,p_2}, v_{p_2,p_3}, \ldots, v_{p_{(N_p-1)},p_{N_p}})$
$\boldsymbol{v} = (v_{p_1}, v_{p_2}, \ldots, v_{p_{N_p}})$
are assumed to be inconstant and their
values are simultaneously estimated.  Reference images for refinement
process are generated in the same way as the coarse process with the
only difference on time-variant velocity as shown in 
Fig.~\ref{fig:makingFineRefPattern}, 
except that the number of samples from each code patterns may not be constant.  
Then, the
reference image patch $I^{fine}(\boldsymbol{v},d_0)$ at initial depth
$d_0$ and velocities $\boldsymbol{v}$ are calculated as follows:
\begin{equation}
I^{fine}(\boldsymbol{v},d_0) = \frac{1}{N_p}\frac{T_E}{T_E^{ref}}
            \sum_{n=1}^{N_p}
               \sum_{d = d_{n-1}}^{d_n}
                  \frac{I^{static}_{p(n+N_{start}), d}}{d_n - d_{n-1} + 1}
\end{equation}
where
%$d_n=\sum_1^n v_{p_n,p_{n+1}} * \Delta t$
$d_n=\sum_1^n v_{p_n} * \Delta t$.
Then, the refined depth and velocity of each pattern are determined as follows:
%●● mm/s.
%
\begin{equation}
%v_{p_1,p_2}, v_{p_2,p_3}, \ldots, v_{p_{(N_p-1)},p_{N_p}}
d, {\boldsymbol{v}}
=
%\argmax_{v_{p_1,p_2}, v_{p_2,p_3}, \ldots, v_{p_{(N_p-1)},p_{N_p}}}
\argmax_{d, \boldsymbol{v}}
    NCC
        \left(
%           \frac{1}{N_p-1}\sum_{n=1}^{N_p-1}
            I^{fine}(\boldsymbol{v}, d), I^{cap}
        \right).
\end{equation}
To reduce the combination number of velocity values, a constraint
between velocities at adjacent intervals is imposed so that the
difference between these values should be less than threshold
(\ie, continuous velocities). 

%\onote{【速度の推定はパターンごと？パターン間ごと？ 速度が分かればdepthの推定は不要？】}
%

\jptext{
fineステップ奥行きのサーチは前後1フレーム(3段階)速度のサーチは前後1フ
レームを見て変化量は1フレーム間に増減5(積分する量(単位がわからないで
す))まで\onote{【←よくわかりません．．．口頭で教えて下さい】}
}

%●その後でBP使用

\jptext{
最後に，得られた復元結果に対してBPを適用し，復元された形状の品質改善を
図る．BPは得られたフレーム毎に独立して行うものとし，NCCにより得られた
コストボリュームを利用する．
}

Finally, belief
propagation (BP)~\cite{Felzenszwalb:IJCV06} is applied to refine the initial shape.
%the cost volume obtained by NCC.
%\knoteA{
The cost volume obtained by NCC is used for data terms and the absolute 
value of depth difference between adjacent pixels is used for regularization terms.
%}

\subsection{Passive synchronization using markers}%\knote{芝（小野先生）}}
\label{ssec:synchronization}

\begin{figure}[tb]
%\vspace{-5mm}
\begin{center}
	\includegraphics[width=0.48\textwidth]{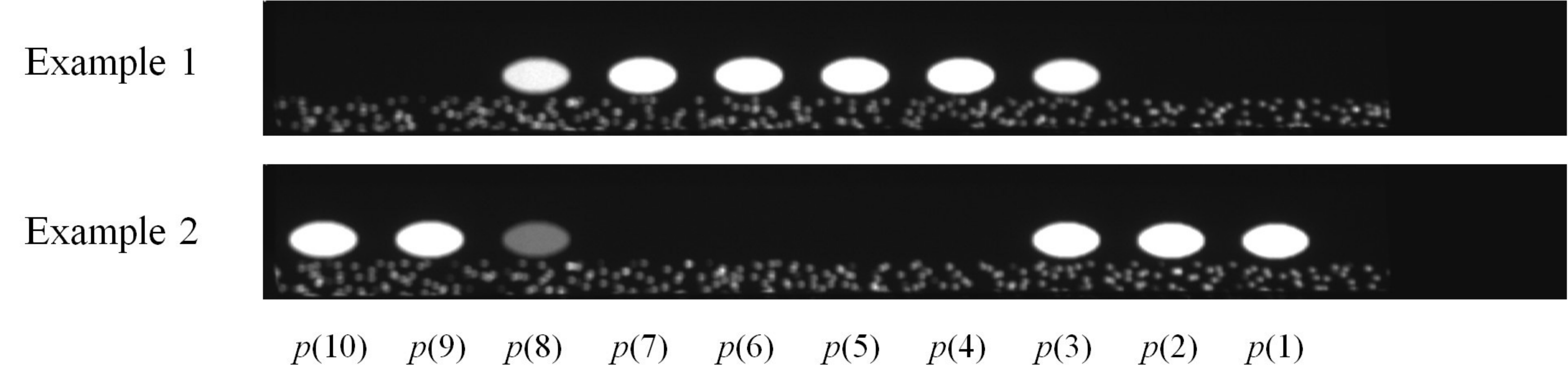}\\
        \caption{Captured image examples of projected pattern
        markers, \eg examples 1 and 2 involve $p(3), p(4), \ldots,
        p(8)$ and $p(8), p(9), \ldots, p(3)$, respectively. In example
        2, $p(8)$ is captured with about 25\%
        % and $p(1)$ are captured with about 25\% and 75\%
        of $\frac{T_E}{T_E^{ref}}$, respectively.}
%        \knote{マーカーの図をのせる}}
	\label{fig:markers}
\end{center}
\end{figure}

%外枠にマーカーをつけた同期
%
%オンザフライ方式なので、これがうまくいく

\jptext{投影パターンの下段に復元に用いない領域を設け，投影されているフレームの
IDに対応したマーカを投影する（図を参照）． これは，ビデオプロジェクタ
において遅延が生じることに対応するためであり，captureされたframeをほぼ
正確に知ることができる．また，プロジェクタカメラ間の同期が不要となり，
露光時間を変更する際に投影パターン側の調整が不要となるなどの利点も生じ
る． 上記マーカはcaptureされたframeを示すだけでなく，各マーカの輝度を
見ることで各frameが露光された時間も知ることができる．}

%\knoteA{
Although the aforementioned steps work effectively without synchronization, the search space 
is vast. If the system is synchronized, the search space can be greatly reduced. We propose a simple method to achieve 
synchronization without any hardware modifications.
%}
%To efficiently know which projected patterns are integrated into the
%captured frame, the proposed technique 
We impose markers corresponding
to each projected pattern in the peripheral region of the projected
pattern, which are not used for measurement.  The markers are captured
as shown in Fig.~\ref{fig:markers} and the intensity of each marker is
calculated in order to accurately estimate which and for how long each
pattern is captured.  
This allows users to use existing structured light systems without special re-configuration. 
%This allows us to eliminate the need for
%synchronization between the camera and the projector, even when a
%delay or a pattern skip occur with the projector due to a hardware or
%software problem, which may occur with consumer products.

\subsection{Temporal super-resolved shape reconstruction}%\knote{芝}} % with ambiguity elimination}

By using the output data from the fine level, the proposed technique 
produces $ N_p$ shapes from a single image.
Since the shape of the first frame and a velocity of each pixel at each frame is already estimated, a
temporal super-resolved shape at $n$-th pattern can be reconstructed by  accumulating 
$\Delta d$ from $1$ to $n-1$ for all the pixels. % for time $t + \Delta t * i.$
To reduce accumulation
error, we conduct a re-optimization step with reverse 
direction and average the depths from both directions.

%It should be noted that the temporal direction of a series of super-resolved shapes 
%cannot be determined. Two solutions to this can be considered as follows:
%\begin{itemize}
%\item If there are consecutive frames, ambiguities can be eliminated by considering
%      consistency with adjacent frames.
%\item If there is no consecutive frame, spatial consistency can be used to 
%      determine the direction, \ie, nearby pixels should have the same direction.
%\end{itemize}
%In this paper, we take the second approach for the sake of simplicity.

\jptext{
時間超解像の形状復元

動きの向きに曖昧さがある。

●前後のフレームが有る場合
→前後の時間的整合性から一意に決まる

●前後のフレームが無い場合
→周辺の空間的整合性から決められる
}

\section{Experiment}%\knote{芝（図と日本語）英語化（川崎＆小野先生）}}

\subsection{Evaluation with planar board}

\begin{figure}[tb]
%\vspace{-5mm}
\begin{center}
	\includegraphics[width=80mm]{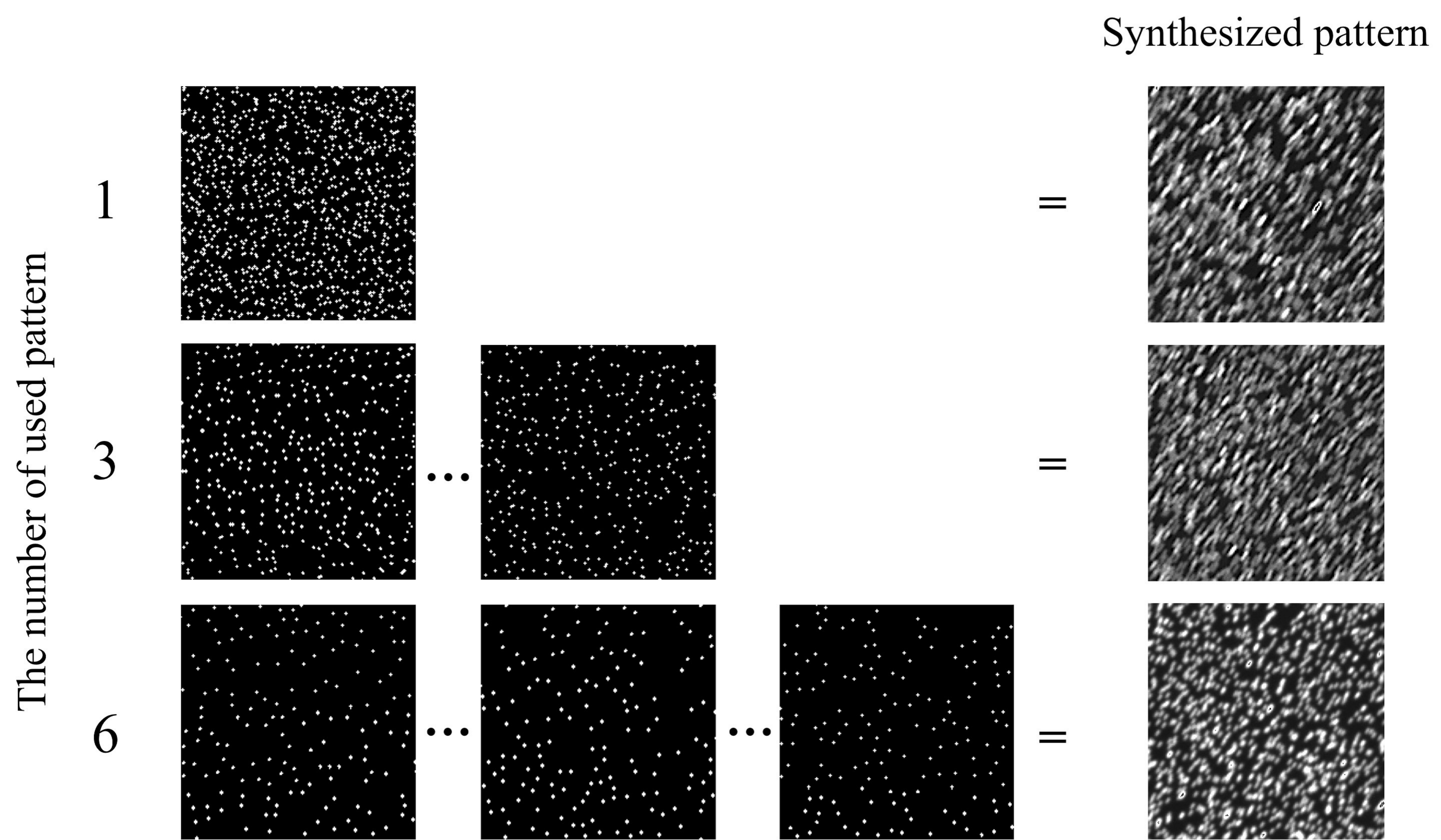}\\
% 	\vspace{10cm}
	\caption{Synthesized patterns with a different number of projected 
    patterns used. Since we assume that the object is 
    moving, motion blurs are observed on the synthesized images; blur is 
    stronger in the single pattern case than the others.}
%    Note that although the density of each 
%    pattern is different, the integrated patterns have the same density.}
	\label{fig:Image_pat_density_equalize}
\end{center}
\end{figure}

\begin{figure}[tb]
\vspace{-5mm}
\begin{center}
	\includegraphics[width=75mm]{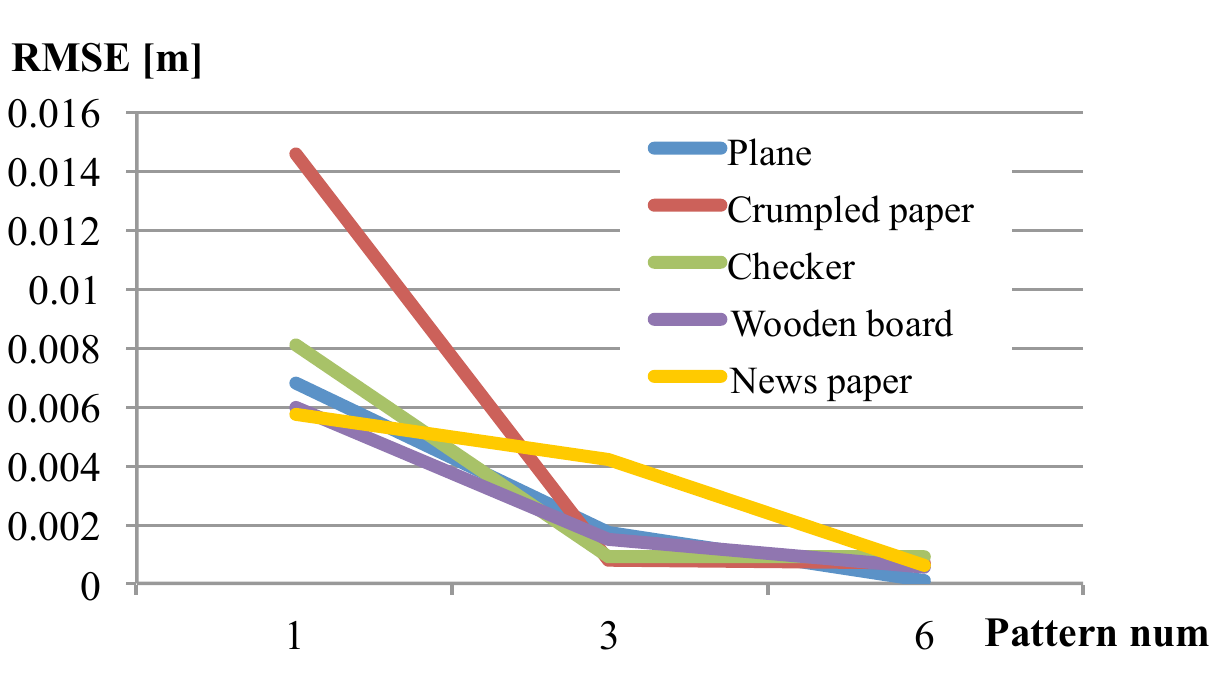}\\
	\caption{Accuracy evaluation by reconstructing a planar board with different texture 
using a different number of 
    projected patterns. RMSEs of reconstructed points from the fitted planes are
shown.
	Horizontal axis represents the number of pattern. It is clearly shown 
    that increased number realize better RMSE with all the textures.}
%	\caption{\knote{グラフ１。横軸枚数、縦軸RMSE。板、テクスチャ、しわくちゃ紙など}}
	\label{fig:Graph_pat_num}
\end{center}
%\end{figure}
%
%\begin{figure}[tb]
		\centering
		\begin{tabular}{llcccc}
		
\vspace{1mm}			
&
\rotatebox{90}{Input}&
	\begin{minipage}{0.15\hsize}
			\includegraphics[clip, width=1.6cm]{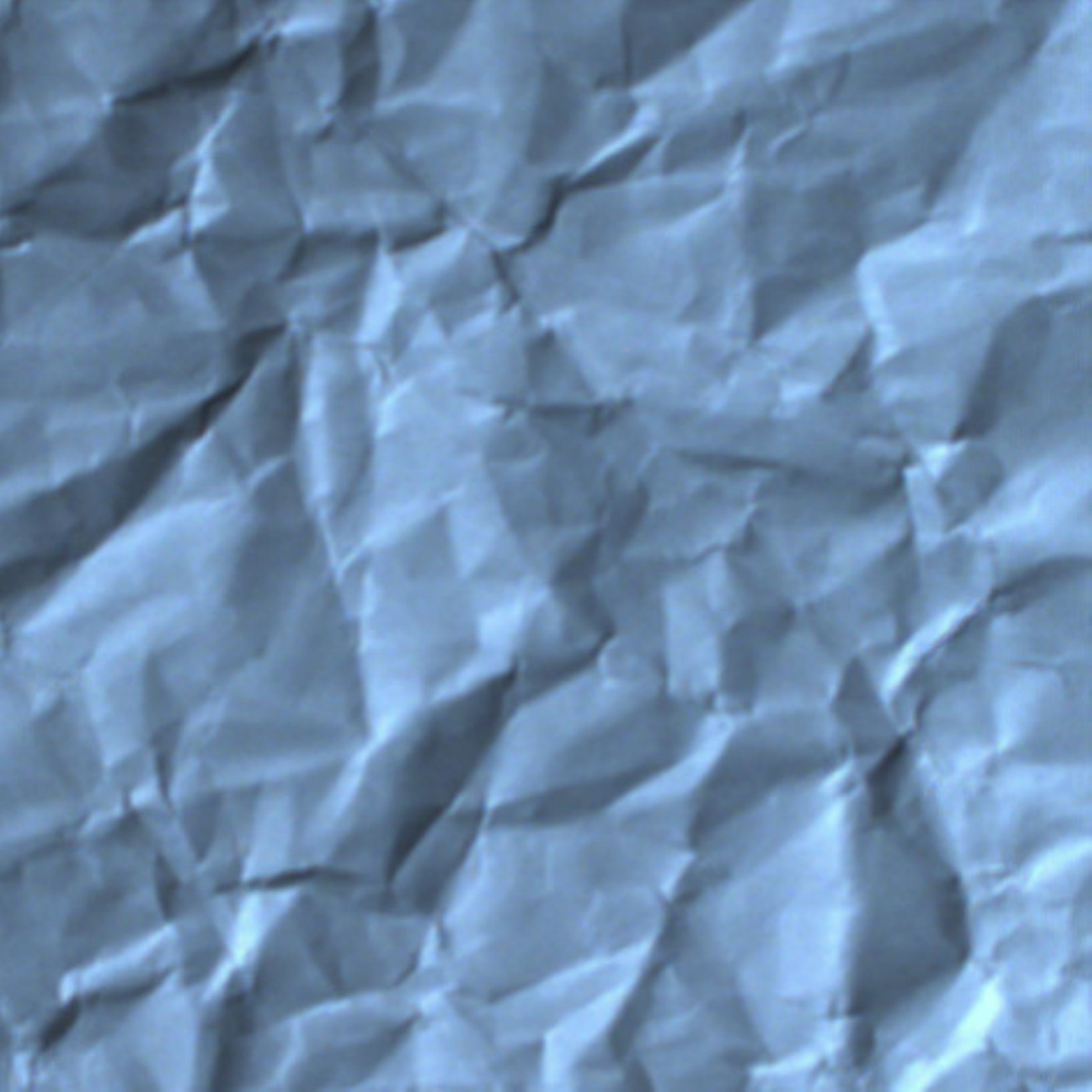} 
	\end{minipage}&
	\begin{minipage}{0.15\hsize}
			\includegraphics[clip, width=1.6cm]{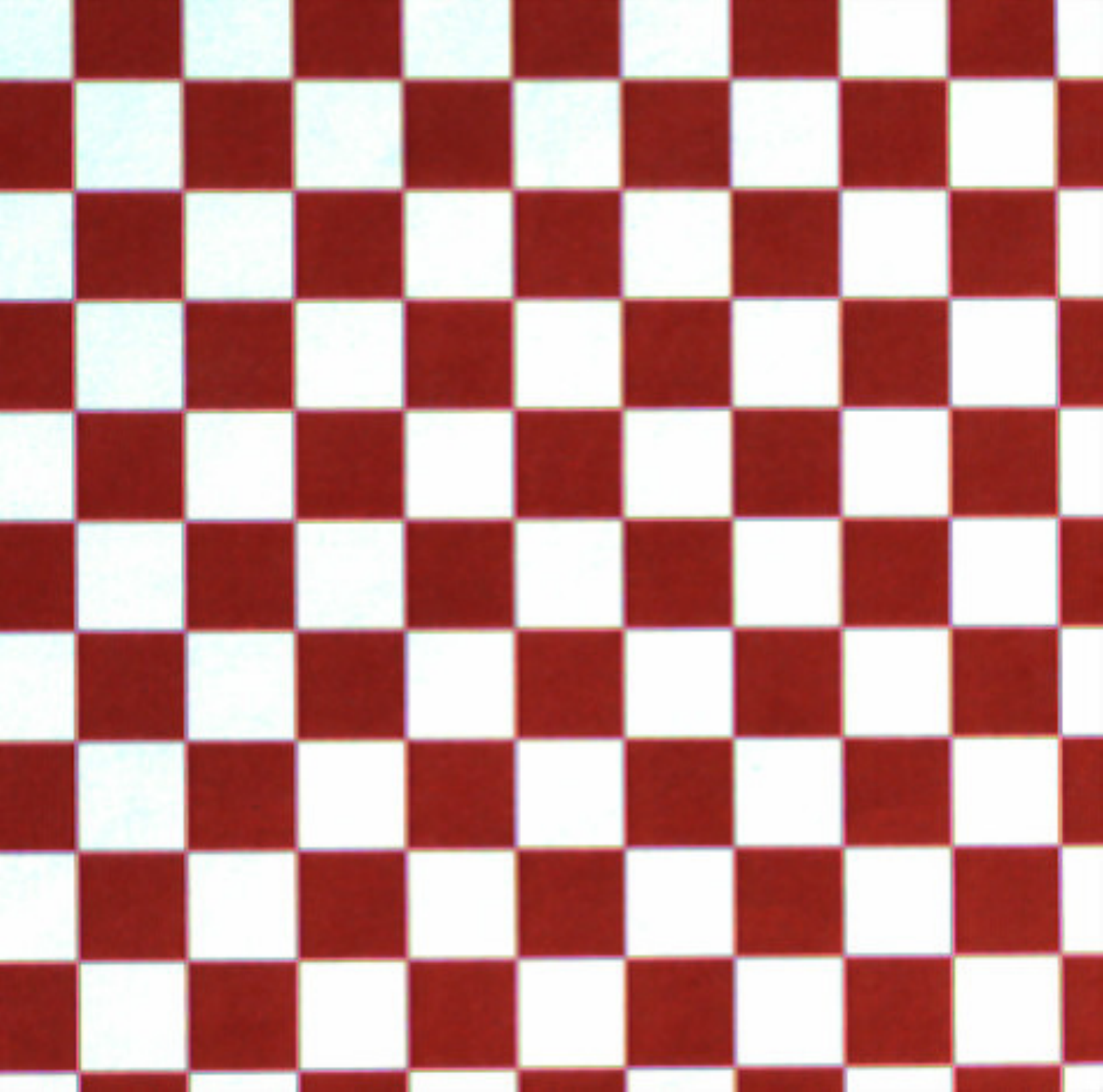}  
	\end{minipage}&
	\begin{minipage}{0.15\hsize}
			\includegraphics[clip, width=1.6cm]{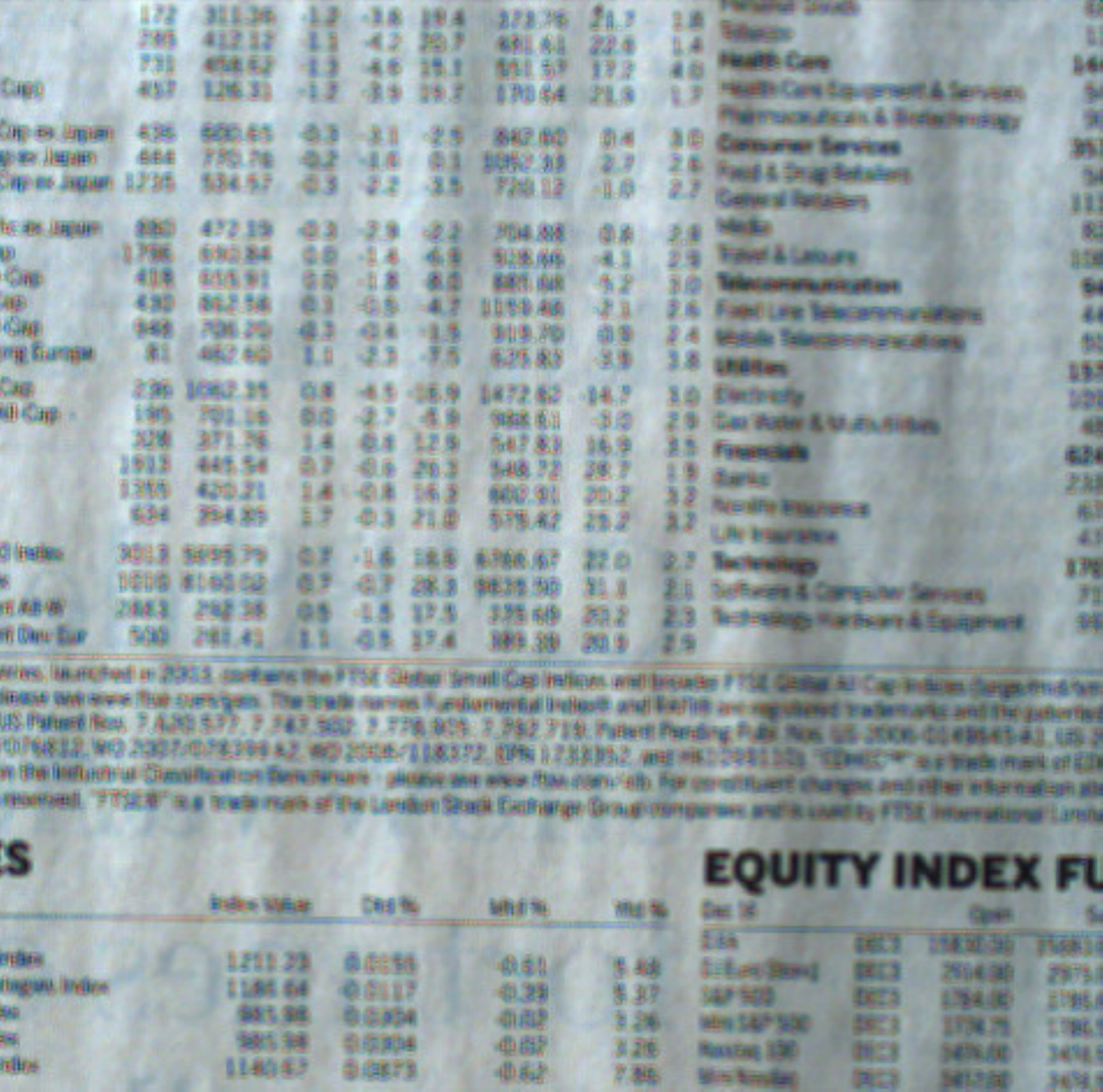}
	\end{minipage}&
	\begin{minipage}{0.15\hsize}
			\includegraphics[clip, width=1.6cm]{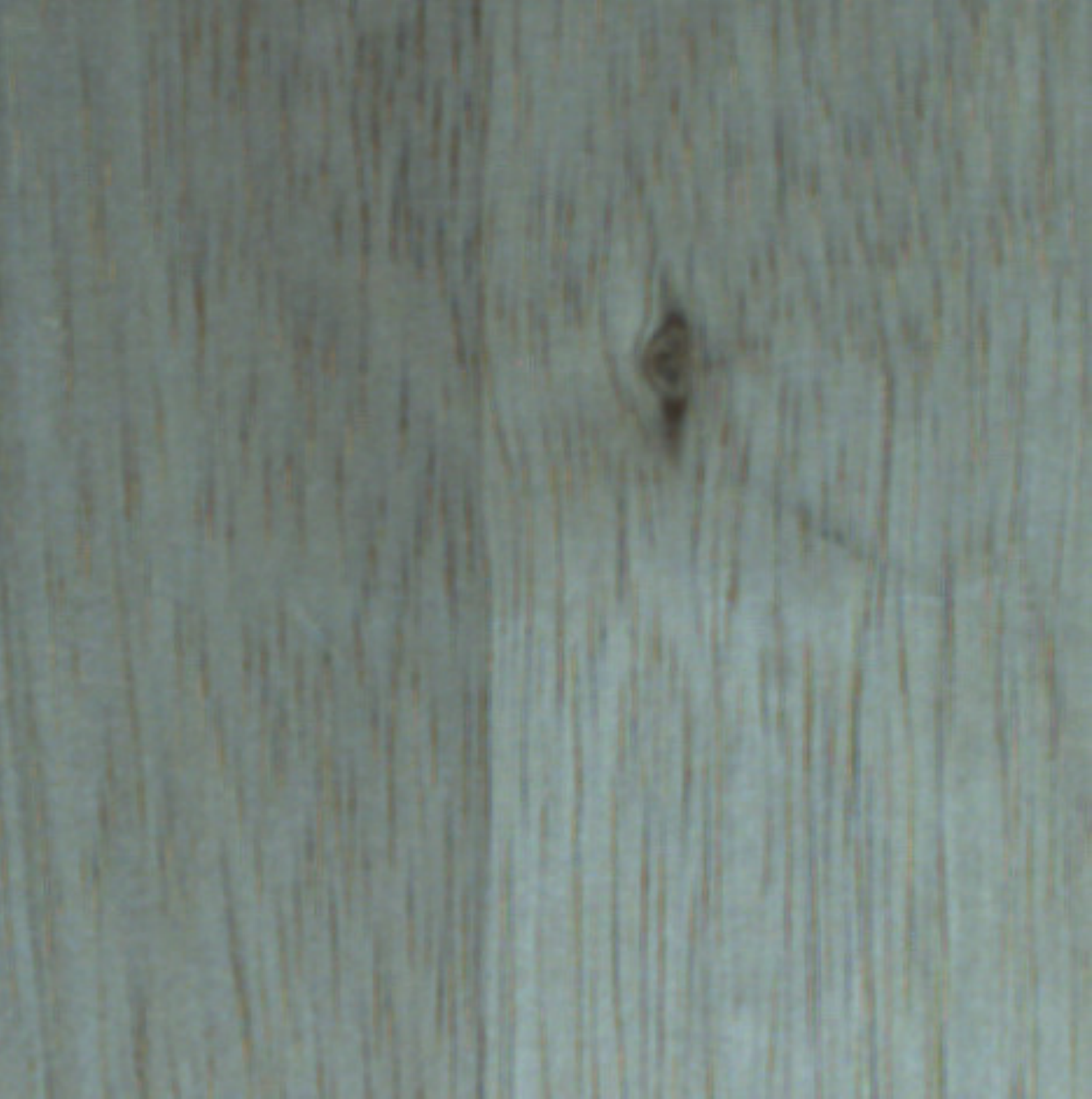}
	\end{minipage}\\

\vspace{1mm}			
\multirow{3}{*}{\rotatebox{90}{The number of used pattern}}&
		1&
	\begin{minipage}{0.15\hsize}
			\includegraphics[clip, width=1.6cm]{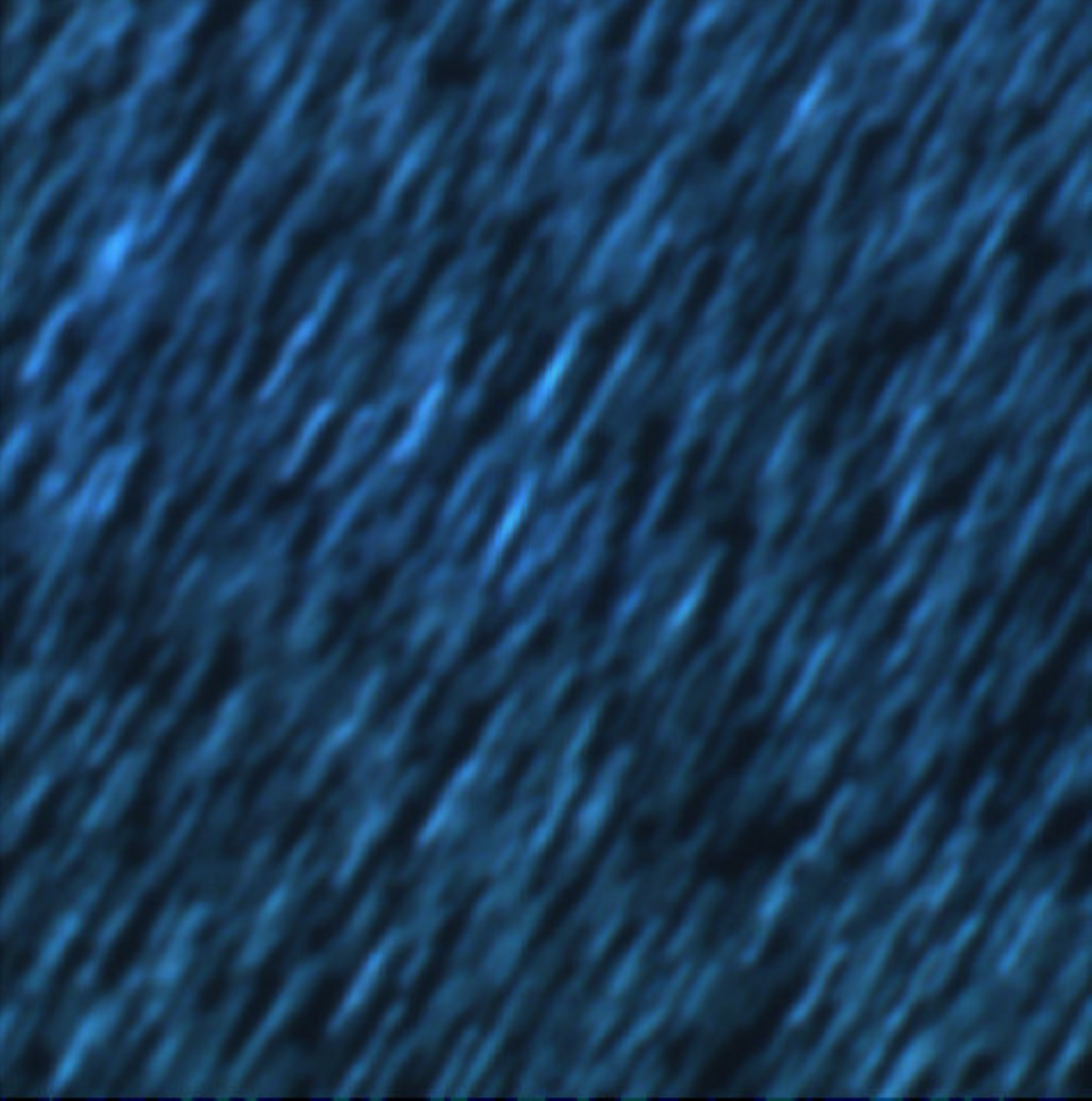} 
	\end{minipage}&
	\begin{minipage}{0.15\hsize}
			\includegraphics[clip, width=1.6cm]{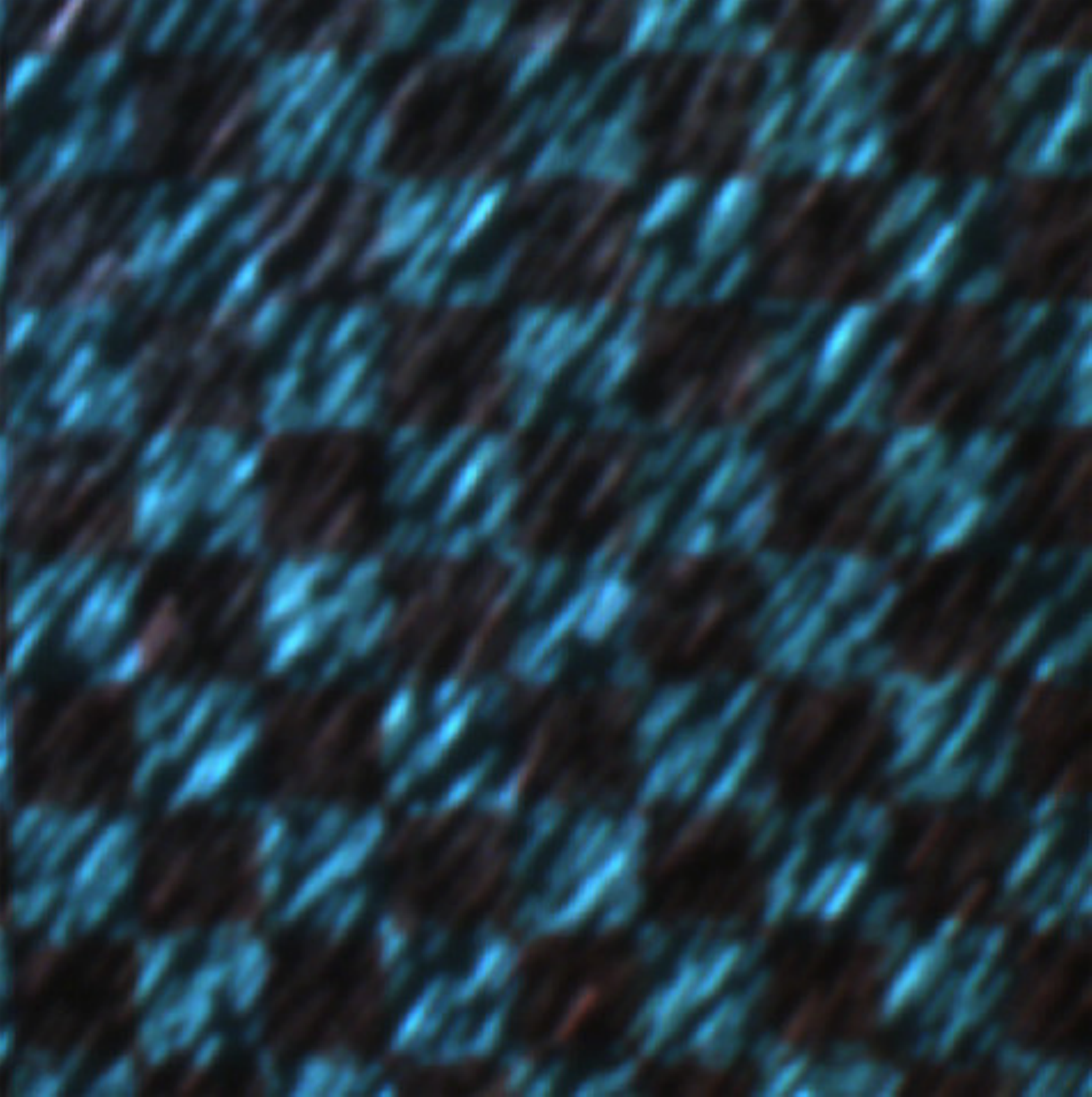}  
	\end{minipage}&
	\begin{minipage}{0.15\hsize}
			\includegraphics[clip, width=1.6cm]{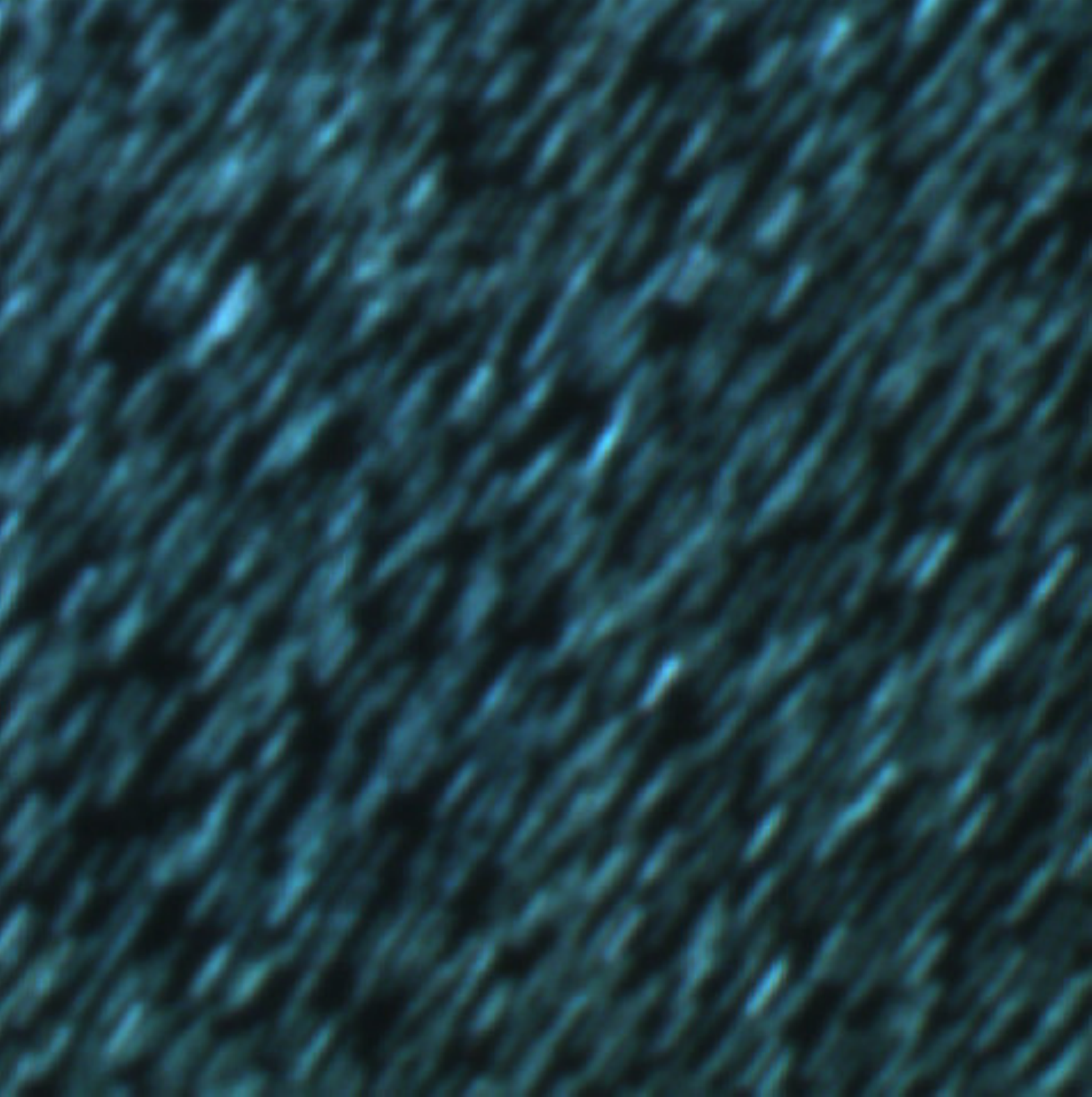}
	\end{minipage}&
	\begin{minipage}{0.15\hsize}
			\includegraphics[clip, width=1.6cm]{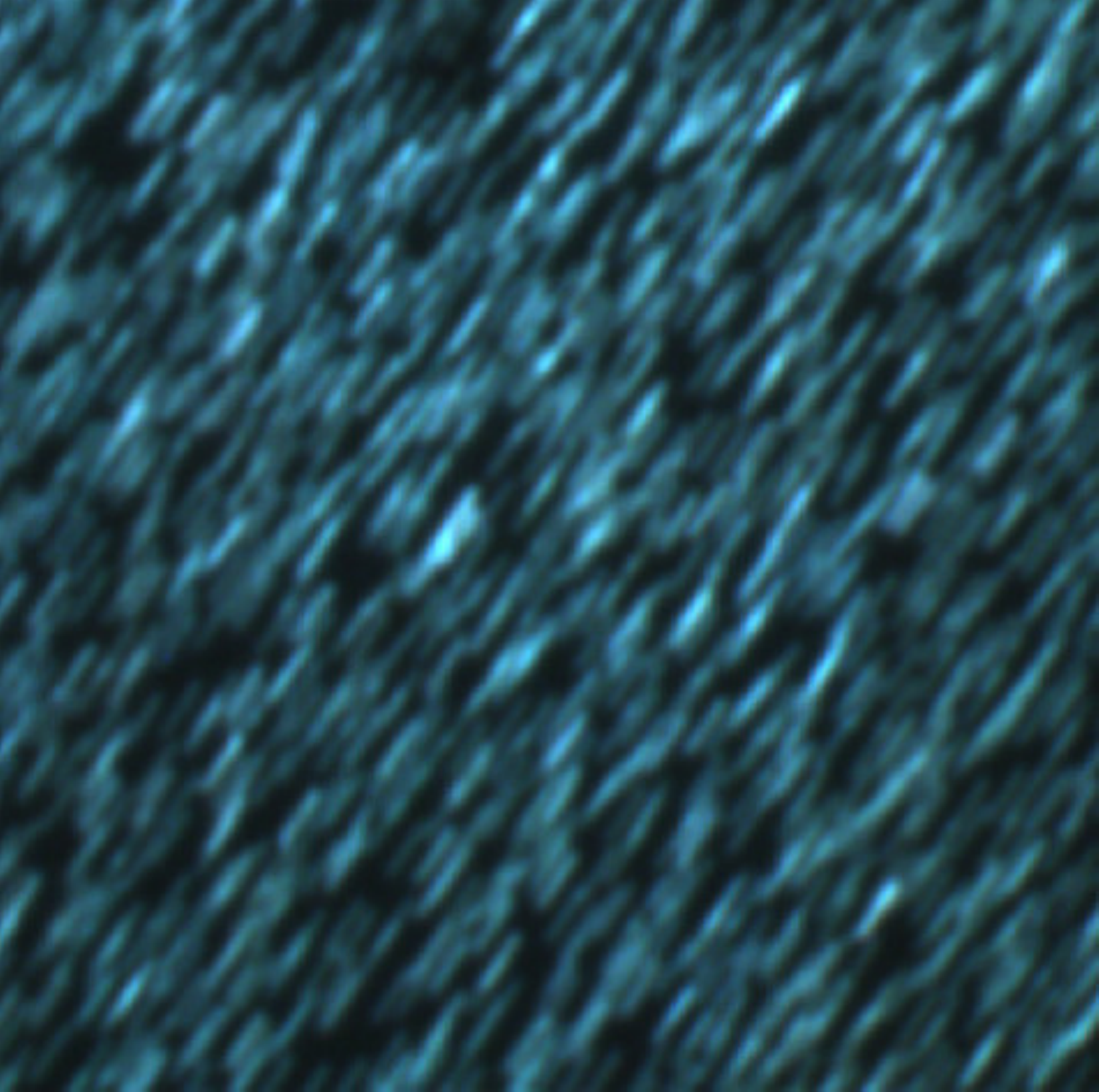}
	\end{minipage}\\

\vspace{1mm}			
&
		3&
	\begin{minipage}{0.15\hsize}
			\includegraphics[clip, width=1.6cm]{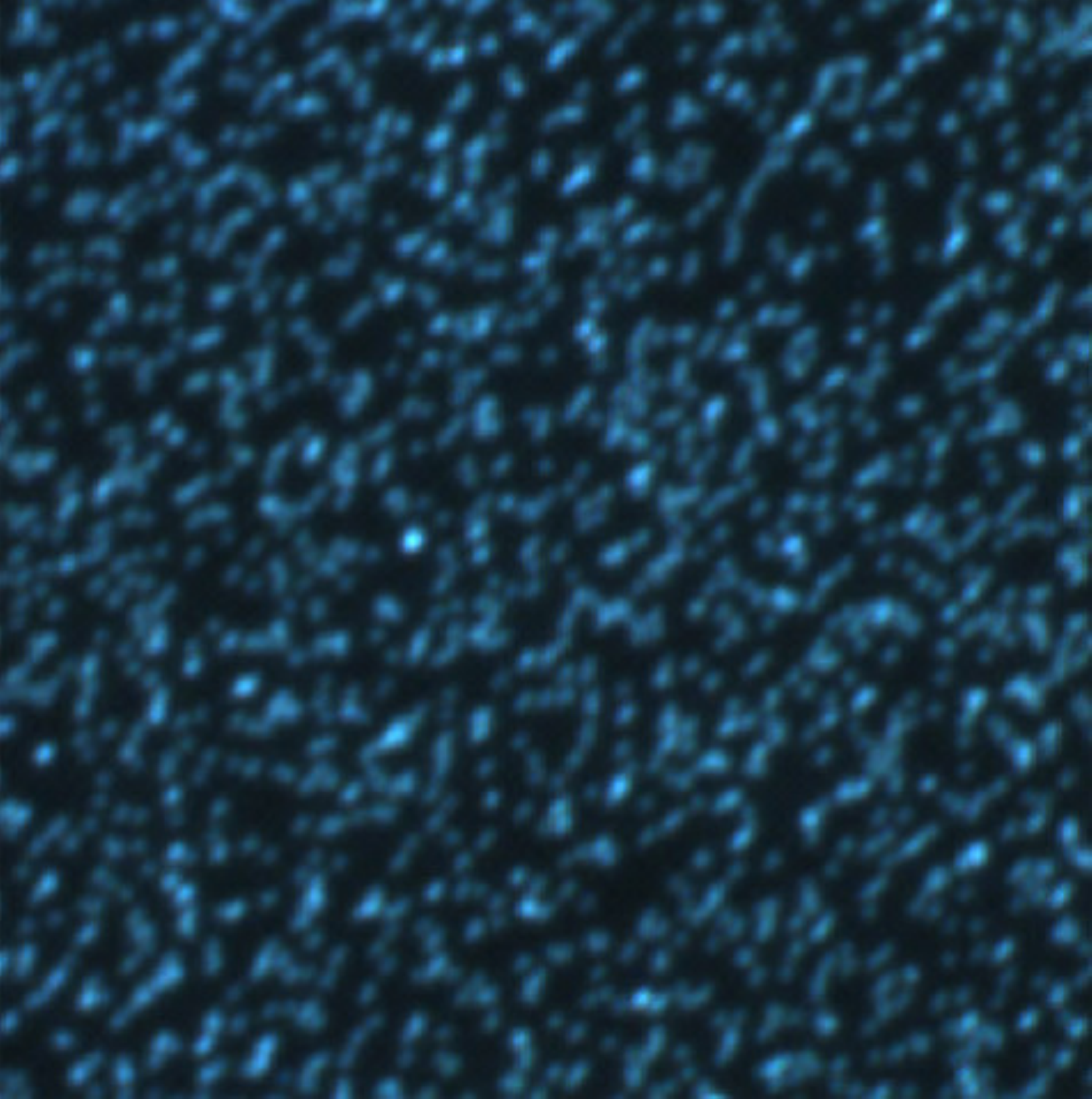} 
	\end{minipage}&
	\begin{minipage}{0.15\hsize}
			\includegraphics[clip, width=1.6cm]{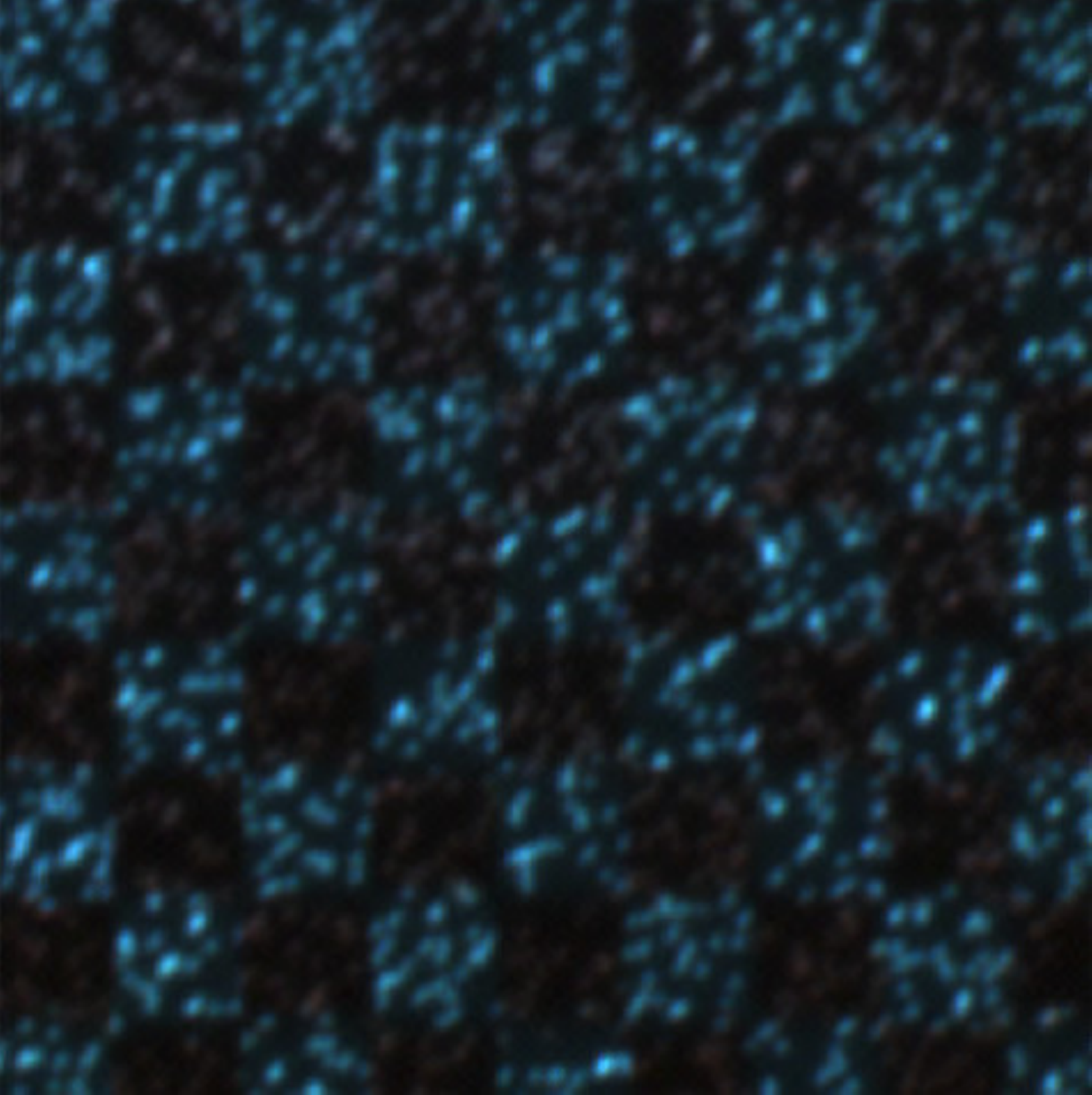}  
	\end{minipage}&
	\begin{minipage}{0.15\hsize}
			\includegraphics[clip, width=1.6cm]{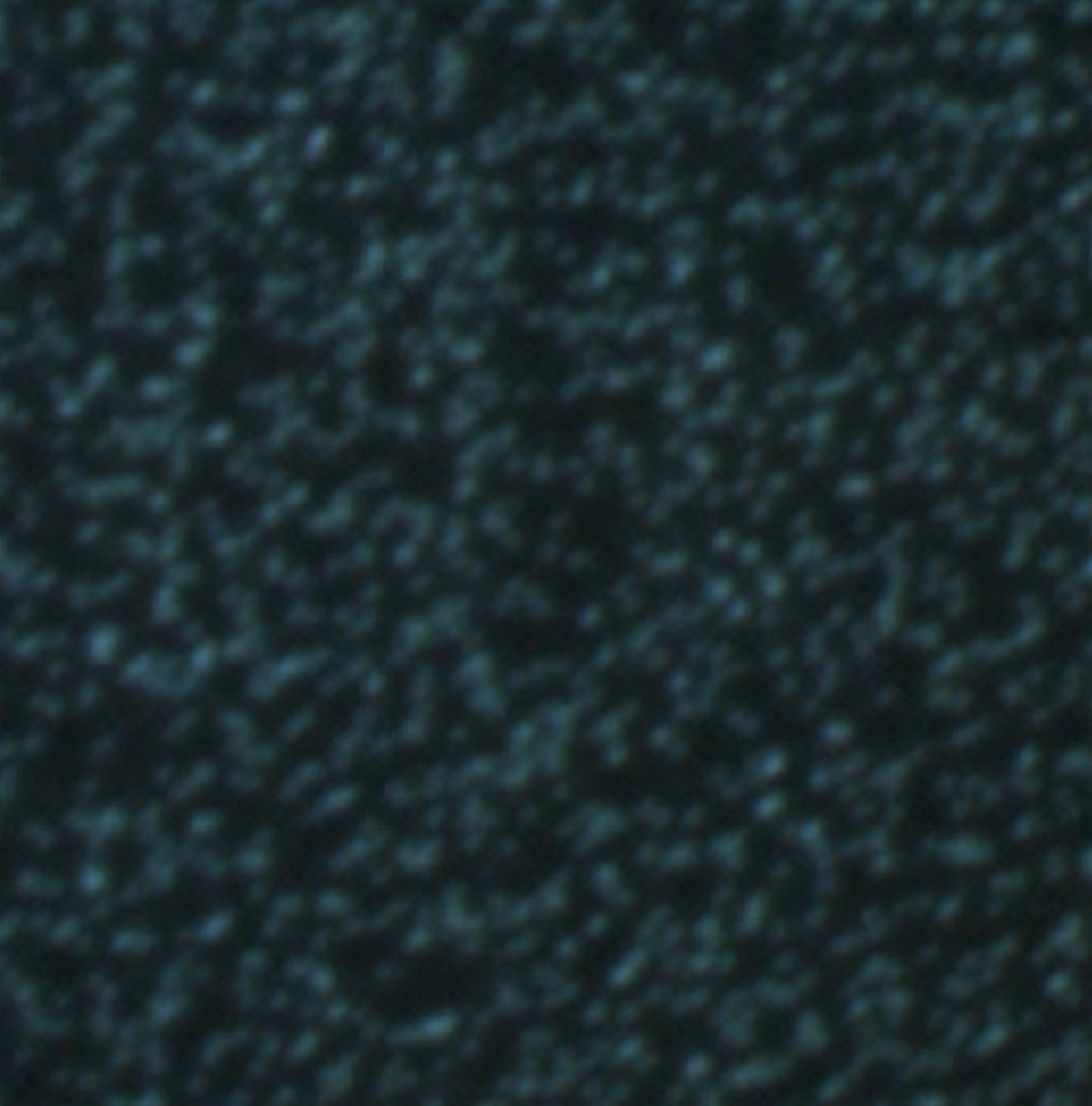}
	\end{minipage}&
	\begin{minipage}{0.15\hsize}
			\includegraphics[clip, width=1.6cm]{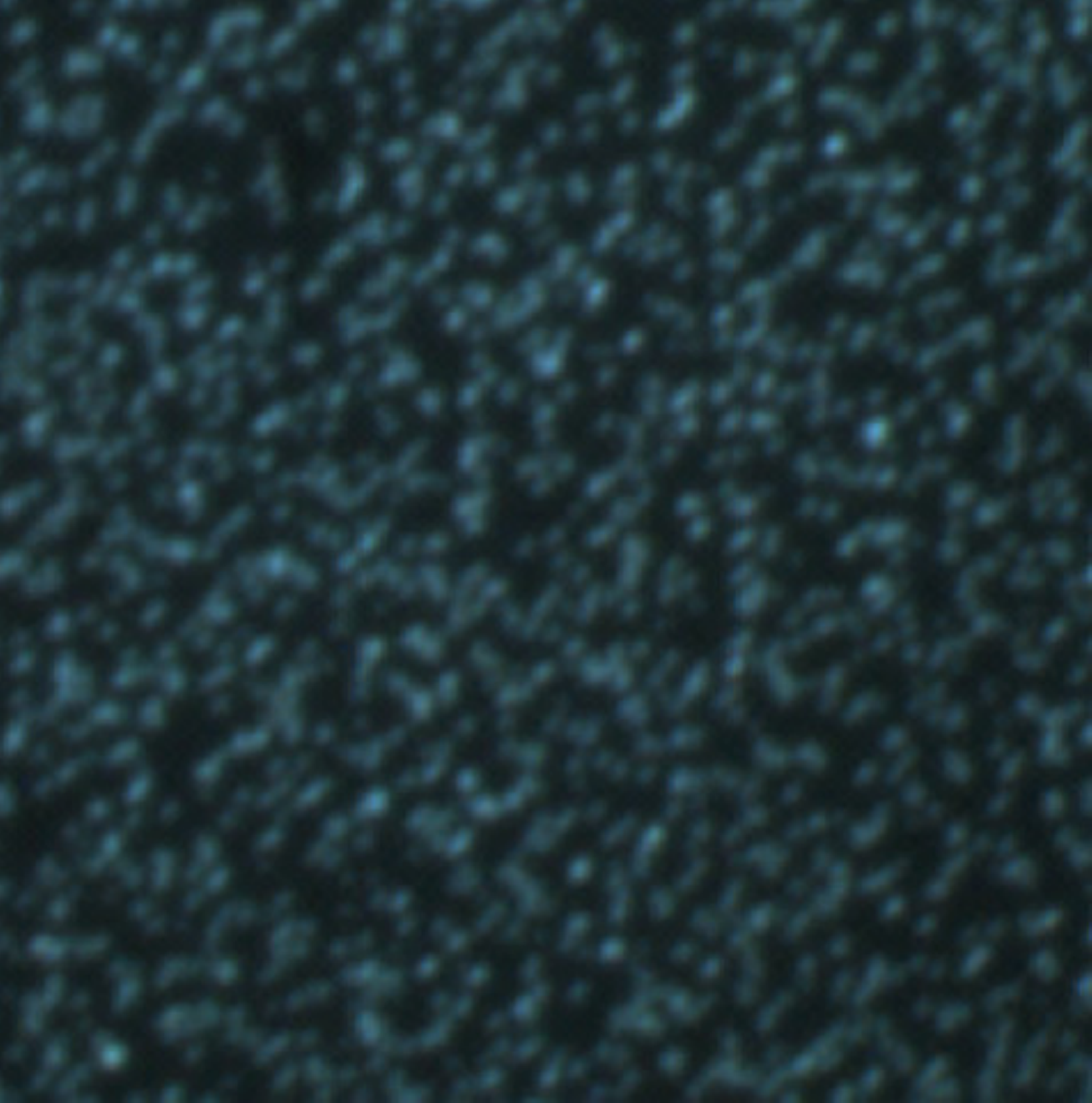}
	\end{minipage}\\

\vspace{1mm}			
&
		6&
	\begin{minipage}{0.15\hsize}
			\includegraphics[clip, width=1.6cm]{arxiv_figs/siwa3-eps-converted-to.pdf} 
	\end{minipage}&
	\begin{minipage}{0.15\hsize}
			\includegraphics[clip, width=1.6cm]{arxiv_figs/check3-eps-converted-to.pdf}  
	\end{minipage}&
	\begin{minipage}{0.15\hsize}
			\includegraphics[clip, width=1.6cm]{arxiv_figs/text3-eps-converted-to.pdf}
	\end{minipage}&
	\begin{minipage}{0.15\hsize}
			\includegraphics[clip, width=1.6cm]{arxiv_figs/wood3-eps-converted-to.pdf}
	\end{minipage}\\
			
&&			(a) & (b) & (c)&(d) \\
		\end{tabular}
		\caption{Example input data for the experiment of 
    Fig.~\ref{fig:Graph_pat_num}. (a) crumpled paper, (b) checker pattern, (c) 
    newspaper and (d) wood under the condition of {\it slow} velocity in Fig.~\ref{fig:Graph_velocity}.}
\vspace{-3mm}			
%		\knote{枚数変化の計測画像例を載せる（テクスチャも）。復元結果も載せる}}
		\label{fig:Image_pat_num}
\end{figure}

\begin{figure}[tb]
\begin{center}
\vspace{-5mm}
	\includegraphics[width=75mm]{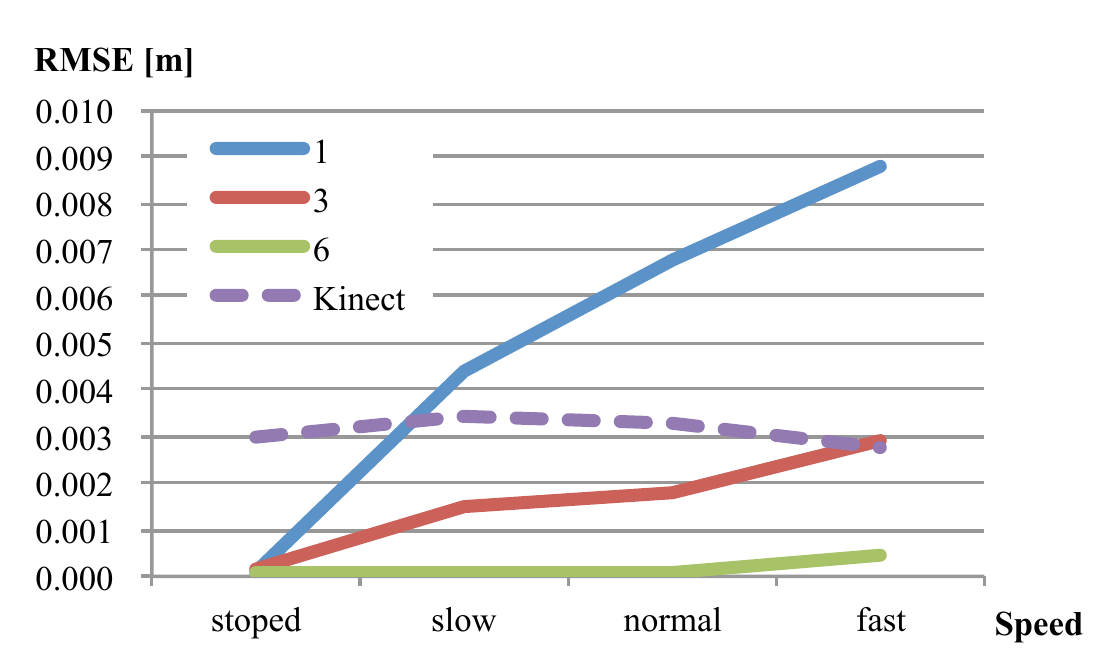}\\
	\caption{Accuracy evaluation by reconstructing a plane board with different velocity 
    of target object. RMSEs of reconstructed points from the fitted planes are shown.
	Horizontal axis represents the velocity of the object. It is clearly shown 
    that accuracy degrades when velocity become faster; typically the number of 
pattern is small.}
%	\caption{\knote{グラフ２。横軸速度、縦軸RMSE。}}
	\label{fig:Graph_velocity}
\end{center}
%\end{figure}
%
%\begin{figure}[tb]
		\centering
		\begin{tabular}{llcccc}
\vspace{1mm}			
\multirow{3}{*}{\rotatebox{90}{The number of used pattern}}&
		1&
	\begin{minipage}{0.15\hsize}
			\includegraphics[clip, width=1.6cm]{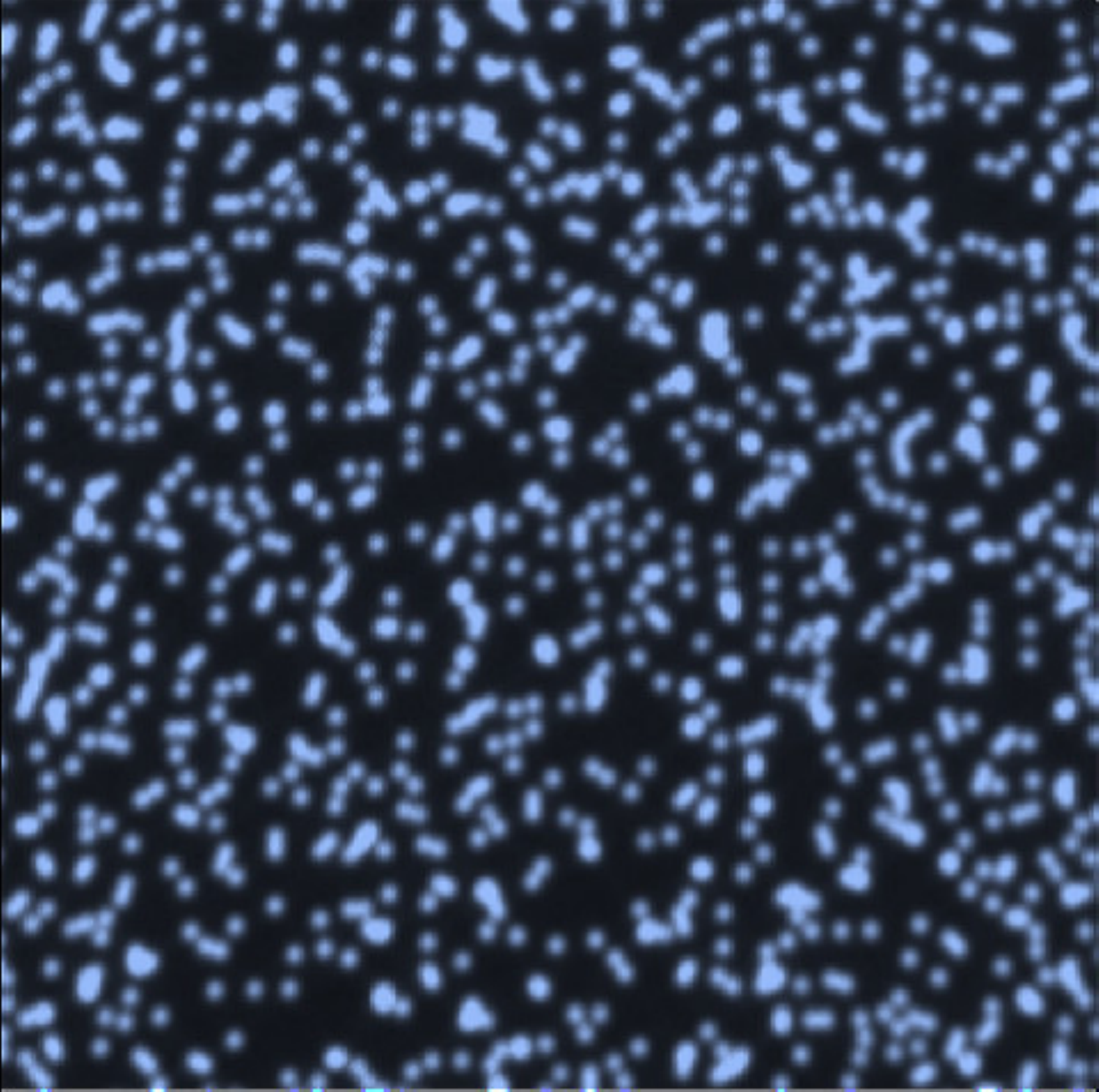} 
	\end{minipage}&
	\begin{minipage}{0.15\hsize}
			\includegraphics[clip, width=1.6cm]{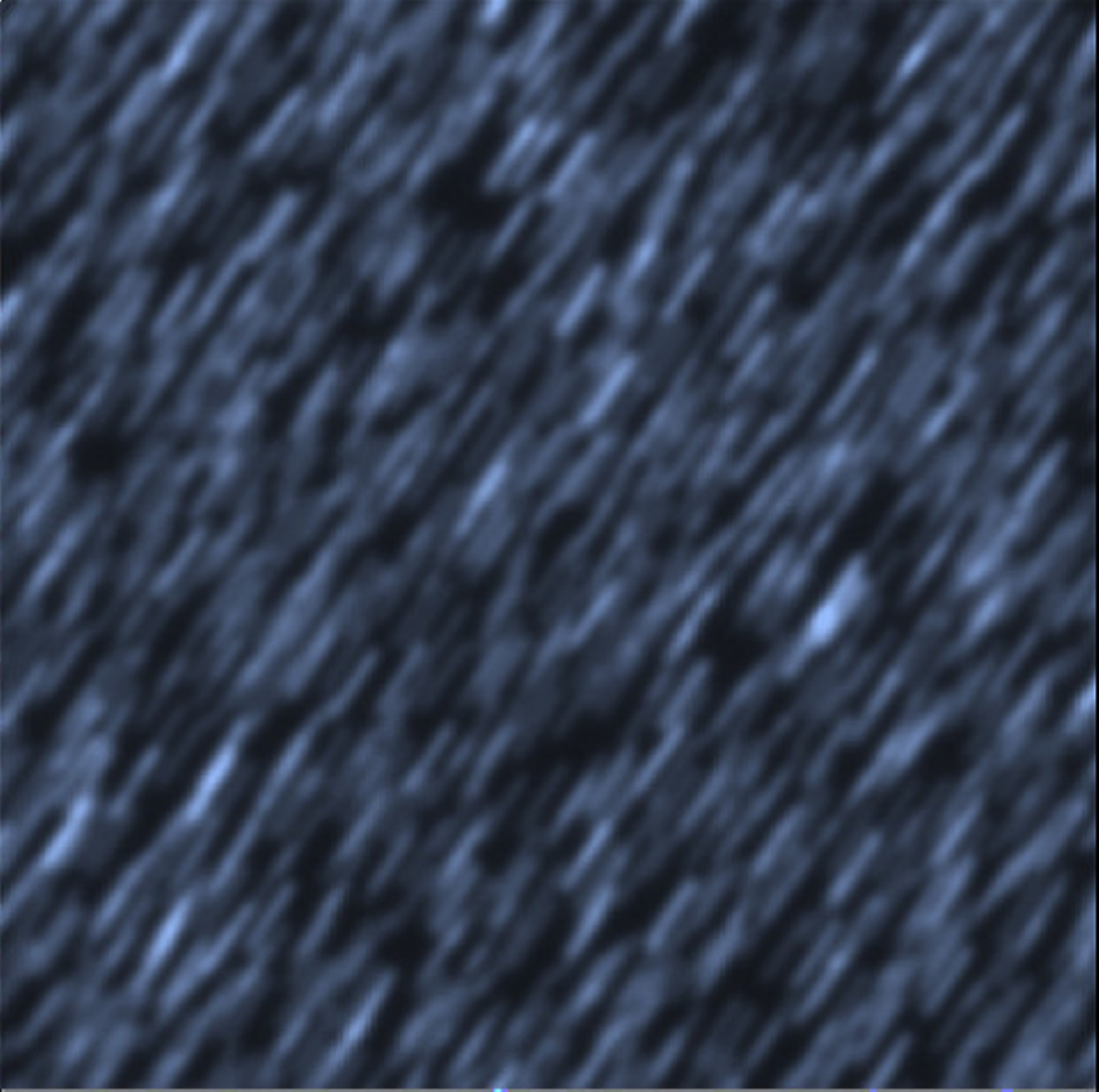}
	\end{minipage}&
	\begin{minipage}{0.15\hsize}
			\includegraphics[clip, width=1.6cm]{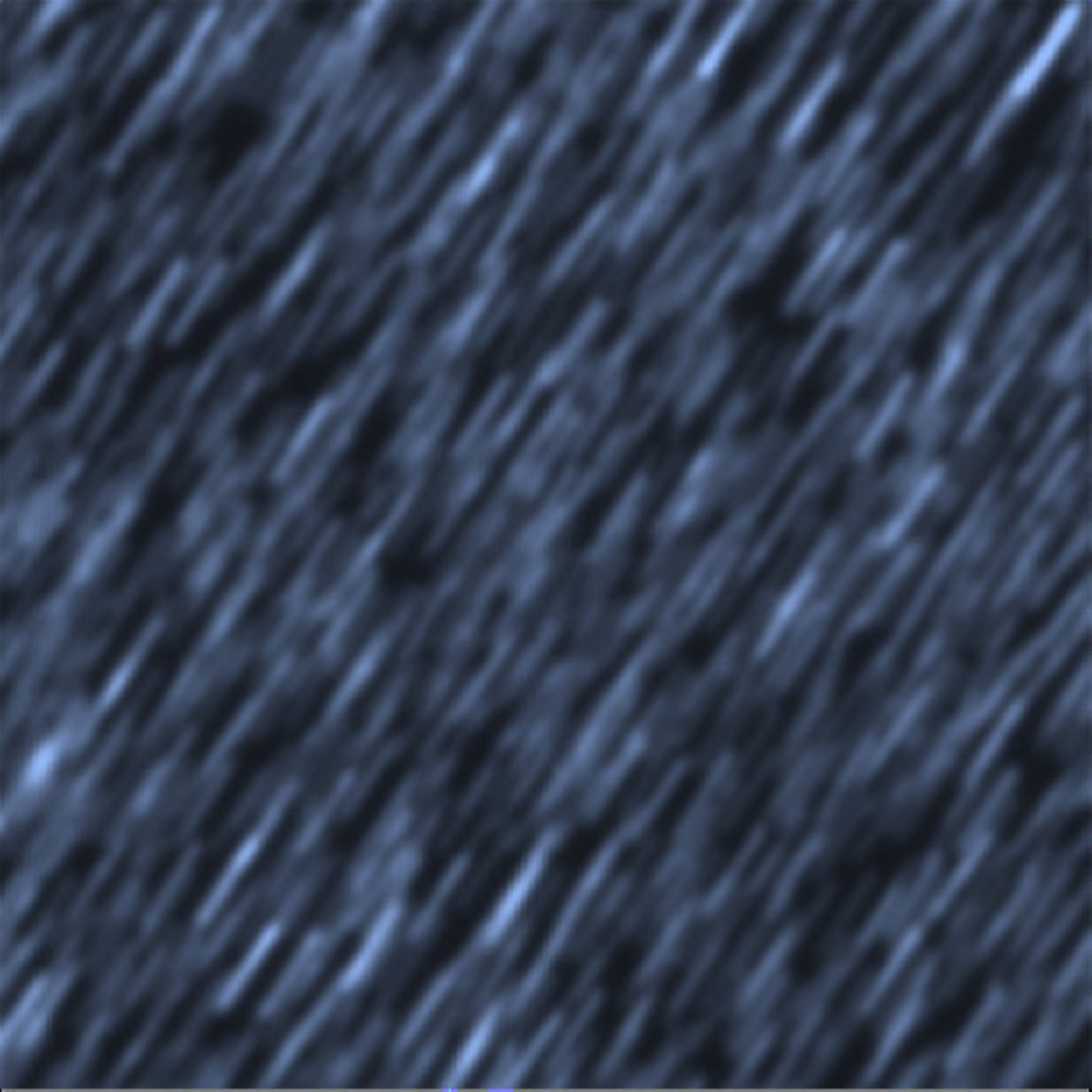}
	\end{minipage}&
	\begin{minipage}{0.15\hsize}
			\includegraphics[clip, width=1.6cm]{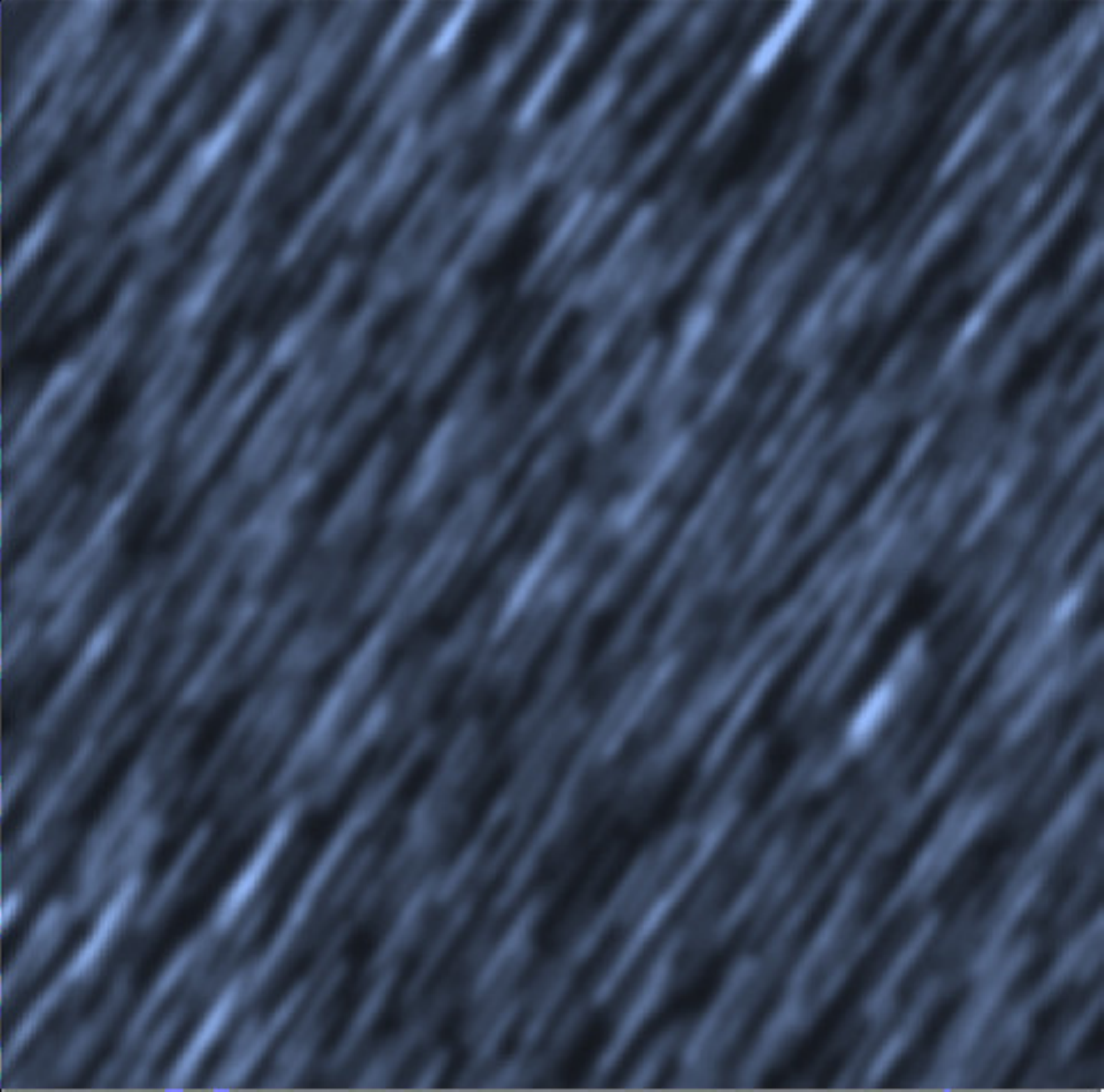}
	\end{minipage}\\
\vspace{1mm}
&
3&
	\begin{minipage}{0.15\hsize}
			\includegraphics[clip, width=1.6cm]{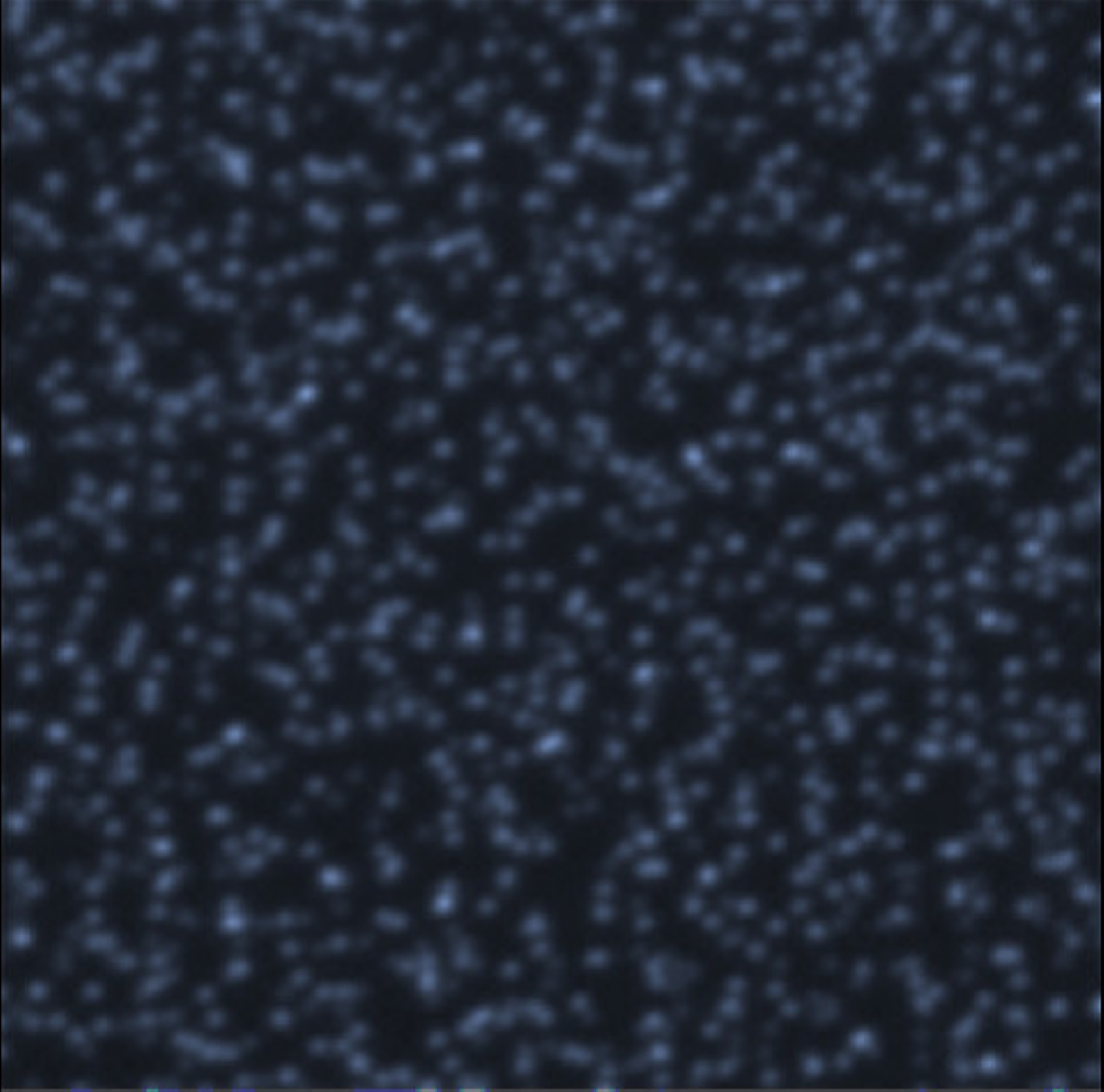}
	\end{minipage}&
	\begin{minipage}{0.15\hsize}
			\includegraphics[clip, width=1.6cm]{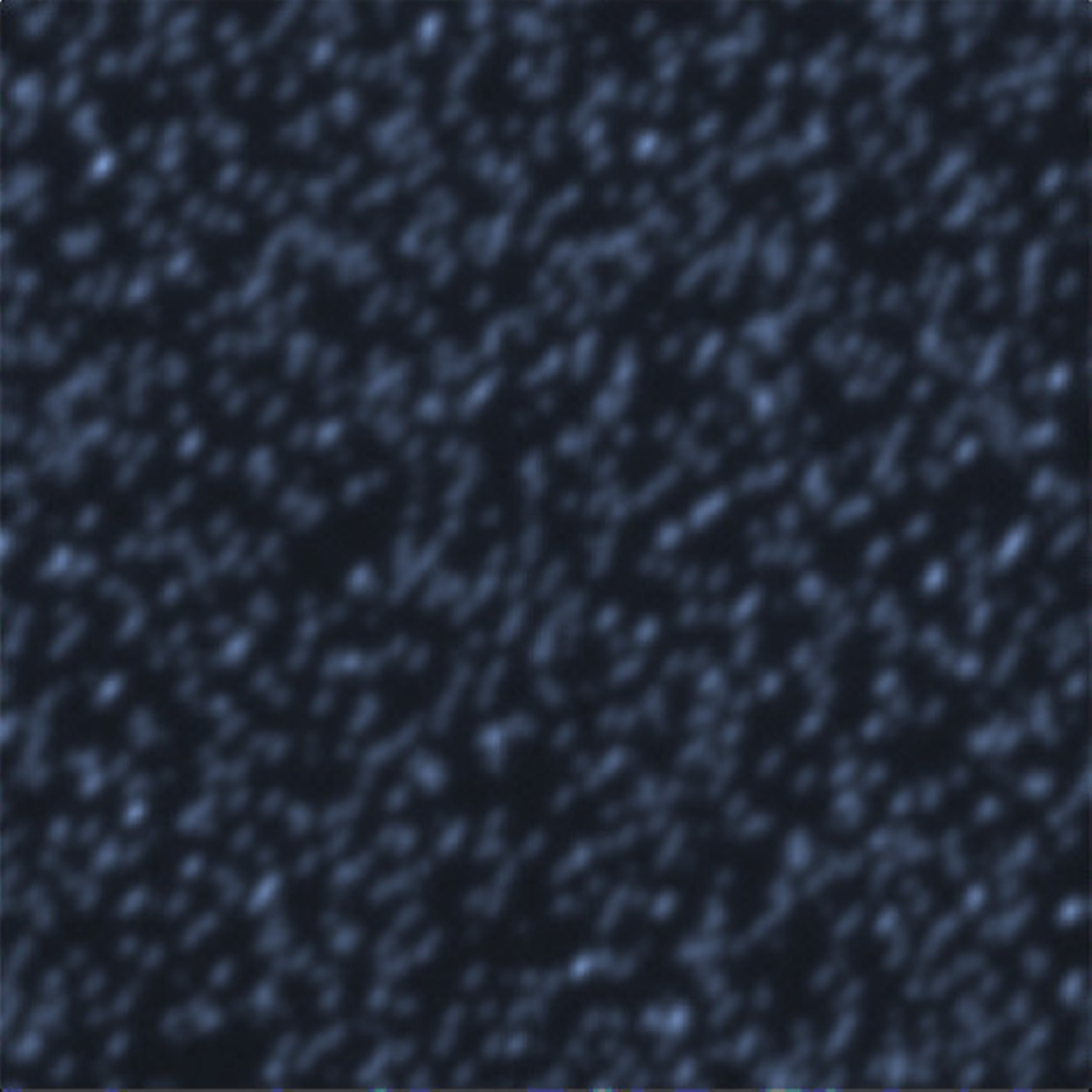} 
	\end{minipage}&
	\begin{minipage}{0.15\hsize}
			\includegraphics[clip, width=1.6cm]{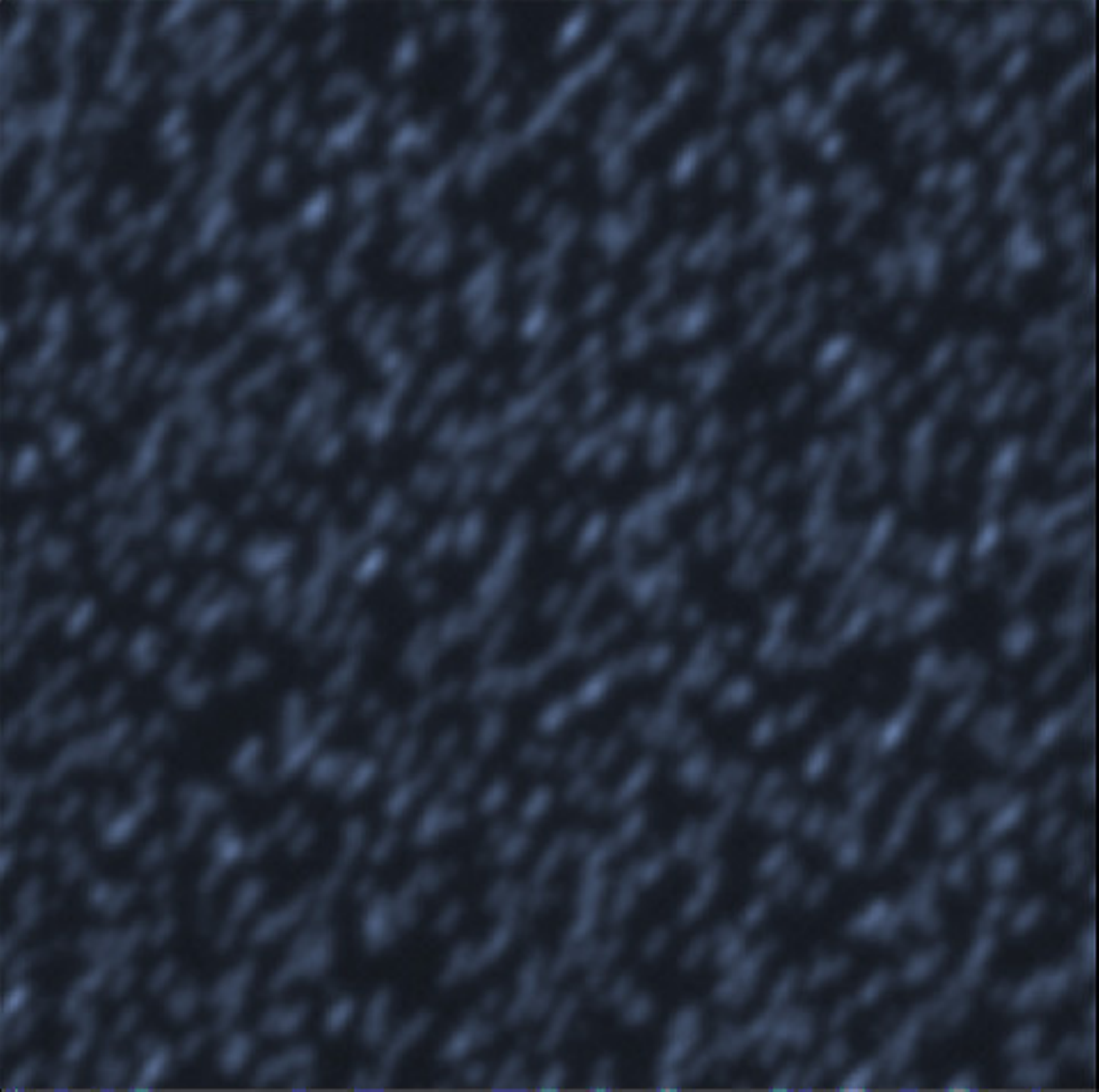}
	\end{minipage}&
	\begin{minipage}{0.15\hsize}
			\includegraphics[clip, width=1.6cm]{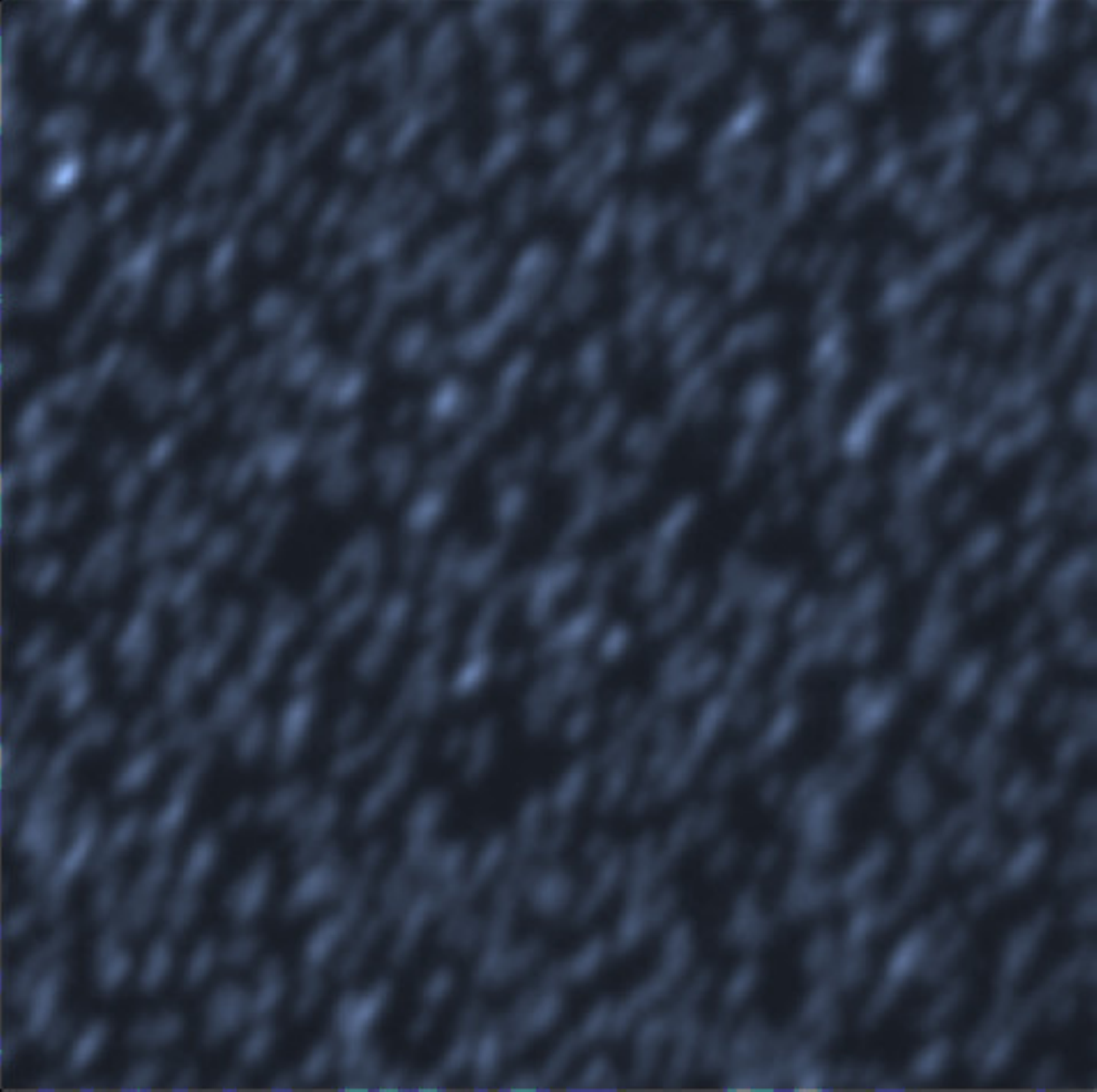}
	\end{minipage}\\
&			
6&
	\begin{minipage}{0.15\hsize}
			\includegraphics[clip, width=1.6cm]{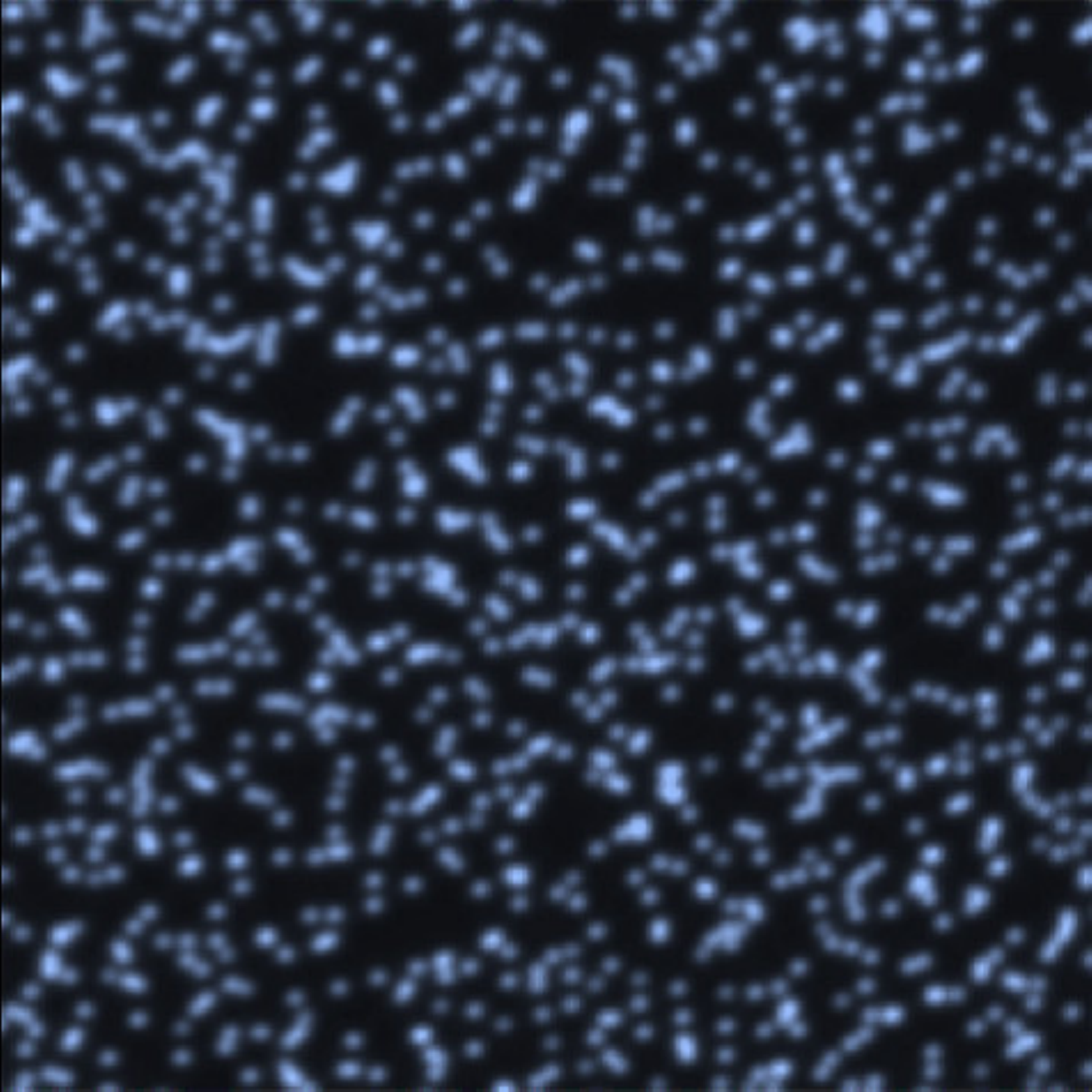}
	\end{minipage}&
	\begin{minipage}{0.15\hsize}
			\includegraphics[clip, width=1.6cm]{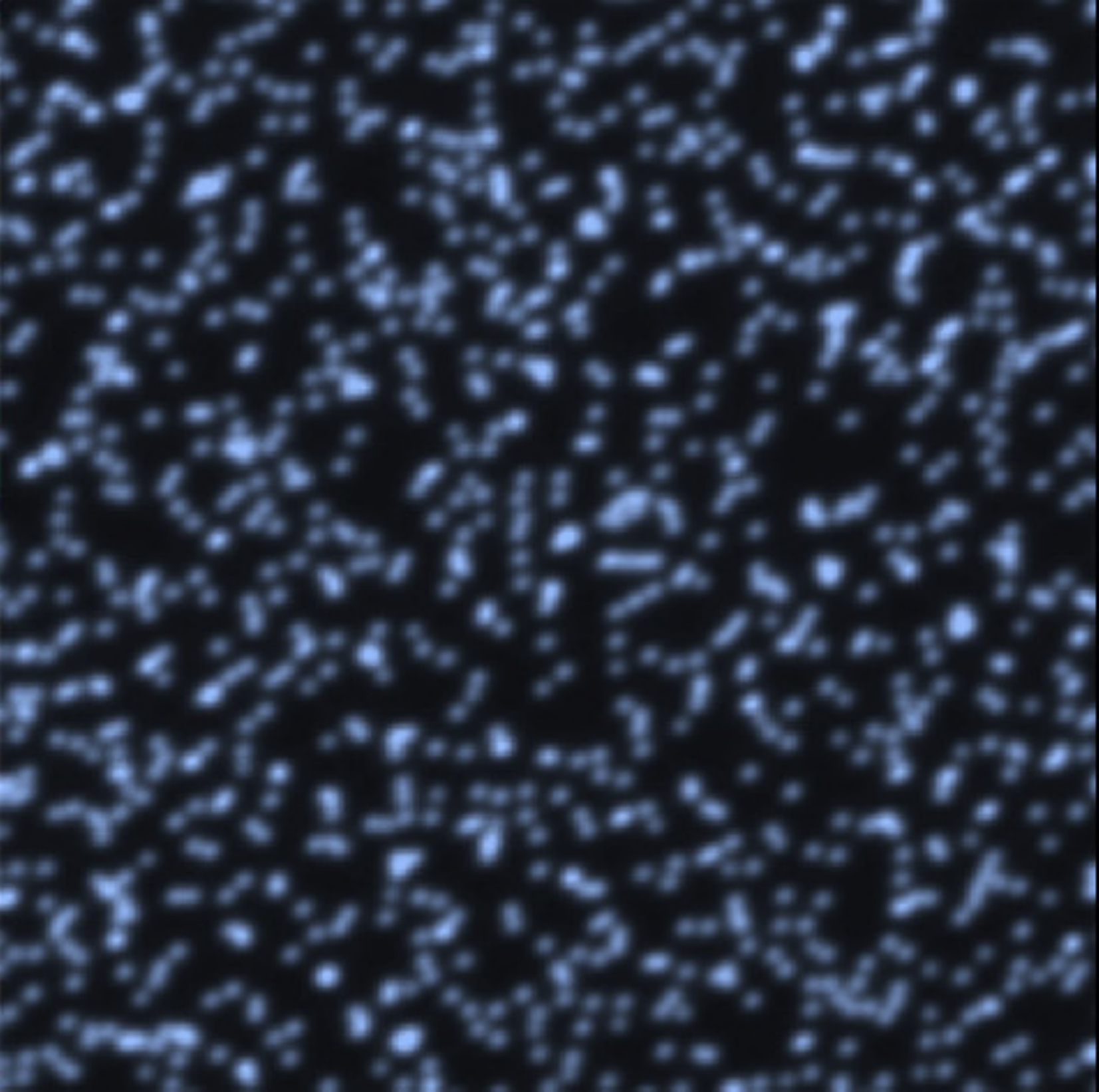} 
	\end{minipage}&
	\begin{minipage}{0.15\hsize}
			\includegraphics[clip, width=1.6cm]{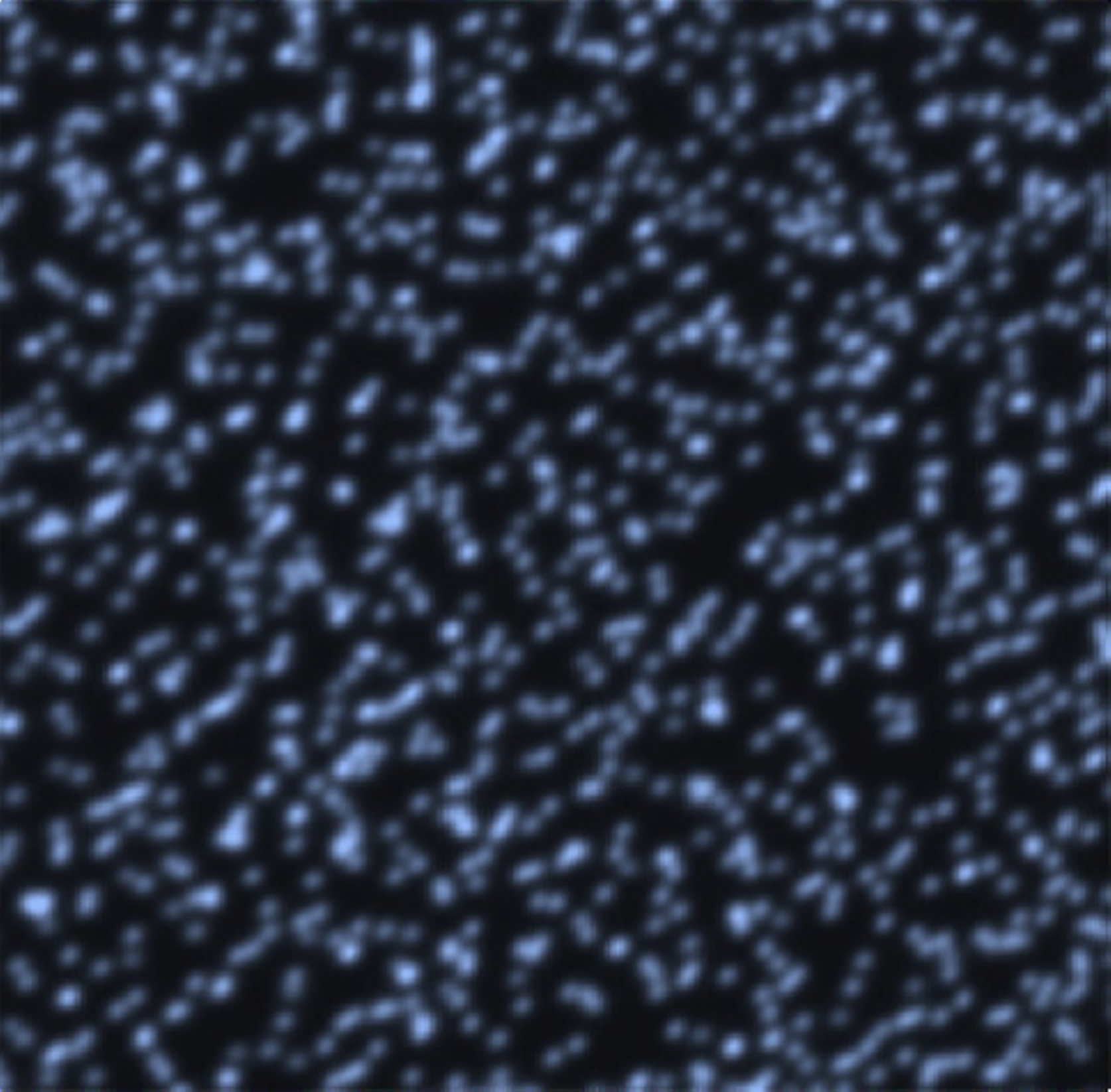}
	\end{minipage}&
	\begin{minipage}{0.15\hsize}
			\includegraphics[clip, width=1.6cm]{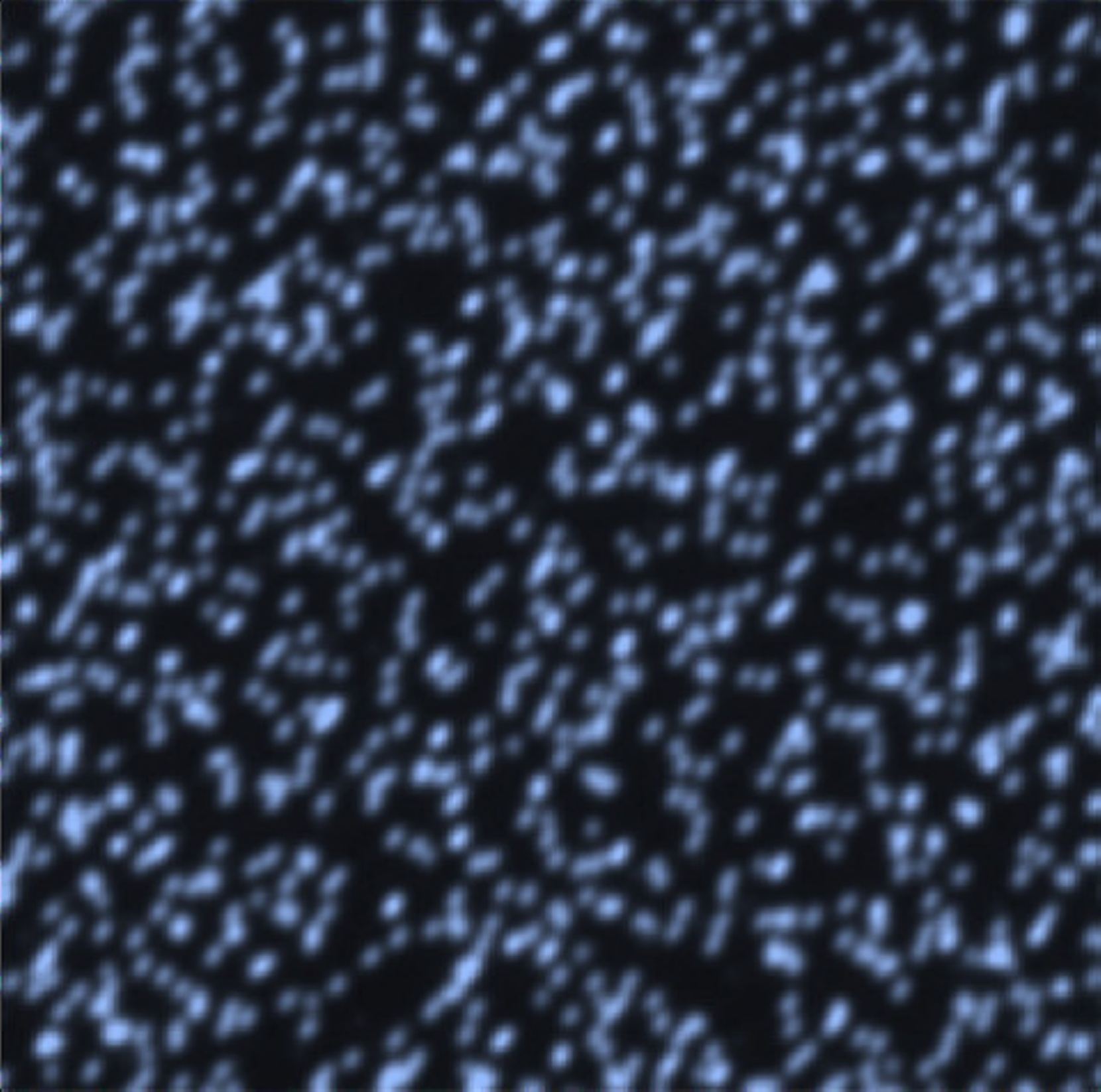}
	\end{minipage}\\
			
&			 & (a) & (b) & (c)&(d) \\
		\end{tabular}
		\caption{Example input data for the experiment of Fig.~\ref{fig:Graph_velocity}. (a) stop, (b) slow,  (c) normal and (d) fast motion, respectively.}
%		\knote{速度変化の計測画像例を載せる。復元結果も載せる}}
		\label{fig:Image_velocity}
\vspace{-3mm}
\end{figure}

The first experiment was conducted by using a video projector and a CCD camera.
% shown in Fig.\ref{fig:actualequipment}. %{fig:HalfMillar}(b).
Reference database was captured by moving the target screen by motorized stage between 500mm-800mm 
from the projector and the camera, 
%Because of the limitation of the length of the motorized stage, we put a close-up 
%lens to change the scale as to be 1/3 of real length. 
%In our set-up, the motion a range of the screen was 500mm-800mm from the projector and the camera, 
%in-focus distance was 650mm $\pm100$mm for the projector, 
and the capturing interval was 1.0mm.
Fps and a resolution of the camera was 3Hz and 1600*1200 pixels and 30Hz and 
1024*768 pixels for the video projector. 
%The baseline between the 
%camera and the projector was approximately 150mm.

For evaluation, we attached a target board onto the motorized stage, and captured it with
the camera while the board was moving under
different conditions, such as with different numbers of projected patterns, 
different velocity and different material of the target object. 
Here, the case of using one pattern is same as the conventional active stereo 
method, % using NCC.
therefore, to equalize the experimental conditions between the measurement using
different numbers of patterns, 
we adjusted pattern density so that the density of the integrated patterns 
becomes the same, as shown in Fig.~\ref{fig:Image_pat_density_equalize}. 

We applied our algorithm to the captured images using the stored database with
matching window size 12*12 pixels. % in this experiment.
%Under the condition, a total of 14 images ware captured.
The depth value was also estimated by
%estimated by using the proposed method and other methods
the commercial device Kinect v1~\cite{Kinect} to show the standard ability on 
scanning moving objects.
%and NCC with a static pattern, which is just an 
%average of multiple patterns,  for a comparison
Results are shown in Fig.~\ref{fig:Graph_pat_num} and \ref{fig:Graph_velocity}.
%Estimated depth was calculated by averaging 1000 points of estimated depths of 
%each frame. 
%In the graph, we can observe only our method and 
%Graycode~\cite{Inokuchi:ICPR84} can recover the correct depth for all the ranges. 
%Further, Graycode method gradually reduces its 
%precision at out-of-focus zone, whereas our method keeps almost same accuracy for all the ranges.

In the first graph in Fig.\ref{fig:Graph_pat_num}, we can clearly observe 
that the increase in the number of projected patterns improves the RMSE for all 
textures and materials. Of all textures and materials, checker 
pattern has the worst RMSE; we consider that this is caused by the small size of checker 
pattern used, which is a similar size to the matching window, thus interference 
occurs during NCC calculation.
Fig.\ref{fig:Image_pat_num} shows examples of the actual captured images. 
As can be seen, a larger number of patterns result in 
a sharper captured image, which in turn results in better RMSE.

In the next graph of Fig.\ref{fig:Graph_velocity}, we can clearly see that an
increase in velocity degrades the RMSE especially when the number of the patterns is 
small. 
%\knoteA{
Since we used an ordinary video 
projector for this experiment and the fps is just 3Hz,
captured 
patterns are significantly blurred for a single pattern (as shown in Fig.~\ref{fig:Image_velocity},) whereas 
sharp patterns are preserved with multiple pattern projection. This
results in the maintenance of the lowest RMSE of all, at all velocities.
The reason why Kinect has almost constant error values at various velocities is that the motorized stage is so slow even at the maximum speed that no motion blur occurred with Kinect. 

\subsection{Temporal super-resolution of nonuniform velocity} %Arbitrary shape reconstruction}

%\knote{これは必要}

\begin{figure}[tb]
%\vspace{-5mm}
    \begin{minipage}{0.20\textwidth}
\begin{center}
	\includegraphics[width=20mm]{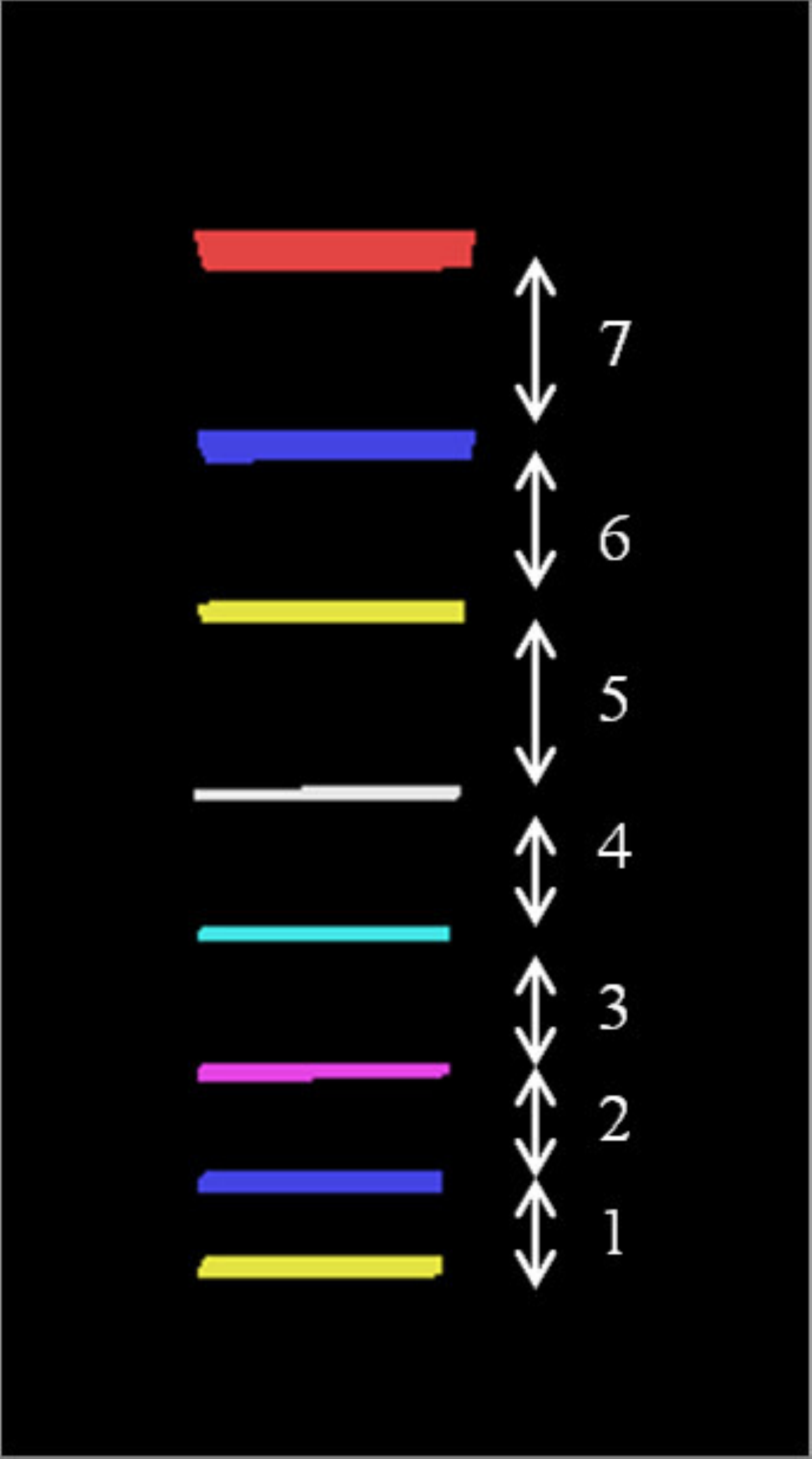}
	\caption{Super-resolved shape of moving board with acceleration. 
    Different intervals are clearly shown.}
%	\caption{\knote{板。計測画像、結果を載せる　超解像の様子　きれいなものと画像差し替え}}
	\label{fig:super_reso_accel}
\end{center}
    \end{minipage}
%\end{center}
%\end{figure}
%
\hspace{1mm}
%
%\begin{figure}[tb]
%\begin{center}
    \begin{minipage}{0.27\textwidth}
\begin{center}
	\includegraphics[width=45mm]{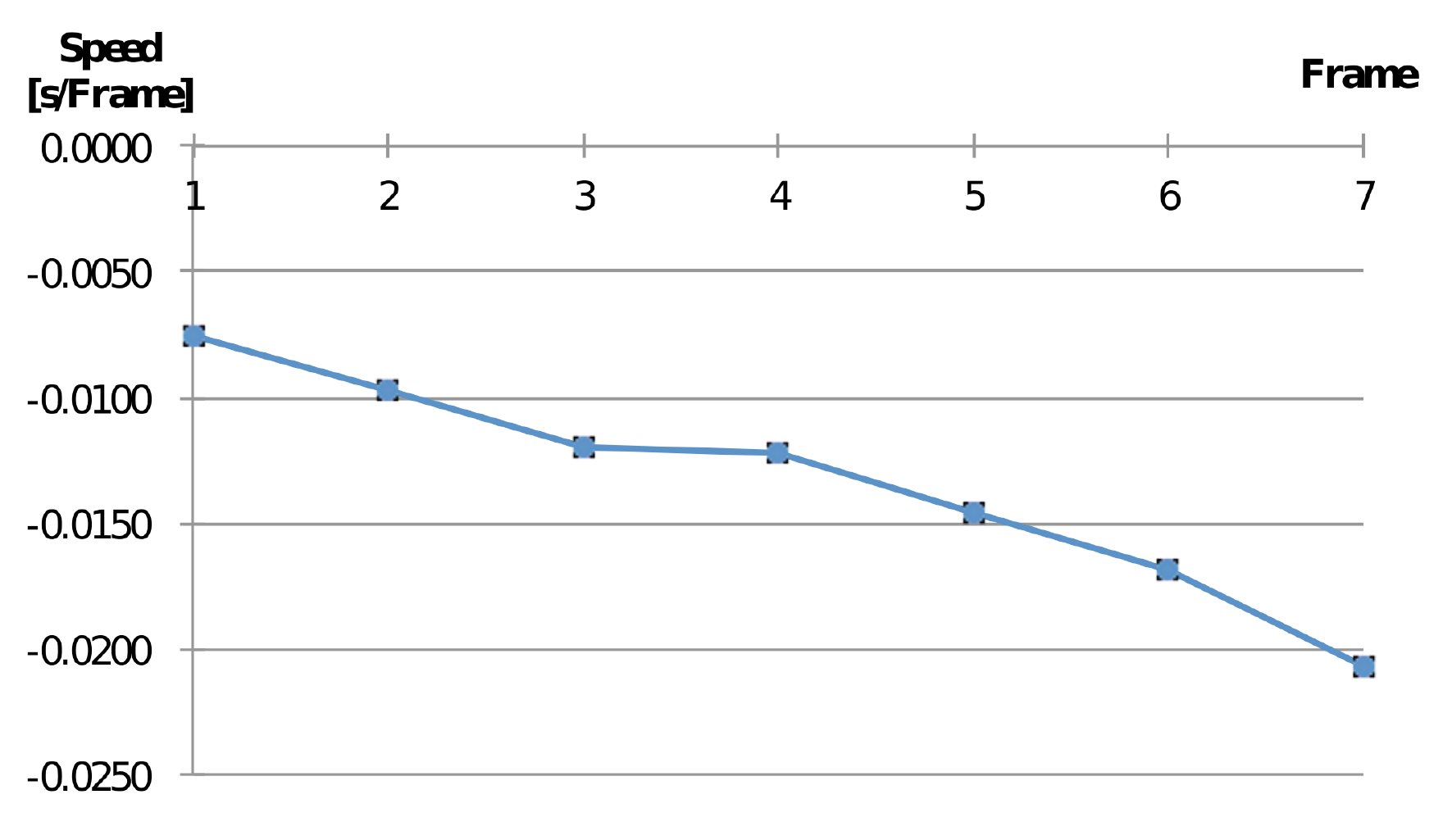}
	\caption{The graph of velocity of each board in Fig.\ref{fig:super_reso_accel}.
%the center of the gravity and 
% showing the velocity of each shape. 
Constant acceleration is confirmed.}
%	\caption{\knote{板。計測画像、結果のグラフ　超解像の加速している様子を表すグラフ}}
	\label{fig:super_reso_accel_image}
\end{center}
    \end{minipage}
\vspace{-5mm}
\end{figure}

Next, super-resolved shapes were reconstructed from a single image input. To confirm 
the effectiveness of the refinement algorithm which can estimate nonuniform velocity, we impose a constant acceleration to the target board using the 
motorized stage. Reconstruction results are shown in 
Fig.\ref{fig:super_reso_accel} and \ref{fig:super_reso_accel_image}.
As shown in the figures, we can confirm that the flat boards were not 
reconstructed at constant intervals, but at squared intervals.

\jptext{
ターゲットは、板か、振り子 %と、自由形状（剛体と非剛体）

1フレーム計測して、数枚に増やす。

その際に、リニアでないことを図で見せる。（特に板で加速している様子が重要）

＊精度評価はしない。%KinectやNCCと比較

%\subsection{Arbitrary shape reconstruction}
%
%ターゲットは、自由形状（剛体と非剛体）
%
%ただし1フレーム
%
%精度評価はしない。KinectやNCCと比較
}

\subsection{Arbitrary shape reconstruction}
%\subsection{Movie data reconstruction} %\knote{6.3と統合するかも}}

Then, we applied our method to shapes with curved surfaces and non-uniform 
texture using off-the-shelf DLP projectors~\cite{LightCrafter4500} with monochrome CMOS sensor.
%, which works with relatively high fps.
%Center of the wooden toy is placed 270mm apart from the lens.
The fps of the camera was 300Hz with 1024*768 resolution and 1800Hz for the 
projector with 912*1140 resolution. Note that 300Hz is almost the maximum fps 
among readily available CMOS sensors, whereas potential fps of consumer DLP 
projector is yet much higher than 1800Hz.

\begin{figure*}[tb]
		\centering
		\begin{tabular}{ccccc}

			\includegraphics[clip, width=3.2cm]{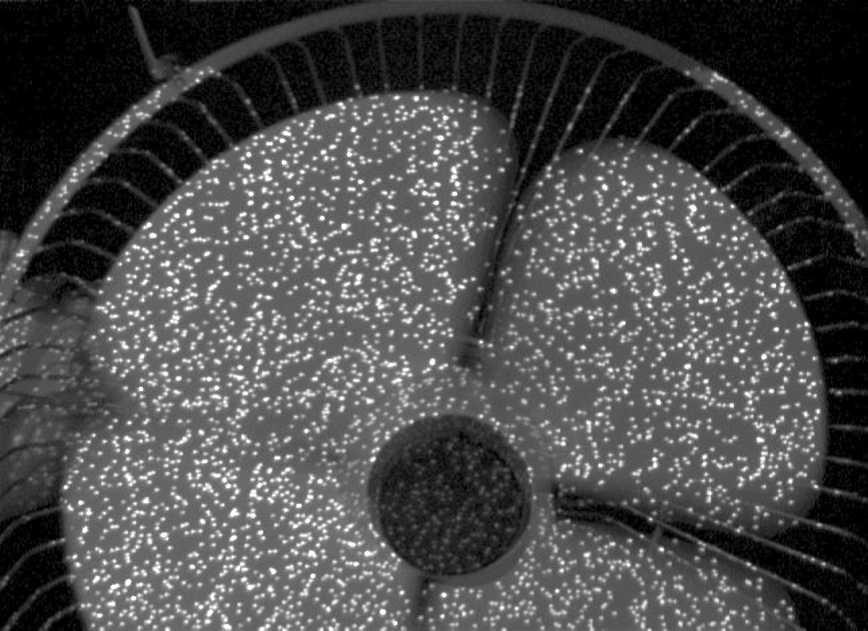} &
			\includegraphics[clip, width=3.2cm]{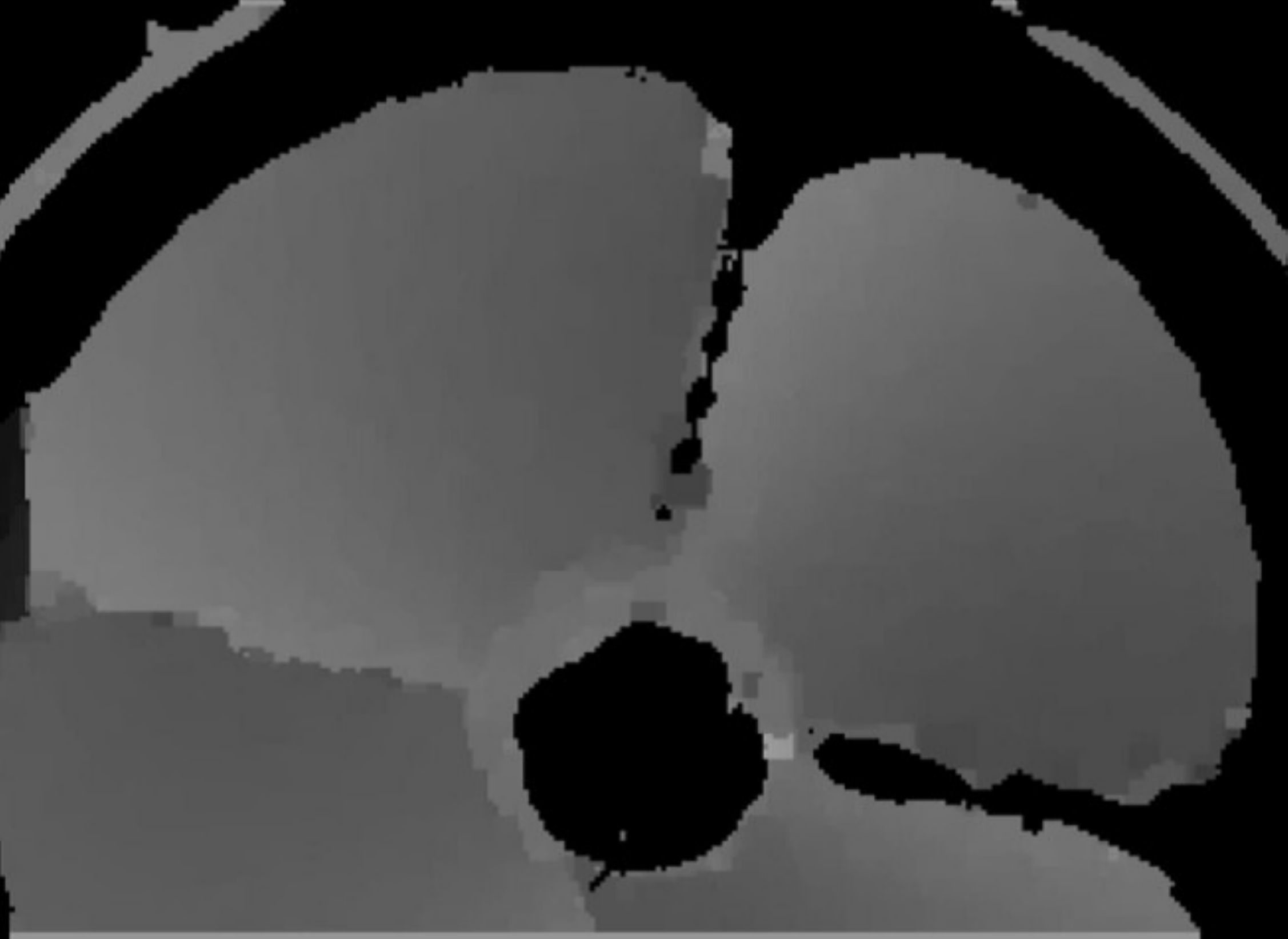} &
			\includegraphics[clip, width=3.2cm]{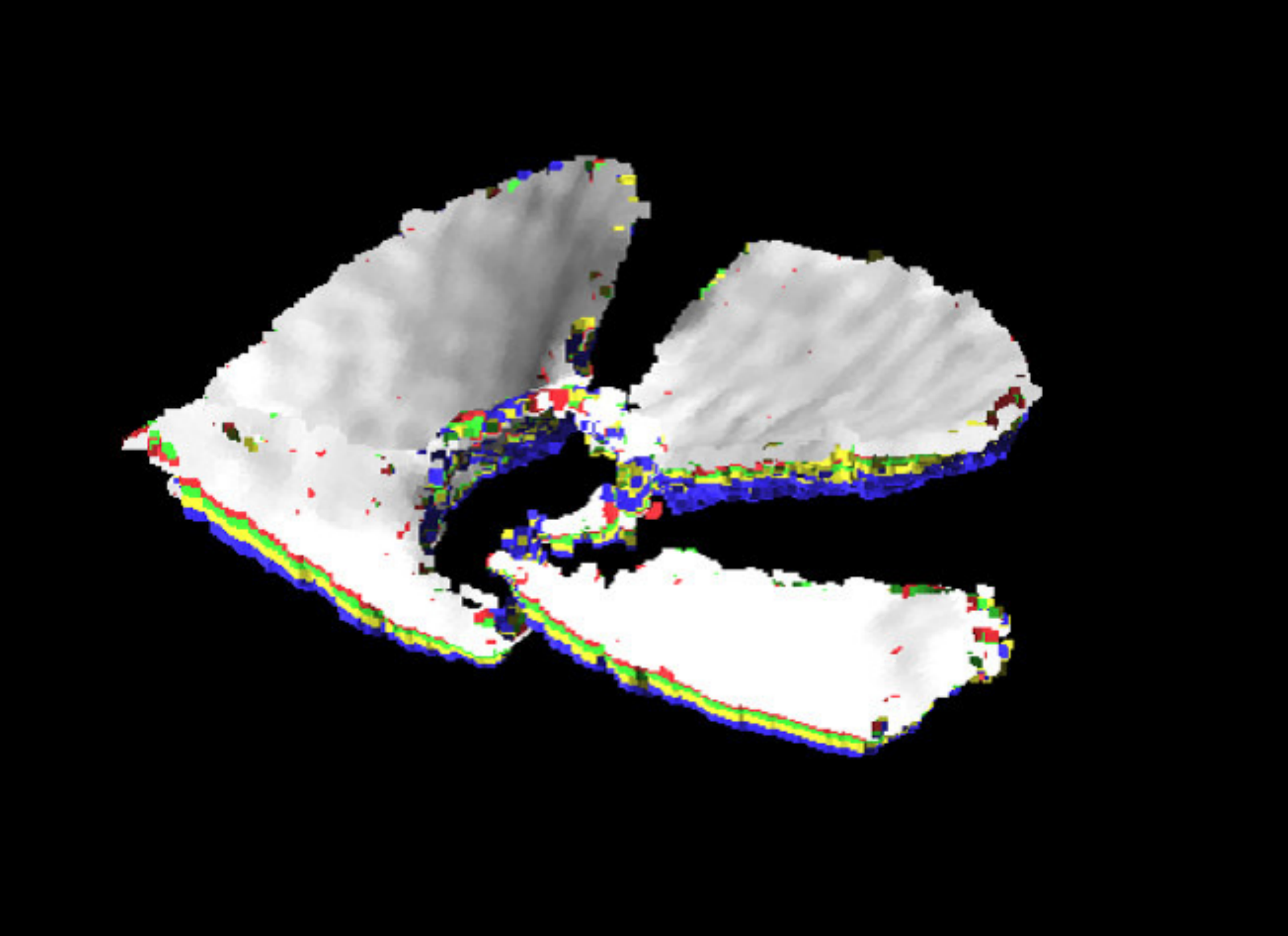}&
			\includegraphics[clip, width=3.2cm]{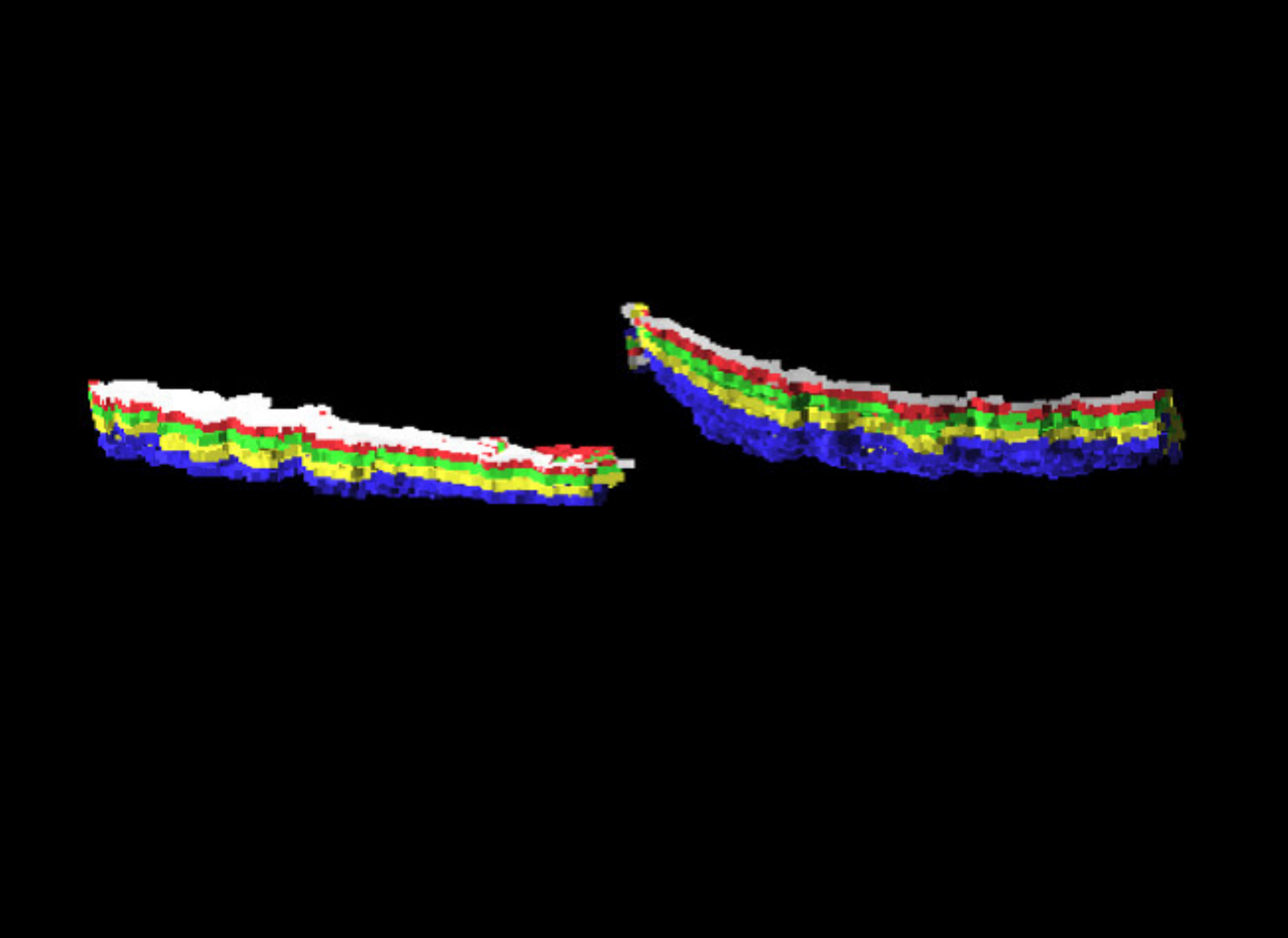}\\

			\includegraphics[clip, width=3.2cm]{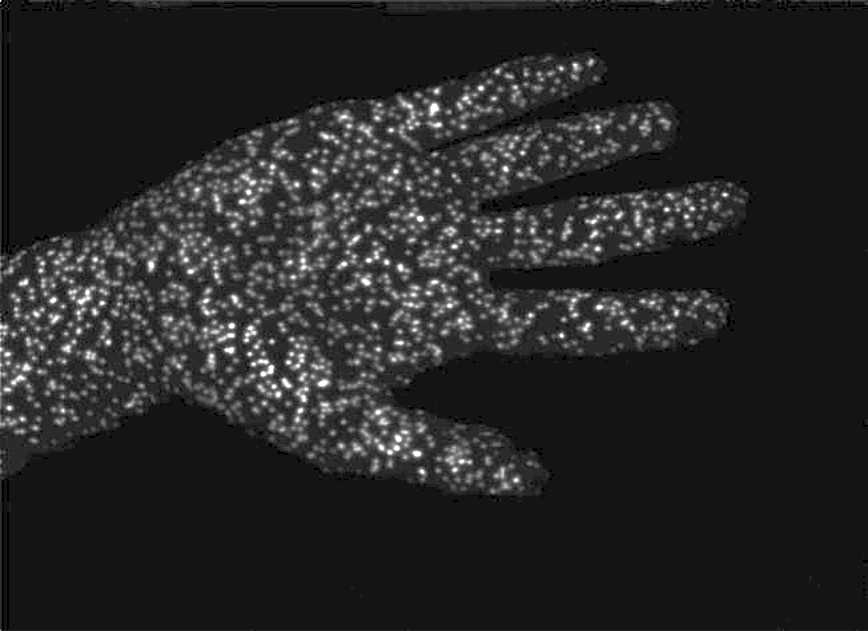} &
			\includegraphics[clip, width=3.2cm]{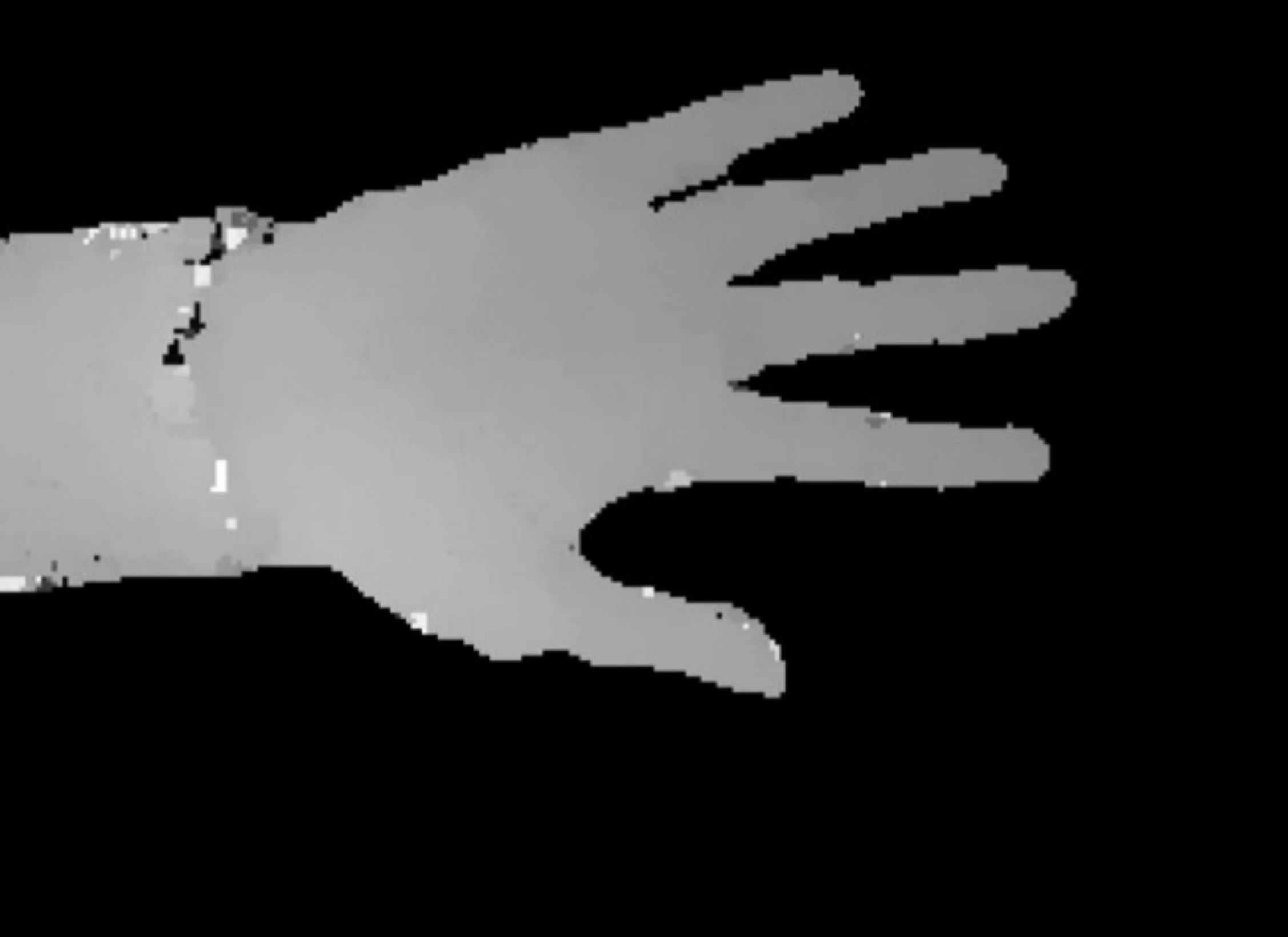} &
			\includegraphics[clip, width=3.2cm]{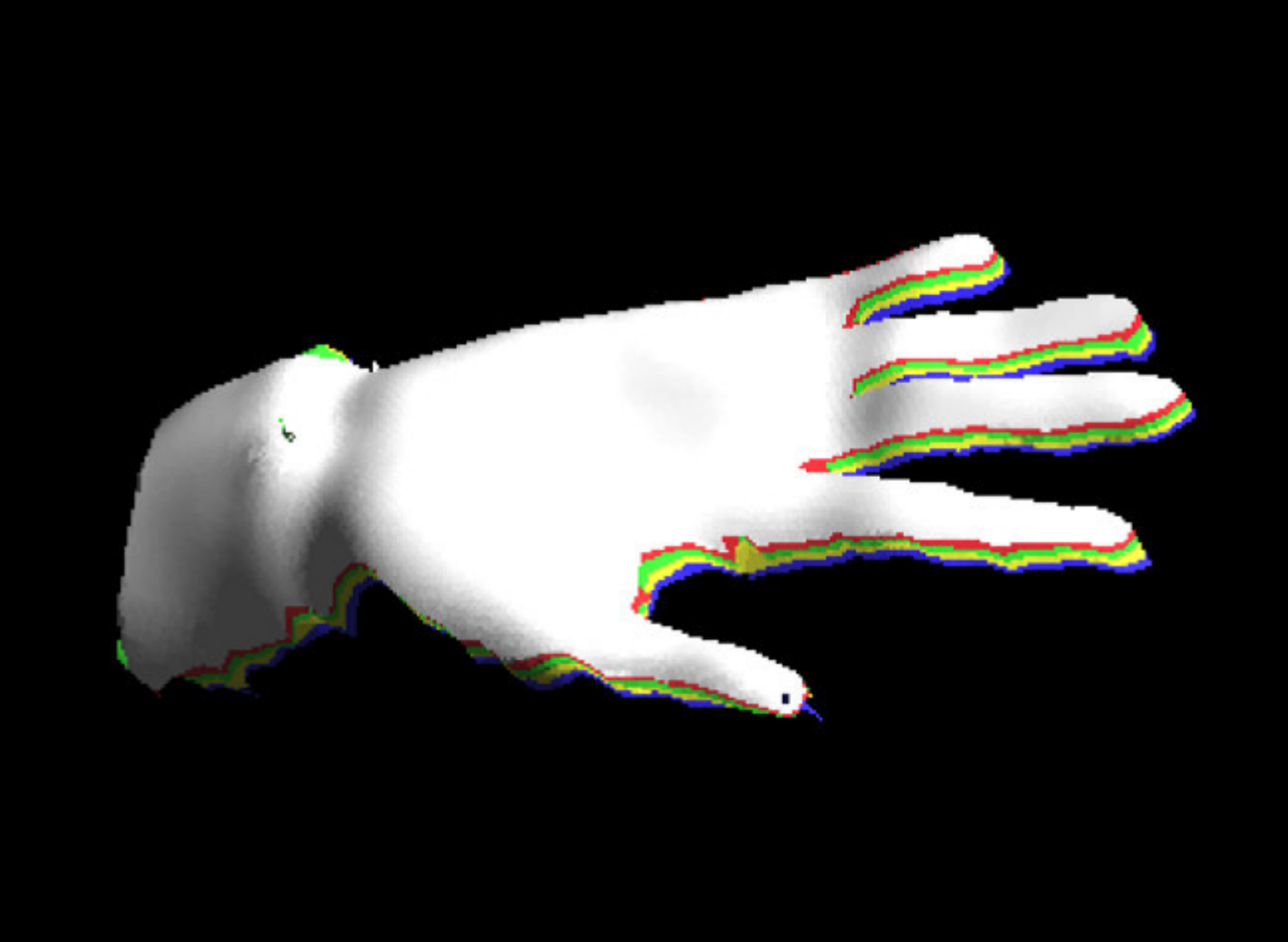}&
			\includegraphics[clip, width=3.2cm]{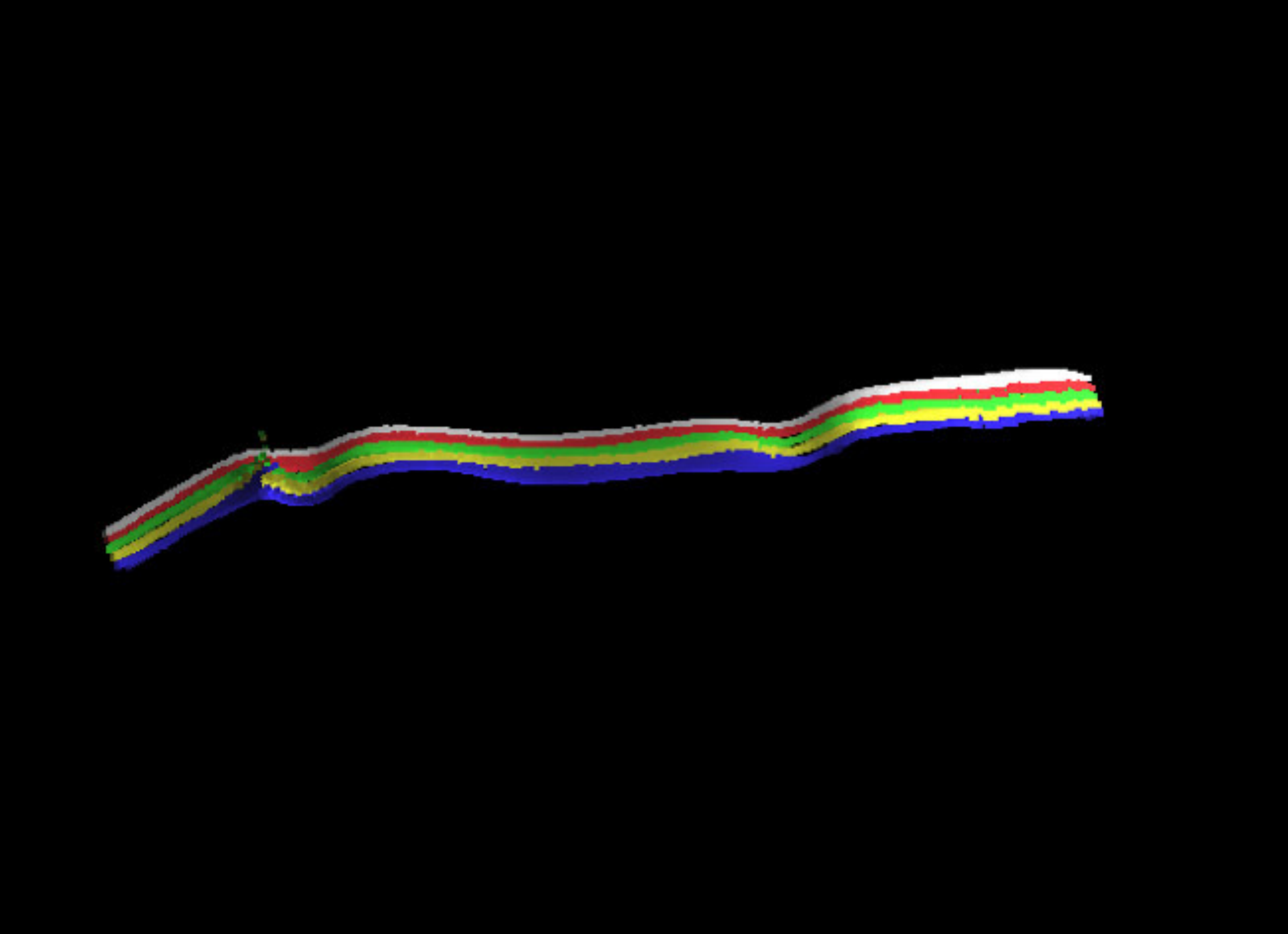}\\

			\includegraphics[clip, width=3.2cm]{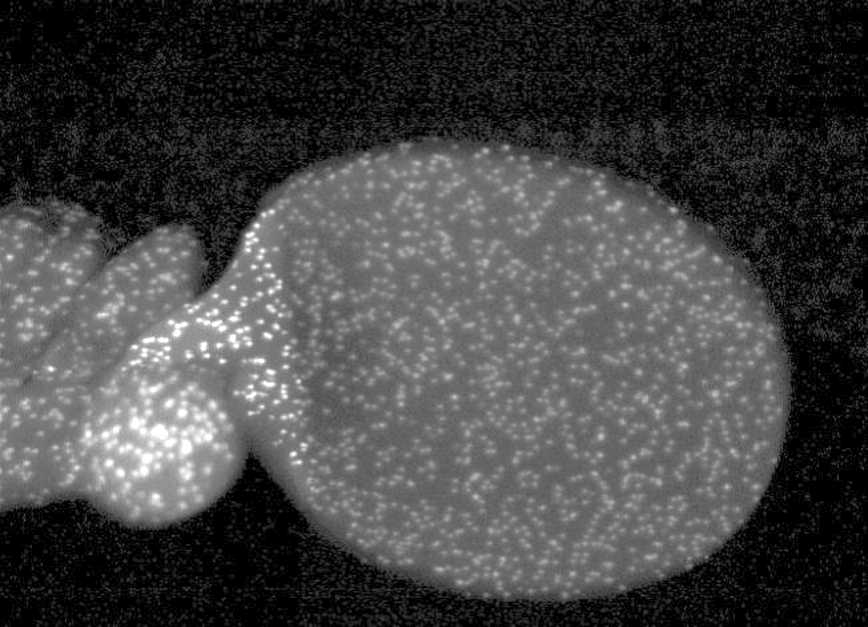} &
			\includegraphics[clip, width=3.2cm]{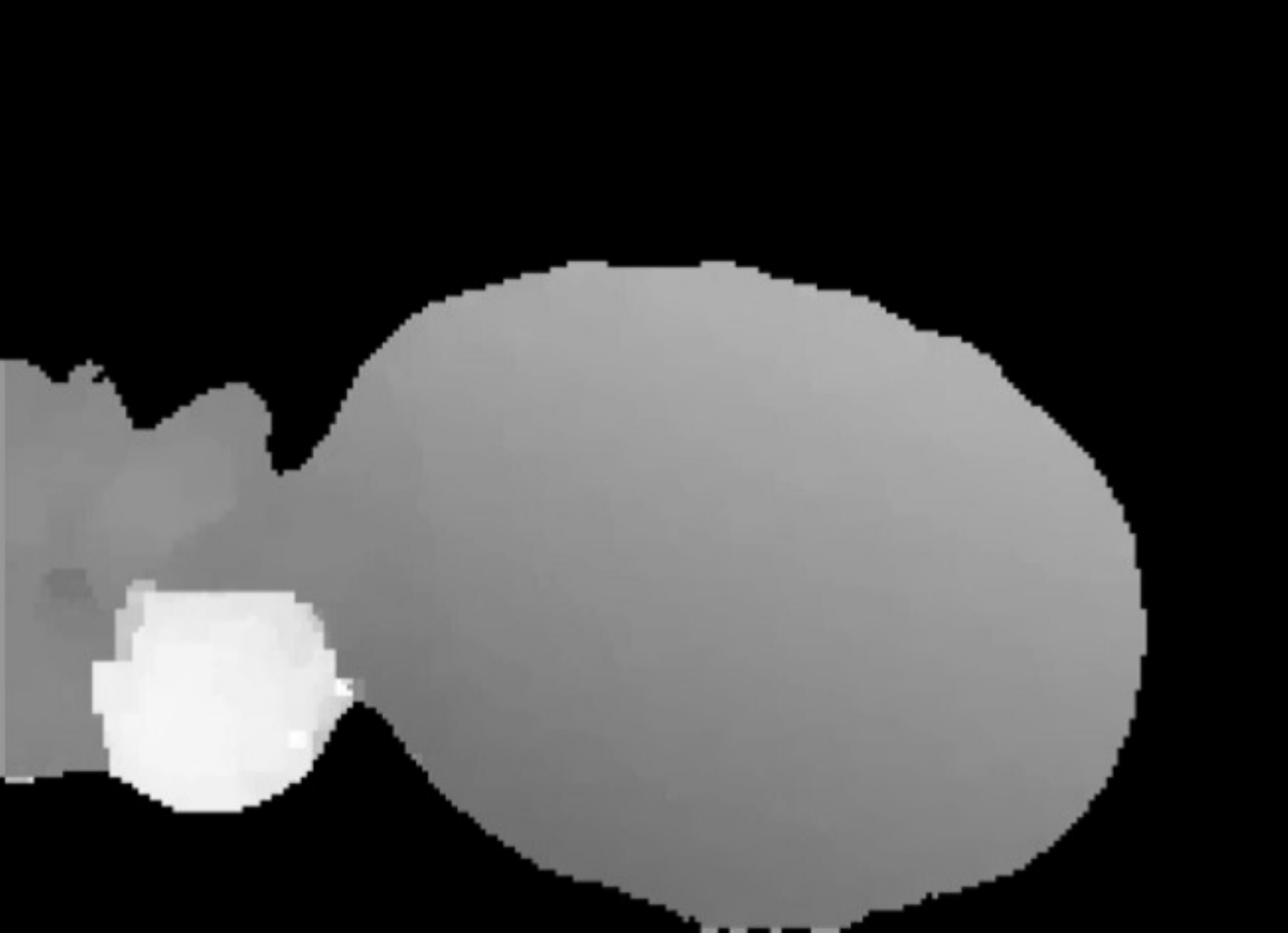} &
			\includegraphics[clip, width=3.2cm]{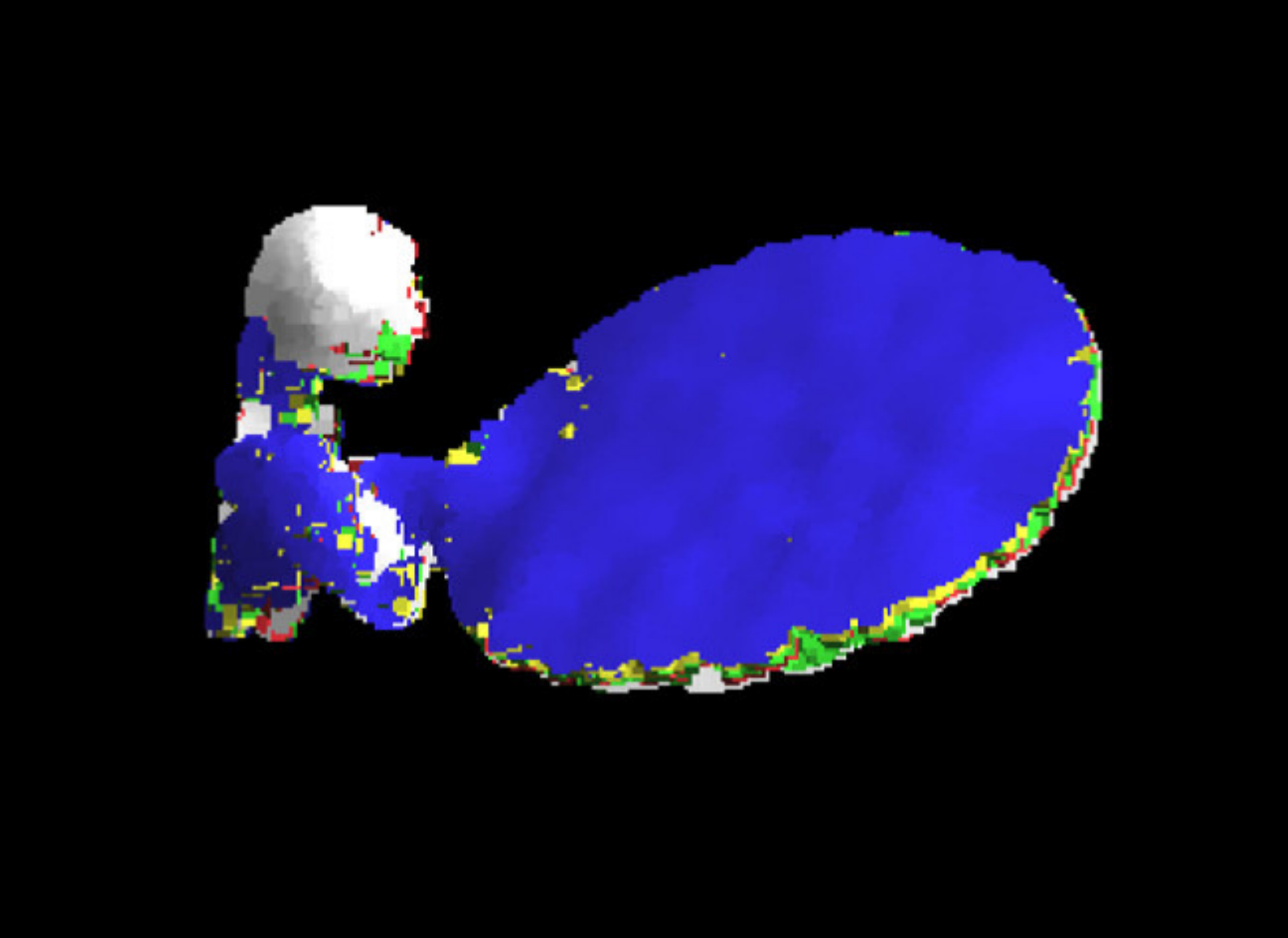}&
			\includegraphics[clip, width=3.2cm]{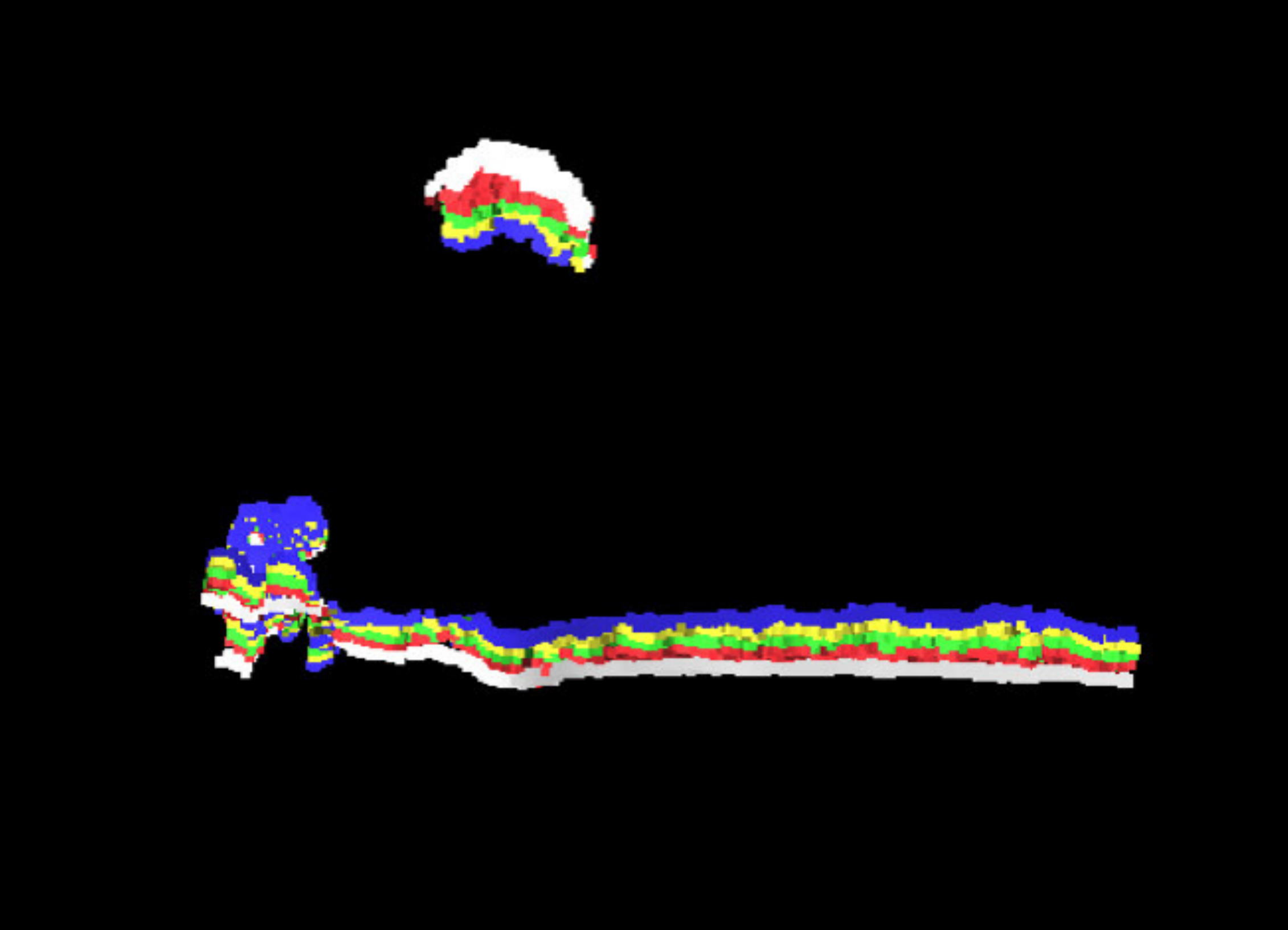}\\

			(a) & (b) & (c)&(d) \\
		\end{tabular}
		%    
		%\vspace{10cm}
	\caption{Results of temporal super-resolution of moving objects. (a) 
    captured image, (b) depth image, (c) reconstructed shape and (d) cross section of (c).}
%\knote{自由形状の結果の図}}
	\label{fig:Sample_ShapeRestoration}
\end{figure*}

The target objects were placed between 500mm and 800mm from the projector.
%Fig.\ref{fig:Sample_ShapeRestoration}(b) shows the reflected pattern and (c) shows the reconstruction results.
\if 0
The middle column of Fig.\ref{fig:Sample_ShapeRestoration} shows the actual captured 
image with pattern projection and the right column shows the reconstruction 
results.
\fi
Captured images are
shown in Fig.\ref{fig:Sample_ShapeRestoration}(a) and 
the middle column (b) %and (d) of Fig.\ref{fig:Sample_ShapeRestoration} 
shows 
the reconstruction 
results with depth image. 3D mesh and  the cross section of super-resolved 
multiple shapes are shown in (c) and (d).
% with constant three pixel intervals for fast calculation 
%and the right column shows all pixel reconstruction to show capability of dense 
%reconstruction. %, respectively.
%Fig.\ref{fig:Sample_ShapeRestoration}(j) shows zoom-in views of the dense reconstruction results of (a).
From the results, we can confirm that the multiple shapes of curved surfaces are recovered accurately. % without any postprocess. %smoothing even if there is texture on it.
%
%In the results, we also observe some parts are missing in the 3D shapes.
%%We consider 
%This is because the texture presents dark areas where no pattern is observed. 
%Efficient high dynamic range capturing system is desired.

\jptext{
連続フレーム（でなくても良い）

布や人体などの非剛体が良い

現状では、タマゴ、けん玉、手、で復元成功

＊比較や数値評価はしない。
}

\subsection{Simultaneous estimation of depths and velocities}

\begin{figure*}[tb]
		\centering
		\begin{tabular}{ccccccc}
			
			\includegraphics[clip, height=2.3cm]{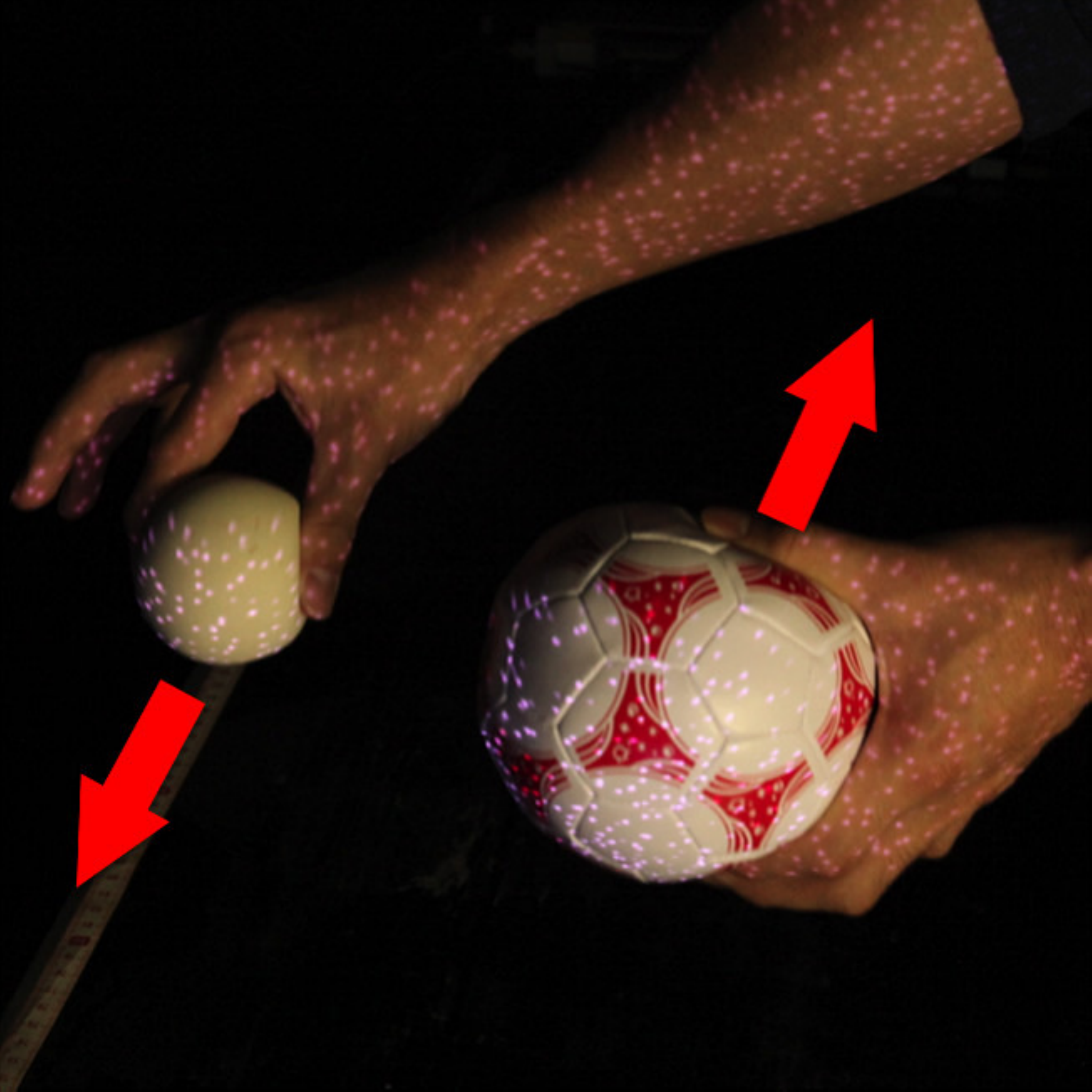} &
			\includegraphics[clip, height=2.3cm]{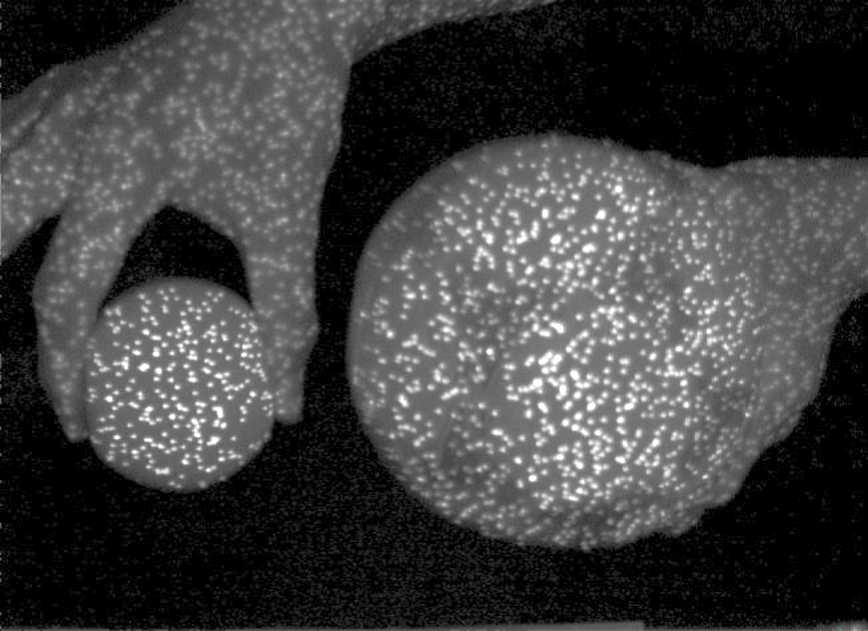} &
			\includegraphics[clip, height=2.3cm]{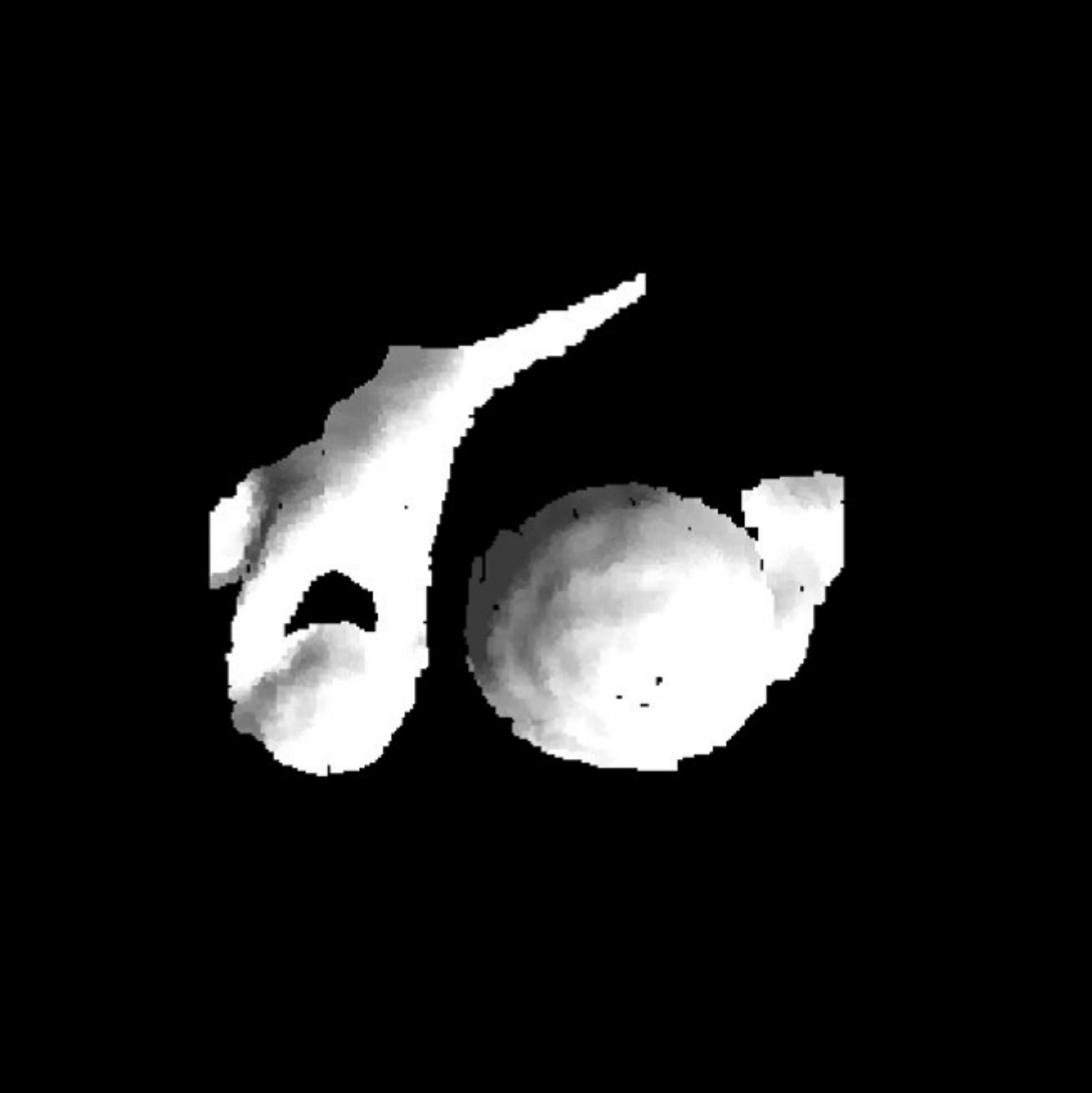} &

			\includegraphics[clip, height=2.3cm]{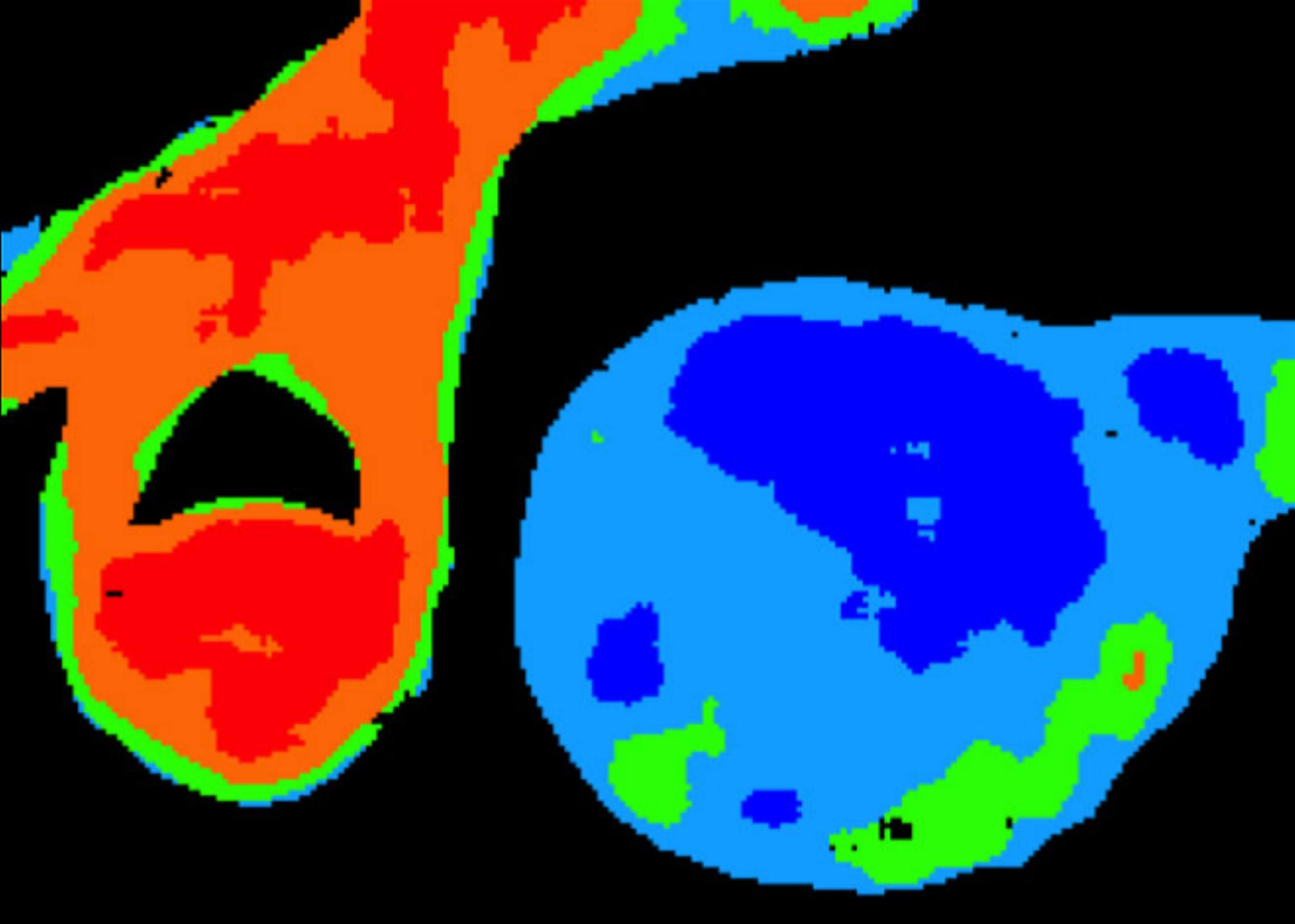} &
			\includegraphics[clip, height=2.3cm]{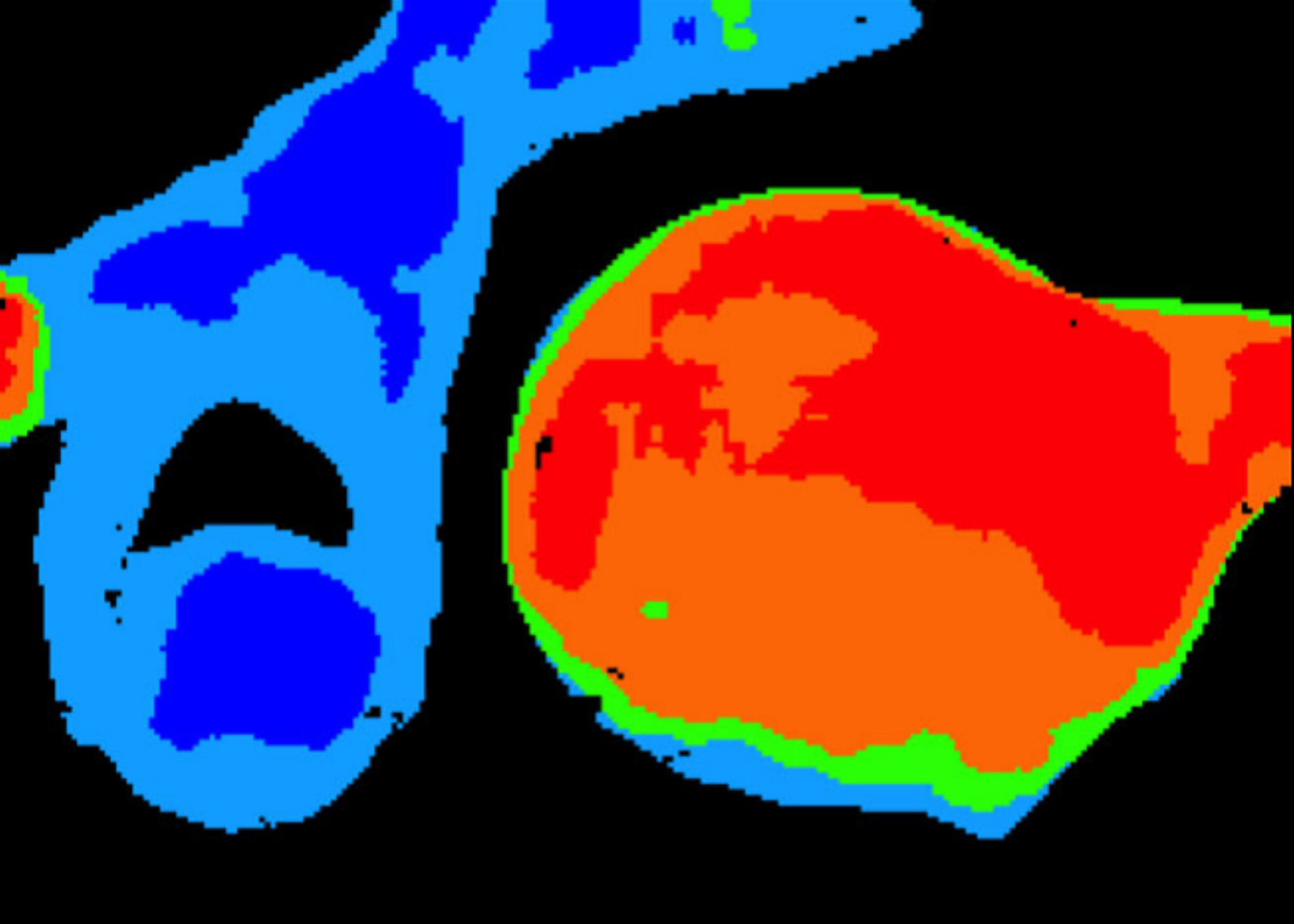} &
			\includegraphics[clip, width=0.7cm]{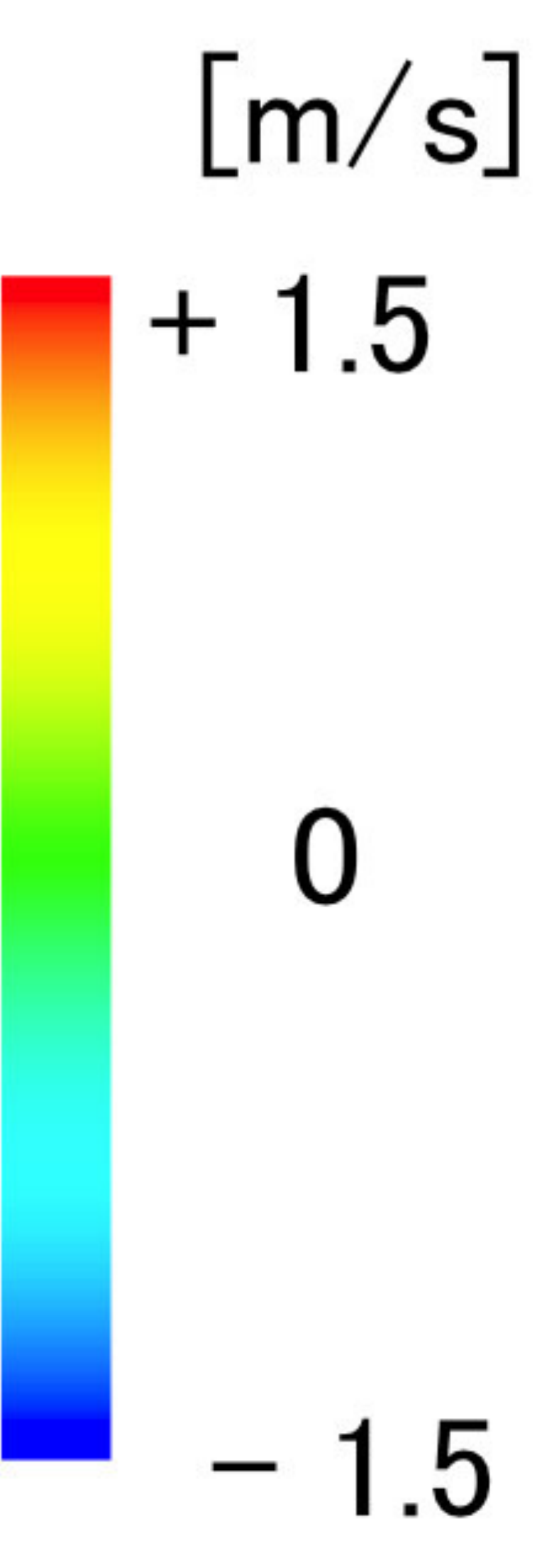}\\
%						(a) & (b) & (c) \\
						(a) & (b) & (c)& (d) & (e)  \\
		\end{tabular}
		%    
		%\vspace{10cm}
	\caption{Velocity estimation: (a) the experimental scene, (b) the captured  image, (c) the reconstructed shape,
	(d,e) velocity maps for two frames.}
%\knote{自由形状の結果の図}}
	\label{fig:clormap_ShapeRestoration}
\end{figure*}

To show the ability of the proposed system to estimate both depth and velocities simultaneously, 
we captured two balls in the both hands, shaken as fast as possible. 
Fig.~\ref{fig:clormap_ShapeRestoration} shows the results, where
(a) shows the capturing scene,
(b) shows the actually captured image for reconstruction,
(c) shows the reconstructed shapes, and
(d) and (e)  show the color-mapped velocity for the direction along the optical axis.
Note that these velocities are estimated from each single frame, rather than from multiple images. 
From the color maps, we can confirm that the left ball is moving toward the camera,
and the right ball is moving from the camera in the frame (d),
%both bolls becomes slow in frame (b), 
and the velocities are altered in frame (e). 

\subsection{Fast motion reconstruction}
\begin{figure*}[htb]
\vspace{-5mm}
\centering
%\begin{minipage}[b]{0.22\textwidth}
        \centering
        \begin{tabular}{cccc}
      \hspace*{-4mm}  \includegraphics[height=2.5cm]{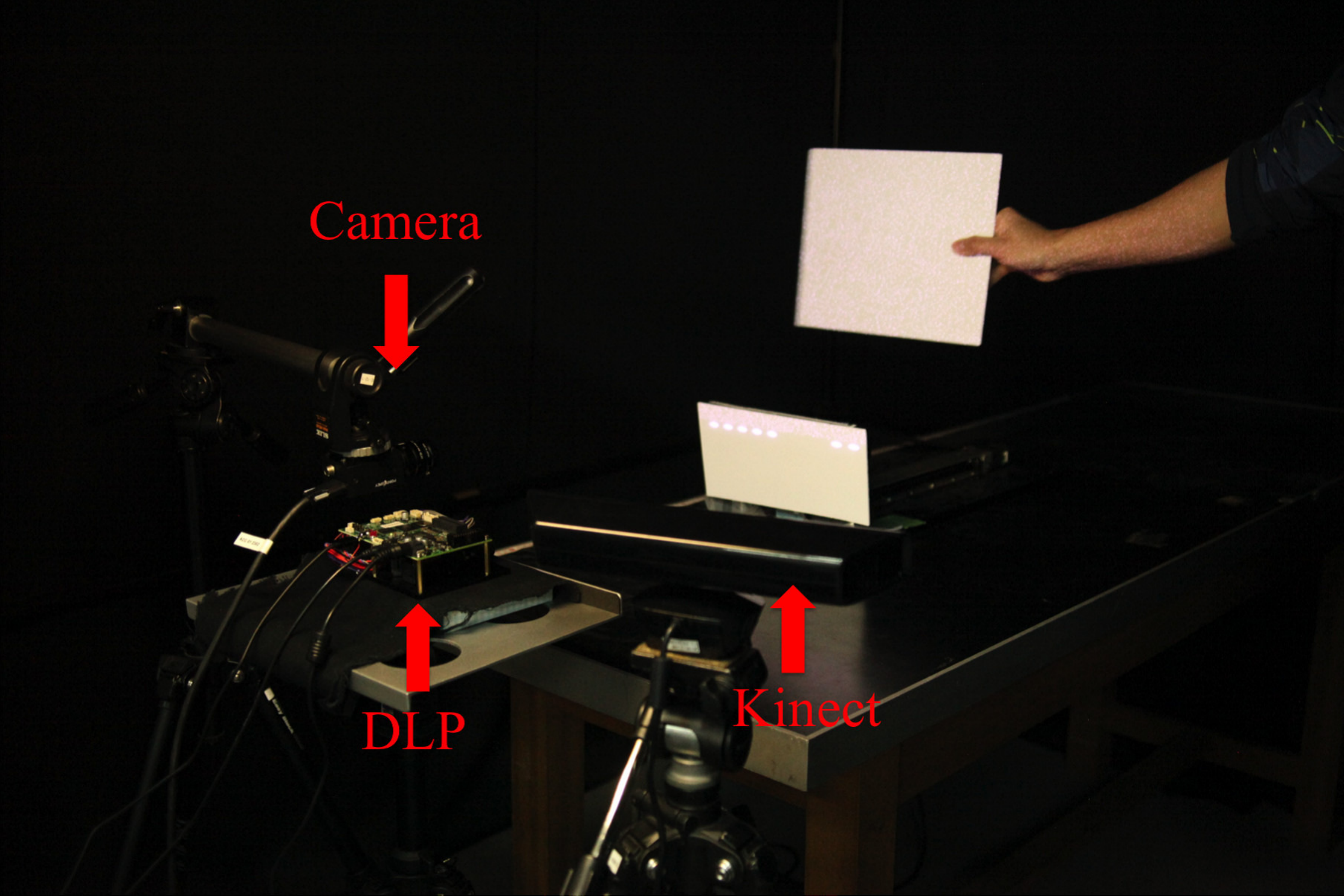}&
%       \hspace*{-4mm} \includegraphics[height=2.5cm]{arxiv_figs/vskinect_input-eps-converted-to.pdf}&
       \hspace*{-4mm} \includegraphics[height=2.5cm]{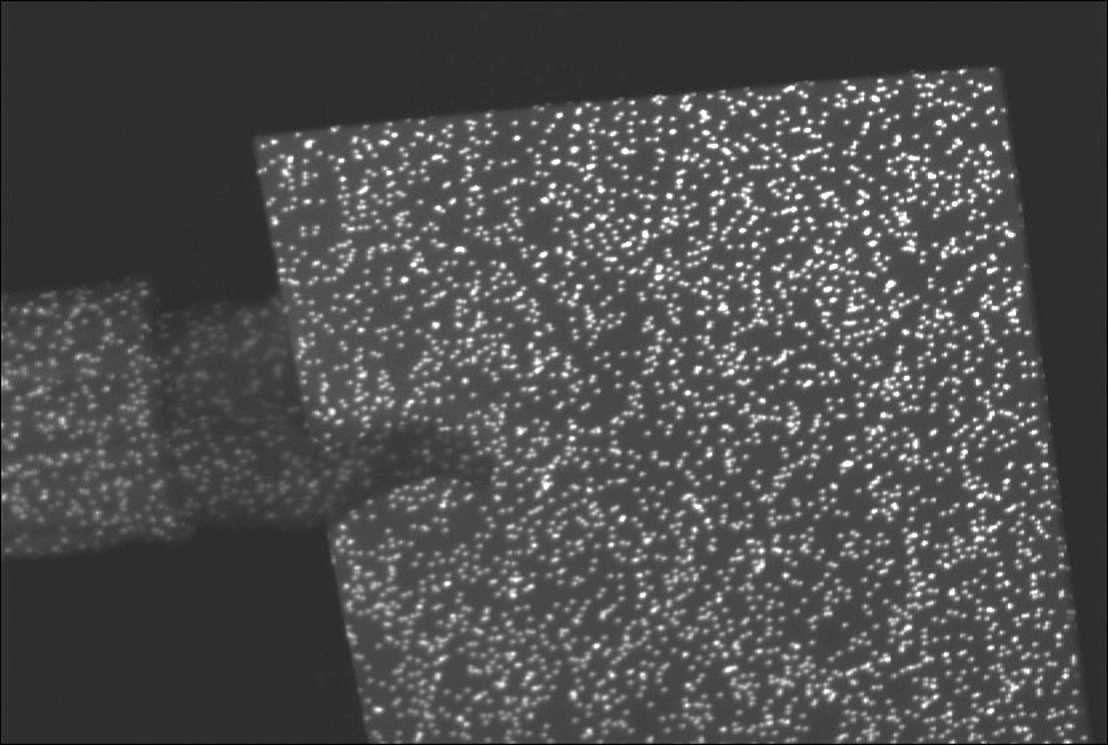}&
        \hspace*{-4mm}		\includegraphics[clip, height=2.5cm]{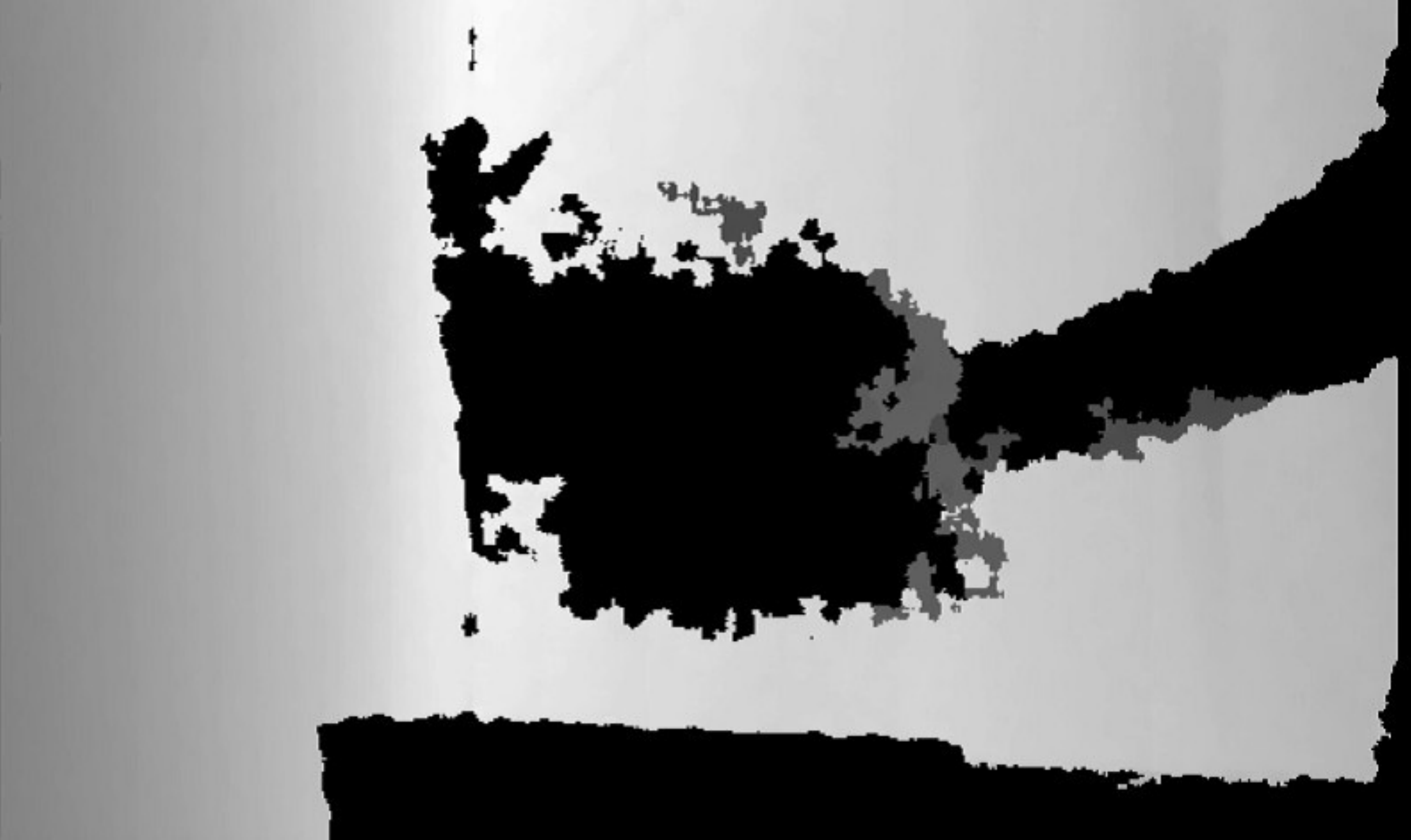} &
\hspace*{-4mm}\includegraphics[clip, height=2.5cm]{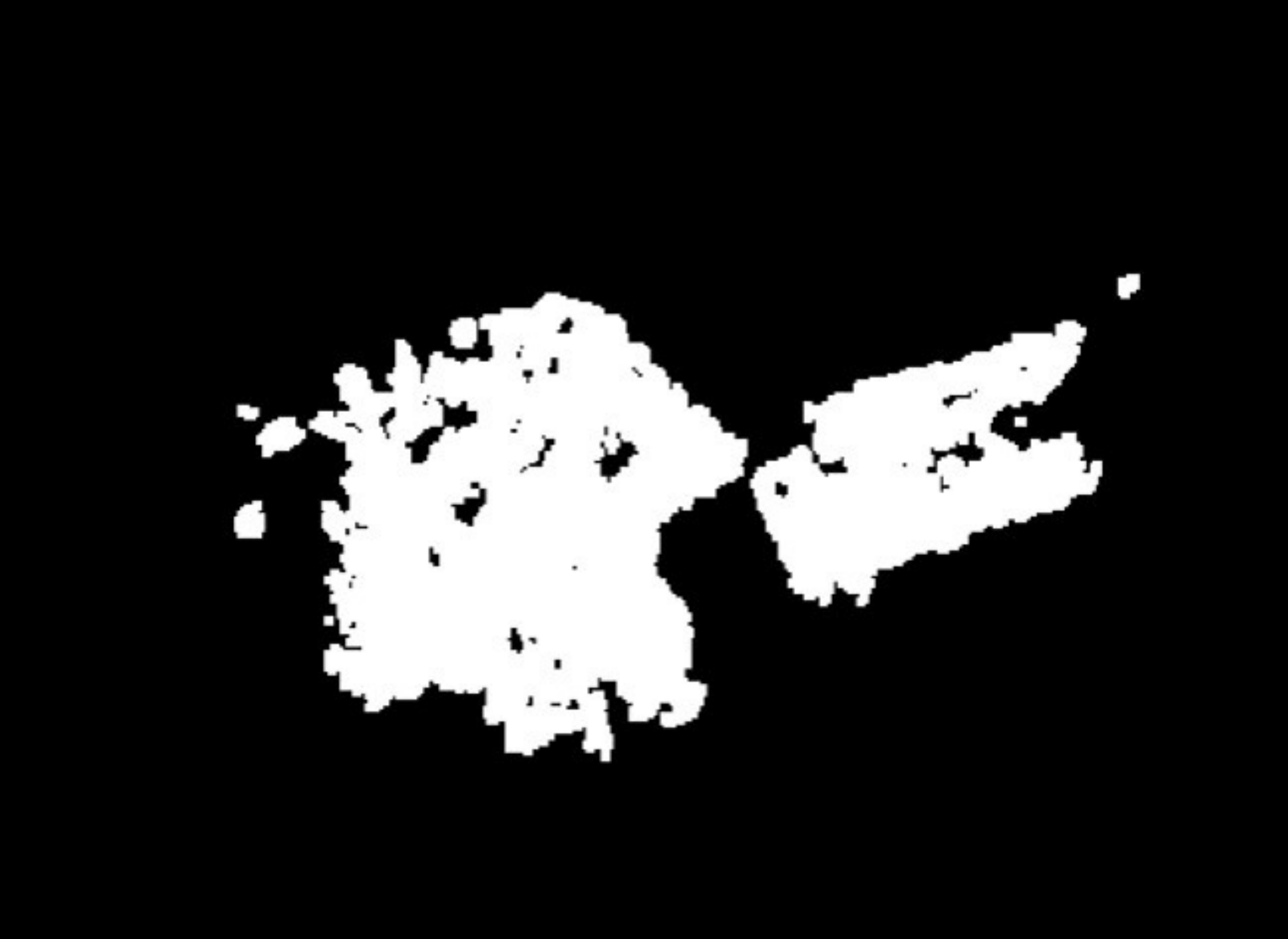} \\
(a)&(b)&(c)&(d)\\
			\hspace*{-4mm}\includegraphics[clip, height=2.5cm]{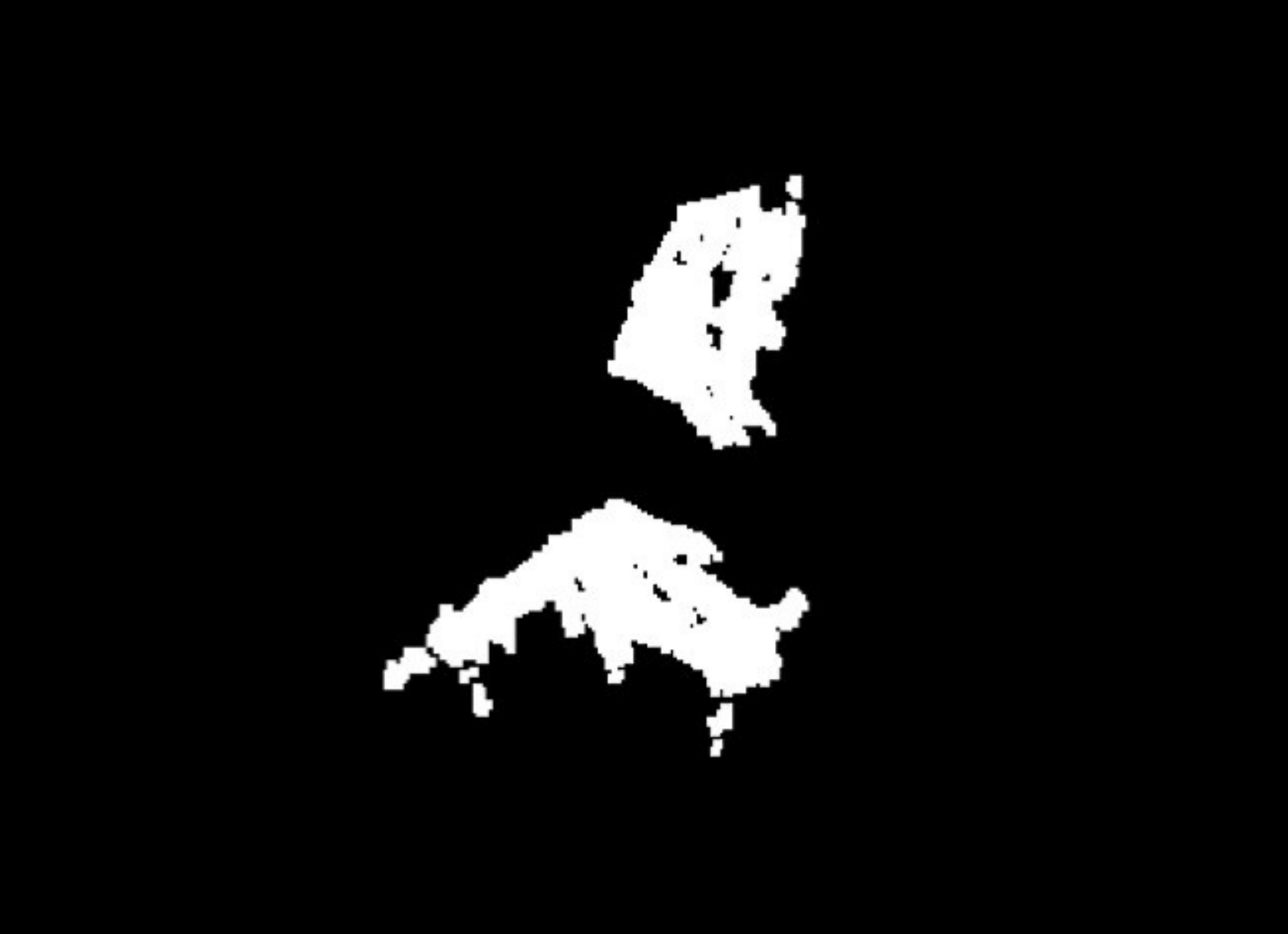} &
		\hspace*{-4mm}	\includegraphics[clip, height=2.5cm]{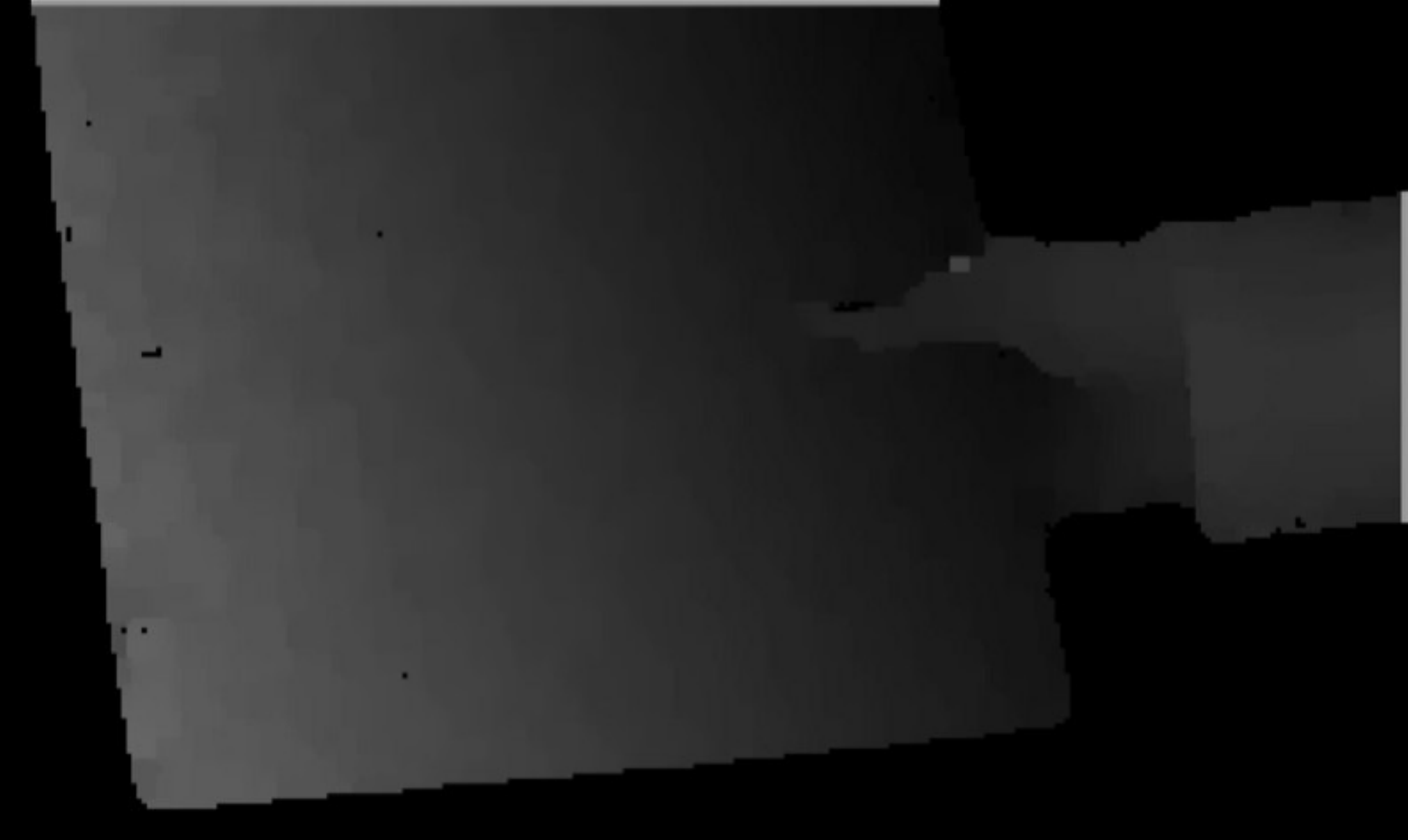}&
		\hspace*{-4mm}	\includegraphics[clip, height=2.5cm]{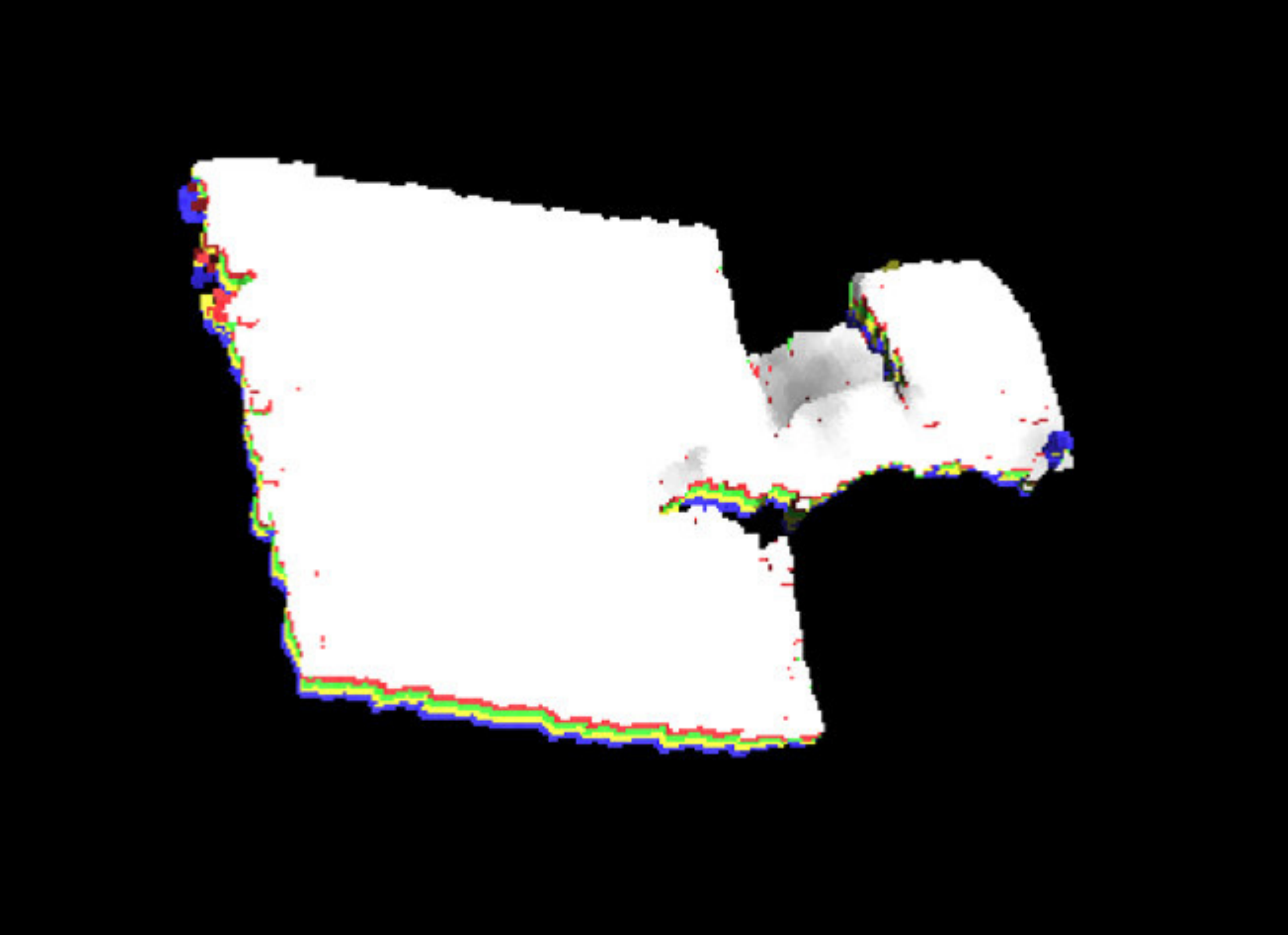}&
	\hspace*{-4mm}\includegraphics[clip, height=2.5cm]{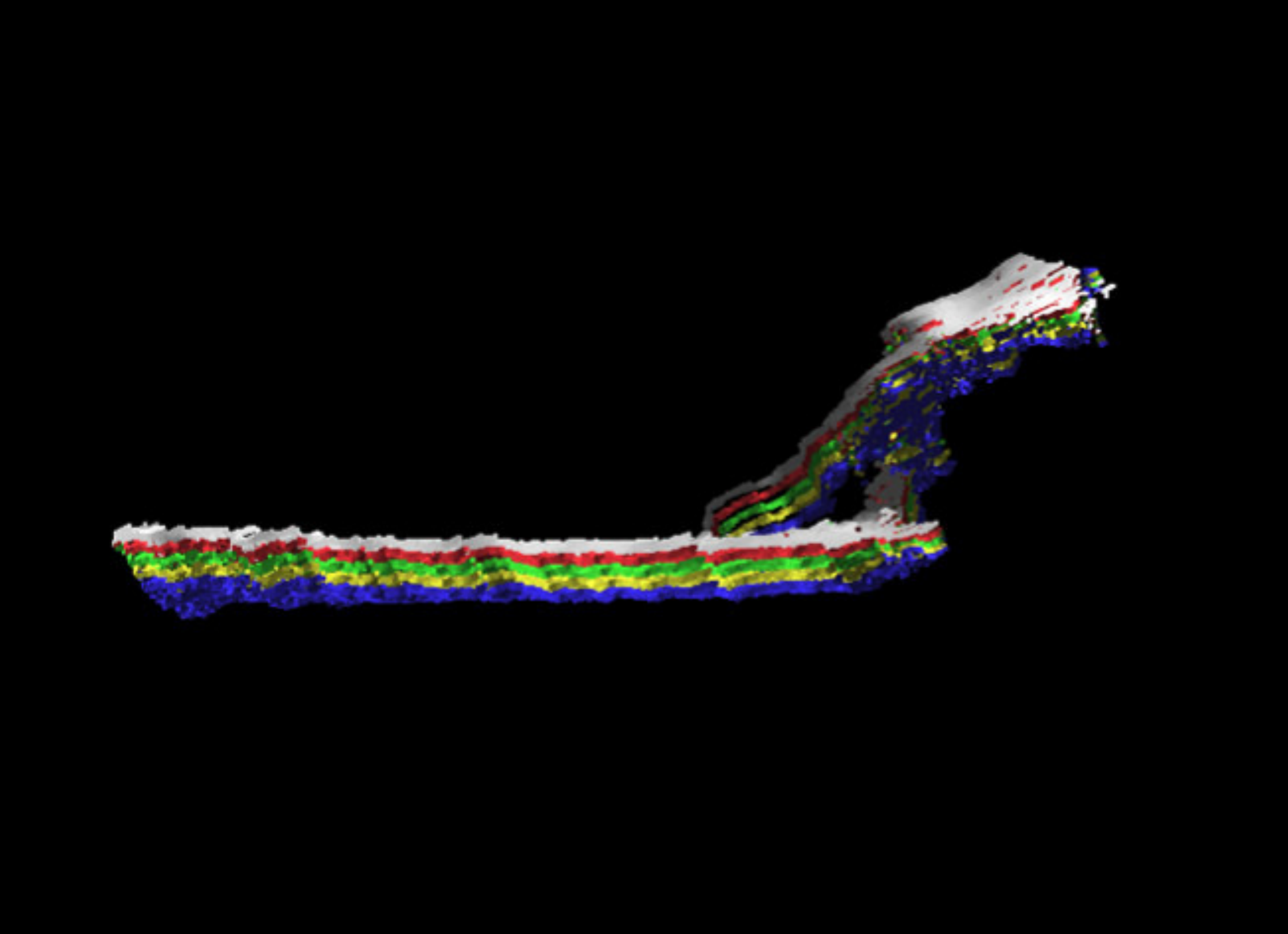}\\
(e)&(f)&(g)&(h)\\
\end{tabular}
	\caption{Capturing fast moving object.
	(a) Actual set-up with 
    Kinect and our system.
    (b) Captured image of fast moving board with normal CMOS camera of
    300 fps and no blur observed. 
%Although our multiple projection mitigate the blur, still blur is observed in the captured image.
    Result of fast motion with Kinect in depthmap (c) and (d,e) 3D shapes.
    Result of the proposed system in depthmap (f) and (g,h) 3D shapes colored by intra-frames.}
%	\caption{\knote{セットアップ画像}}
	\label{fig:comp_Kinect}
%\end{minipage}
%\\
\vspace{-5mm}
\end{figure*}

%\knote{間に合わなければカット。インパクトの上では載せたい}

Finally, we captured the fast moving object by both our technique and  Kinect 
v1~\cite{Kinect}; the board was swung as fast as possible by hand and 
the same scene was captured by both devices as shown in Fig.~\ref{fig:comp_Kinect}(a).
Since Kinect is not designed for scanning a fast moving object, 
the purpose of using Kinect is to show the potential of our method compared with standard 
3D scanning technique, but not to intend to say that our technique is better than Kinect.
The fps of the camera and the projector is as same as the 
previous experiment, \ie 300Hz and 1800Hz, respectively.
As can be seen in Fig.~\ref{fig:comp_Kinect}(b), there is no blur captured  
by our method, and accumulated sharp patterns are observed.
% in the captured image. 
%using a common CMOS camera. 
% set at 15fps with a shutter speed of 50msec:
%standard specs for common consumer cameras.
As the results shown in Fig.~\ref{fig:comp_Kinect}(c-e), only corrupted shapes are reconstructed 
%near maximum speed 
with Kinect, whereas our method can recover the temporally super-resolved shapes 
with high accuracy as shown in Fig.~\ref{fig:comp_Kinect}(f-h).

\section{Limitations}

Since our method detects the change rate of distance on the line of sight 
corresponding to each pixel, not all object movements can be recovered, 
especially, the motion perpendicular to the optical axis. One simple 
solution to detect 3D motion along the optical axis is to use multiple bands; 
another camera for visible light for estimating optical flow added to the IR 
pattern and IR camera for our method.

Similar to the previous limitation, if the object shifts in X-Y directions with 
textures of high spatial frequency and high contrast, negative effects can be 
caused. However, these are open problems for all the active 3D scanning 
techniques, and not specific to our method. Using IR illuminations could also be a 
practical solution. 

Another limitation is occluding boundaries in the X-Y direction. With the 
current implementation, the first and last patterns are fixed for all the pixels, 
and the shape reconstruction sometimes fails or gets worse. We tested a
simple solution of enlarging the search space, however, it significantly 
increases the processing time and sometimes causes 
unstable results. Therefore, to prioritize the stability, we did not adopt this 
technique in the experiments and a finding a practical solution is an important 
next step in our research.

%As with the texture issue, the depth-of-field problem is not specific to our method, but is an open problem for active 3D scanning methods and is essentially not related to temporal super-resolution. However, since high speed systems are usually lacking in light intensity and depth-of-field tends to be shallow, we resolved it with the database approach for calibration, which is proposed in the paper. We will add such discussion in the revised version.
%}

\section{Conclusion}%\knote{川崎}}

In this paper, we propose a 
temporal super resolution technique for structured light systems to recover a
series of 3D 
shapes from a single image. %n initially reconstructed  single shape. 
To achieve this, we project multiple patterns with higher fps than the camera 
can capture, to 
    embed temporal information of object depth into a single captured image.
Since the projected pattern on the surface varies depending on the depth and motion 
of the object, those parameters should be estimated simultaneously.
%As for the solution, 
    Object depth as well as velocity for each pattern are sought to best explain the captured image, by 
    synthesizing the image from an image database which is created in advance.
Experiments were conducted to confirm the effectiveness our high-fps multiple 
pattern projection technique through
quantitative evaluation using a planar board with various materials.
We also show that temporally super-resolved shapes not captured in between frames can be 
successfully reconstructed using our method.

\section*{Acknowledgment}
\vspace{-0.2cm}
This work was supported in part by JSPS KAKENHI Grant No. 15H02758, 15H02779 and 16H02849, 
MIC SCOPE 171507010 and MSR CORE12.

%-------------------------------------------------------------------------
\clearpage
\clearpage
%\newpage
{\small
\bibliographystyle{ieee}
\bibliography{../bib/shortSTRING,../bib/JabRef,../bib/bibref,../bib/furukawa,temporal_superreso}
}

\end{document}